%% file: main.tex
\documentclass{article}

\usepackage{PRIMEarxiv}

\usepackage[utf8]{inputenc} % allow utf-8 input
\usepackage[T1]{fontenc}    % use 8-bit T1 fonts
\usepackage{hyperref}       % hyperlinks
\usepackage{url}            % simple URL typesetting
\usepackage{booktabs}       % professional-quality tables
\usepackage{amsfonts}       % blackboard math symbols
\usepackage{nicefrac}       % compact symbols for 1/2, etc.
\usepackage{microtype}      % microtypography
\usepackage{graphicx}       % graphics
\usepackage{natbib}
\usepackage{makecell}
\usepackage{algorithm2e}
\usepackage{comment}
\usepackage[table,x11names,dvipsnames]{xcolor}
\input{math_commands_old}

\title{Beyond FVD: Enhanced Evaluation Metrics for Video Generation Quality
}

\author{
  Ge Ya (Olga) Luo$^*$, Gian Mario Favero$^*$, Zhi Hao Luo \\
  Mila - Quebec Artificial Intelligence Institute \\
   \And
  Alexia Jolicoeur-Martineau \\
  Samsung - SAIT AI Lab, Montreal \\
  \AND
  Christopher Pal\\
  Mila - Quebec Artificial Intelligence Institute \\
  Canada CIFAR AI Chair 
}

\begin{document}
\maketitle
\def\thefootnote{*}\footnotetext{These authors contributed equally to this work}
\def\thefootnote{\arabic{footnote}}

\vspace{-0.55cm}
\begin{abstract}
The Fr\'echet Video Distance (FVD) is a widely adopted metric for evaluating video generation distribution quality. However, its effectiveness relies on critical assumptions. Our analysis reveals three significant limitations: (1) the non-Gaussianity of the Inflated 3D Convnet (I3D) feature space; (2) the insensitivity of I3D features to temporal distortions; (3) the impractical sample sizes required for reliable estimation. These findings undermine FVD's reliability and show that FVD falls short as a standalone metric for video generation evaluation. After extensive analysis of a wide range of metrics and backbone architectures, we propose \textbf{JEDi}, the \textbf{J}EPA \textbf{E}mbedding \textbf{Di}stance, 
based on features derived from a Joint Embedding Predictive Architecture, measured using Maximum Mean Discrepancy with polynomial kernel.  Our experiments on multiple open-source datasets show clear evidence that it is a superior alternative to the widely used FVD metric, requiring only 16\% of the samples to reach its steady value, while increasing alignment with human evaluation by 34\%, on average.\footnote{\textbf{Project page:} \url{https://oooolga.github.io/JEDi.github.io/}; \textbf{Code: }\url{https://github.com/oooolga/JEDi}.}
\end{abstract}

% and improving alignment with human evaluations
% , we propose JEDi, the \textbf{J}EPA \textbf{E}mbedding \textbf{Di}stance, based on features derived from a Joint Embedding Predictive Architecture, measured using Maximum Mean Discrepancy with a polynomial kernel.  
%, we propose \textbf{V-JEPA-MMD} -- a \textbf{V}ideo quality metric based on the use of a \textbf{J}oint \textbf{E}mbedding \textbf{P}redictive \textbf{A}rchitecture, and measured using \textbf{M}aximum \textbf{M}ean \textbf{D}iscrepancy. 

% keywords can be removed
% \keywords{Video quality metrics \and Fréchet Video Distance (FVD) \and Inflated 3D Convnet (I3D) \and Video Joint Embedding Predictive Architecture (VJEPA)}
%\vspace{-.3cm}
\section{Introduction}
\begin{figure}[!h]%
    \vspace{-.32cm}
    \centering
    {\includegraphics[width=0.55\linewidth]{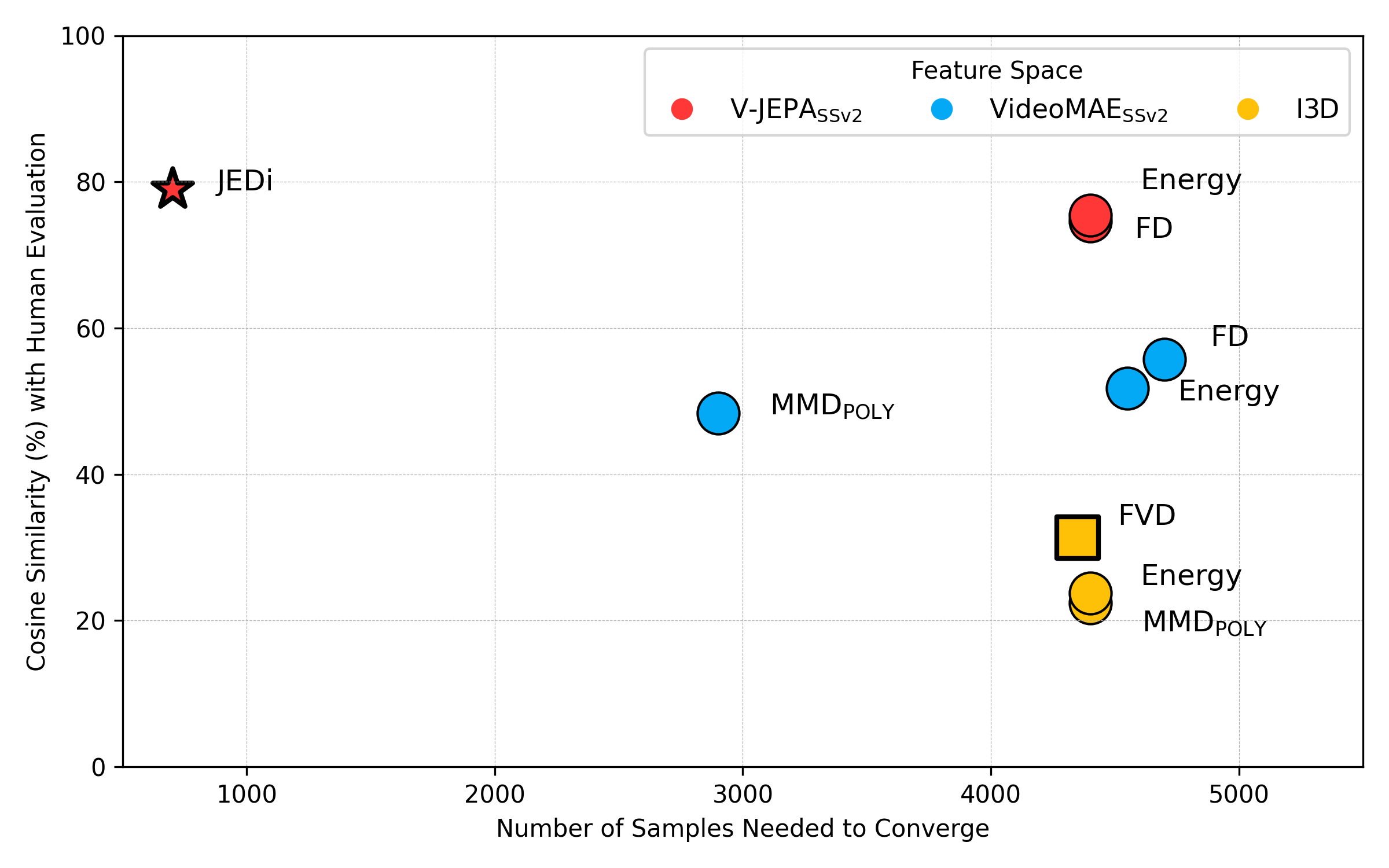} }%
    %[FOR ICLR]{\includegraphics[width=0.7\linewidth]{figs/ucf_samples_plot.png} }%
    \caption{Comparing the number of samples that Fr\'echet Distance (FD), Energy, and $\text{MMD}_{\text{POLY}}$ need to converge against its alignment with human evaluation on the UCF-101 dataset. JEDi, the feature space of a V-JEPA model (\vjepaft) in combination with a Maximum Mean Discrepancy (MMD) metric, is a vastly more efficient framework for evaluating distributions of generated videos than conventional methods. The current standard, FVD (FD+I3D), underperforms in terms of both sample efficiency and alignment with human evaluation.}
\end{figure}
\input{sections/introduction}

\vspace{-.25cm}
\section{Background and Notations}
\hypertarget{Background-and-Notations}{}
\input{sections/background}

%\vspace{-.3cm}
\section{Examining FVD: Feature Spaces and the Gaussianity Assumption}
\hypertarget{Examining-FVD}{}
\input{sections/fvd_linear_analysis}

%\vspace{-.3cm}
\section{The Dual Challenge of Convergence: High-Dimensional Feature Spaces and Limited Samples}
\hypertarget{Dual-Challenges}{}
\input{sections/convergence_discussion}

%\vspace{-.3cm}
\section{Metric Distance Analysis: Noise, Generative Models, and Human Study\label{sec:noise_and_human_study}}
\hypertarget{Metric-Analyses}{}
\input{sections/noise_and_human_study}

%\vspace{-.3cm}
\section{Conclusion}\label{sec:conclusion}
\input{sections/conclusion}

\section*{Acknowledgments}
\addcontentsline{toc}{section}{Acknowledgement}
We are deeply grateful to \emph{Mufan (Bill) Li} for his expert consultation on statistical proofs during the formative stages of this project. Also, we thank \emph{Songwei Ge} for his valuable assistance in configuring the evaluation code for this project. Special thanks to \emph{Benno Krojer} for taking the time to review and provide valuable feedback on our work. Moreover, we extend our gratitude to all individuals who generously participated in our survey study, sharing their valuable insights and time. We thank \emph{NSERC}, \emph{CIFAR} and the \emph{Samsung SAIT AI Lab Montreal} for supporting this work.
%Bibliography
\bibliographystyle{bib_style}
\bibliography{references}
\clearpage
\appendix
\input{sections/appendix}

\end{document}

%% file: math_commands_old.tex
\usepackage{amsmath,amsfonts,bm,amssymb}
\usepackage{graphicx}
\usepackage{subfig}
\usepackage{array,booktabs}
\newcolumntype{M}[1]{>{\centering\arraybackslash}m{#1}} % w/ horizontal centering
\usepackage{subcaption} % for 'subfigure' environment

\def\eqref#1{equation~\ref{#1}}
\def\1{\bm{1}}

\newcommand{\mw}{\text{MW}_2}

\def\rx{{\textnormal{x}}}
\def\ry{{\textnormal{y}}}
\def\rz{{\textnormal{z}}}

\def\rmX{{\mathbf{X}}}

\def\vb{{\bm{b}}}

\def\mA{{\bm{A}}}

\DeclareMathAlphabet{\mathsfit}{\encodingdefault}{\sfdefault}{m}{sl}
\SetMathAlphabet{\mathsfit}{bold}{\encodingdefault}{\sfdefault}{bx}{n}
\newcommand{\tens}[1]{\bm{\mathsfit{#1}}}

\def\tX{{\tens{X}}}
\def\tY{{\tens{Y}}}

\def\gE{{\mathcal{E}}}
\def\gF{{\mathcal{F}}}

\def\gN{{\mathcal{N}}}

\def\sR{{\mathbb{R}}}

\def\sX{{\mathbb{X}}}

\newcommand{\E}{\mathbb{E}}

\newcommand{\R}{\mathbb{R}}

\newcommand{\vjepapt}{$\text{V-JEPA}_{\text{PT}}$}
\newcommand{\vjepaft}{$\text{V-JEPA}_{\text{SSv2}}$}
\newcommand{\videomaept}{$\text{VideoMAE}_{\text{PT}}$}
\newcommand{\videomaeft}{$\text{VideoMAE}_{\text{SSv2}}$}
\newcommand{\ourmetric}{JEDi}

\DeclareMathOperator{\Tr}{Tr}

\usepackage{arydshln}
\usepackage{array,multirow,graphicx}

\usepackage[inline]{enumitem}
\usepackage{diagbox}

%% file: sections/introduction.tex
Video generation research has experienced a significant surge recently, yielding cutting-edge models that produce high-quality videos~\citep{liu2024sora, blattmann2023stable, zeng2023pixeldance, he2023latent}. However, evaluating their generation quality poses a substantial challenge.

A video generation model, must not only produce high-quality images, but also have high temporal consistency and produce diverse videos with diverse features. For instance, at the object level, it is undesirable to have exclusively cars of specific brands or colors when generating automobile videos; at the motion level, it is undesirable to observe the same type of motion repeatedly when generating human action videos. Thus, an ideal video generation metric must tackle many aspects in order to be a reliable tool for evaluating generative video models.

%In generative tasks, models may occasionally produce samples with certain properties because they find them “easy” to produce. %Understanding the distribution of data generation is essential for training models that generalize effectively. 
%For instance, at the object level, it is undesirable to have exclusively cars of specific brands or colors when generating automobile videos; at the motion level, it is undesirable to observe the same type of motion repeatedly when generating human action videos. Understanding the distribution of data generation is essential for training models that generalize effectively. Thus, it is important to have a reliable tool for evaluating generation distributions.
 
%When evaluating the performance of a generative model, two main aspects are considered: \emph{the quality of individual outputs} and \emph{the overall quality of the entire generation distribution}. The \emph{quality}, \emph{realism}, and \emph{accuracy} of each generated sample are assessed when 
%evaluating \emph{the quality of individual outputs}. The \emph{diversity}, \emph{coverage}, and \emph{similarity} of the overall generation distribution are evaluated when assessing \emph{the generation distribution quality}.

Researchers have created a range of evaluation metrics and tools to assess the quality of individual outputs from video generation models. Many of the older video generation metrics are derived from image quality assessments: LPIPS, MSE, SSIM, PSNR. These metrics fail to quantify the temporal consistency between frames~\citep{zhang2018lpips, hore2010psnr_ssim}. Recent video distance metrics focus on assessing both temporal and spatial qualities, as well as diversity, by measuring distances in distribution with respect to the real videos. The most popular such metric for video is the Fr\'echet Video Distance (FVD) ~\citep{unterthiner2019fvd}. An analog exists for measuring image quality (but not temporal consistency): the Fr\'echet Inception Distance (FID)~\citep{heusel2017fid}. These metrics have emerged as a leading tool for assessing the quality of video and image generation models. 

FID, primarily a image metric, is also used in video generation to compare key-frames. It computes the Fr\'echet distances of the frame features from Inception v3 ~\citep{szegedy2014inception, ioffe2015inception_with_bn} trained on ImageNet. Building upon the principles of FID metrics, FVD evaluates the FD between the generated and data distributions in the Inflated 3D ConvNet (I3D)'s feature space~\citep{carreira2018i3d}, which is trained on the Kinetics dataset~\citep{kay2017kinetics}.  FVD's use of features extracted from 3D-ConvNet allows it to capture a more comprehensive range of visual and temporal information compared to FID.

Recent studies have highlighted limitations in the reliability of the FVD measure. Specifically, \citet{brooks2022generating_long_video} demonstrated that FVD is not effective in capturing long-term realism and is more suitable for comparing generation model variants of the same architecture. Moreover, ~\citet{skorokhodov2021styleganv} showed that FVD overlooks motion collapse and is biased towards image quality, rather than video quality. Additionally, they pointed out that FVD is excessively sensitive to minor implementation details, such as the specific image storage formats used (e.g., JPEG compression levels or file encoding), which can lead to inconsistent and non-comparable results across different studies. 

A comprehensive study on FVD was conducted by~\citet{ge2024content}, which compares Fr\'echet distances of features extracted by the I3D network~\citep{carreira2018i3d} and VideoMAE network~\citep{wang2023videomae}. The study shows that the FVD prioritizes per-frame quality over temporal consistency when using I3D features, which they refer to as content-bias. Further, they suggest that using features from self-supervised models trained on content-debiased data can effectively mitigate this bias in FVD. Our methodology draws inspiration from previous analysis that highlights shortcomings with FID~\citep{borji2021image_metric_analysis, kynkäänniemi2023role, soloveitchik2022cfid, sajjadi2018assessing}.

\begin{comment}
Furthermore, the concept of FVD metrics is derived from the image quality metric FID, which has been extensively analyzed in numerous studies for its performance~\citep{borji2021image_metric_analysis, kynkäänniemi2023role, soloveitchik2022cfid, sajjadi2018assessing}. Research on FVD is more limited, emphasizing the need for further investigation in this area. By drawing on insights from prior FID research, we can inform and enhance our studies on FVD, ultimately advancing our understanding of video quality assessment.
\end{comment}

A separate and distinct method of evaluating videos is on the sample level, rather than the distributional level. For example, \citep{huang2023vbench} recently developed VBench, a comprehensive video benchmark that analyzes the evaluation of individual generated outputs on subject consistency, background consistency, temporal flickering, motion smoothness, aesthetic quality, among others. VBench addresses the challenge of evaluating both temporal and spatial consistency in video generation, but naturally fails at efficiently evaluating the generational capabilities of a model on a distributional level, or the robustness of a model in generating videos outside its benchmark distribution.

In our proposed framework, \textbf{JEDi}, we address many of the problems affecting existing evaluation strategies:
% [For ICLR] \vspace{-0.3cm}
\begin{enumerate}[leftmargin=*,nolistsep]%,noitemsep,nolistsep]
    \setlength{\itemsep}{1pt} % Adjust this value for desired spacing
    \item \ourmetric~employs a Maximum Mean Discrepancy (MMD) metric with a polynomial kernel, eliminating the need for parametric assumptions about the underlying video distribution, unlike FVD which relies on the Gaussianity assumption to make its metric feasible.
    \item \ourmetric~significantly reduces the number of samples needed to make an accurate estimate by using an MMD metric in a V-JEPA feature space, enabling reliable use in smaller datasets that do not meet the requirement when using FVD.
    \item JEDi leverages the robust representations of a V-JEPA model, which are found to be more aligned with human evaluations compared to FVD.
\end{enumerate}

%% file: sections/background.tex
\subsection{Video Feature Representation}
\textbf{Inflated 3D ConvNet:} The Inflated 3D ConvNet (I3D)~\citep{carreira2018i3d} is a convolutional neural network model based on the pre-trained Inception-v1. It extends the 2D convolutional filters to 3D by replicating them along the temporal dimension. I3D, pre-trained on Kinetics, has demonstrated excellent classification performance on UCF-101~\citep{soomoro2012ucf101}, HMDB-51~\citep{kuehne2011hmdb}, and Kinetics datasets~\citep{kay2017kinetics}, proving to be a valuable network for video recognition tasks.

The original FVD work by \citet{unterthiner2019fvd} explores the use of I3D features trained on the Kinetics datasets. They analyze the features from the logits layer, as well as the features from the last pooling layer trained on the Kinetics-400 and Kinetics-600 datasets. Their experiments suggest that the features from the logits layer trained on the Kinetics-400 dataset are the most suitable for the FVD metric.

\textbf{Video Masked Autoencoder:} The Video Masked Autoencoder (VideoMAE-v2)~\citep{wang2023videomae} is a self-supervised pre-training method that leverages a vision transformer (ViT) backbone~\citep{dosovitskiy2020vit} to learn efficient video representations. According to ~\citeauthor{ge2024content}, the giant-VideoMAE-v2 model, pretrained on a diverse set of unlabeled datasets and fine-tuned on Something-something-v2~\citep{goyal2017something} with a masked autoencoder objective, effectively captures both spatial and temporal distortions in its encoded feature space. We leverage two variants of the VideoMAE-v2 model in our study: (1) \emph{\videomaept}: the self-supervised pre-trained giant VideoMAE-v2 model and (2) \emph{\videomaeft}: the fine-tuned giant VideoMAE-v2 model.

\textbf{Video Joint Embedding Predictive Architecture:} Video Joint Embedding Predictive Architecture (V-JEPA)~\citep{bardes2024revisiting} is a self-supervised training paradigm that learns by predicting missing or masked parts of a video in an abstract representation space. V-JEPA excels in ``frozen evaluations'', where its encoder and predictor are pre-trained through self-supervised learning and then left unchanged. For new tasks, only a small, lightweight layer or network is trained on top of the pre-trained components, enabling quick and efficient adaptation to new environments. In this study, we employ both (1) the pre-trained variant of the model trained with the self-supervised objective, \emph{\vjepapt}, as well as (2) a version that was fine-tuned on Something-something-v2~\citep{goyal2017something} with an attentive classification probe such that its pre-logit features could be used for distributional analysis metrics, \emph{\vjepaft}. 

\subsection{Fr\'echet Distance and Fr\'echet Video Distance}

Fr\'echet Distance (FD), also known as 2-Wasserstein distance ($W_2$), is a way of measuring how similar two distributions are~\citep{frechet1957frechet_distance,dowson1982frechet_gaussian, zilly2020frechet_generation_gap}. The Fr\'echet distance between two distributions \(P\) and \(Q\) is defined as the minimum distance between all pairs of random variables \(\rx\) and \(\ry\) from the distributions. Assuming \(P\) and \(Q\) are multivariate Gaussian distributions, it can be expressed as:
\begin{comment}
\begin{equation}
\small
    D_{\text{Fr\'echet}}^2(P,Q) = \left( \inf_{\gamma \in \Gamma (\mu, \nu)} \int_{\R^n \times \R^n} \|x-y\|^2 \, \mathrm{d} \gamma (x, y) \right)^{\frac{1}{2}},\label{eq:squared_FD_distance}
\end{equation}
where $\Gamma(\mu, \nu)$ is the set of all measures on $\R^n \times \R^n$ with marginal distributions $\mu$ and $\nu$.

Unfortunately, this equation is intractable and hard to solve. %The general case of calculating the Fréchet distance between two distributions is known to be difficult. 
However, by assuming that $P$ and $Q$ are multivariate Gaussian distribution, the distance becomes tractable and we obtain the closed-form solution ~\citet{dowson1982frechet_gaussian}:
%~\citet{dowson1982frechet_gaussian} provides a closed-form solution for the scenario where the two distributions are multivariate Gaussian:
\end{comment}
\begin{equation}
    D_{\text{Fr\'echet}}^2(P,Q) = (\mu_P-\mu_Q)^2 + \Tr(\Sigma_P+\Sigma_Q-2(\Sigma_P\Sigma_Q)^{\frac{1}{2}})\label{eq:squared_FD_gaussian_distance}
\end{equation}

where $\mu_P$ and $\mu_Q$ are the means, while $\Sigma_P$ and $\Sigma_Q$ are the covariance matrices of the two Gaussian distributions. Without making this assumption, the Fr\'echet Distance is intractable and becomes much more arduous to obtain.

The Fr\'echet Inception Distance (FID) and Fr\'echet Video Distance (FVD) correspond to the above equations, but the distance is applied in the space of InceptionV3 and I3D network features, respectively, instead of directly in raw image space in order to obtain more meanful distance that better align with human preferences ~\citep{szegedy2014inception, unterthiner2019fvd}. 

%The Fr\'echet Video Distance (FVD) is a metric used to assess the quality and diversity of generated videos in comparison to the original dataset. FVD calculates the Fr\'echet Distance between these two distributions under the multivariate Gaussian assumption~\citep{unterthiner2019fvd}. 

%In the upcoming sections, we will denote the generation distribution as $ G $ and the real distribution as $ R $:
%\begin{equation}
%    \text{FVD} = (\mu_G-\mu_R)^2 + \Tr(\Sigma_G+\Sigma_R-2(\Sigma_G\Sigma_R)^{\frac{1}{2}})\label{eq:fvd}
%\end{equation}

\subsection{Other Distribution Distance Metrics}
\hypertarget{other-distribution-metrics}{}
This study also explores the application of alternative statistical methods to compute probability distribution distances in video feature spaces, including Mixture Wasserstein ($\mw$), Energy Statistics, and kernel-based methods such as Maximum Mean Discrepancy (MMD). 

The detailed backgrounds of these metrics are provided in Appendix ~\ref{sec:distribution_distance_overview}.

% \subsection{Notation and Symbols}
% In this paper, we reuse mathematical notations throughout. For quick reference, this glossary defines the key symbols and notations used.

% TODO

%% file: sections/fvd_linear_analysis.tex
The Fr\'echet distance (FD) measures the difference between means and covariances. This can offer insights into the first two moments of the distributions, but fails to do so with respect to higher-order moments (e.g., skewness, kurtosis) that arise when either the real or generated data distribution is non-Gaussian. According to \citet{jayasumana2024rethinkingFID}, the reliance on Gaussianity assumptions in FID research can lead to substantial inaccuracies when the underlying image distribution does not come from such a distribution. This part of the study focuses on the video feature spaces, investigating the accuracy of Gaussian assumptions and considering the consequences of the Fréchet Video Distance (FVD) when those assumptions are not met.

%(TODO: Notably, their toy experiment (Table 2) demonstrates that data samples generated from diverse mixtures of Gaussians exhibit significant deviations from multivariate Gaussian samples, yet the Fr\'echet Distance remains invariant at 0, revealing intriguing inconsistencies.) 
%Similarly, video representations may not adhere to multivariate Gaussian distributions in their feature spaces, resulting in flawed measurements. This part of the study focuses on the video feature spaces, investigating the accuracy of Gaussian assumptions and considering the consequences of Fréchet Video Distance (FVD) metrics when these assumptions are not met. These analyses further uncover the underlying interpretations and constraints of FVD in situations where Gaussian assumptions do not hold true.
\begin{comment}    
Similar to images, the feature space of videos may not adhere to the multivariate Gaussian distribution, resulting in flawed measurements. This part of the study focuses on the video feature spaces, investigating the accuracy of Gaussian assumptions and considering the consequences of the Fréchet Video Distance (FVD) when those assumptions are not met. We show below that \emph{video representations are not Gaussian}, and \emph{this violation of Gaussianity becomes more severe as duration length increases}, thus making this assumption especially problematic for video evaluation. 
\end{comment}

Using each of the I3D, VideoMAE, and V-JEPA networks under comparison, we extract 48,501 features from 11 distinct video datasets, with each feature representing a 32-frame clip. Specifically, we extract a maximum of 5000 features from the training set of each dataset, which include: Anime-Run-v2~\citep{li2022animerun}, BAIR~\citep{ebert2017bair}, BDD100k~\citep{yu2020bdd100k}, DAVIS~\citep{ponttuset2017davis}, Fashion Modeling~\citep{zablotskaia2019ubcfashion}, HMDB-51~\citep{kuehne2011hmdb}, How2Sign~\citep{duarte2021how2sign}, KITTI~\citep{geiger2013kitti}, Something-Something-v2~\citep{goyal2017something}, Sky Scene~\citep{xiong2018skyscene}, and UCF-101~\citep{soomoro2012ucf101}.

The notion that I3D video features do not follow multivariate Gaussian distributions was investigated using the widely-accepted Mardia's Skewness~\citep{mardia_skewness}, Mardia's Kurtosis~\citep{mardia_skewness}, and Henze-Zirkler~\citep{hz_test} normality tests, following~\citep{jayasumana2024rethinkingFID}. The null hypothesis that the I3D features were drawn from a multivariate Gaussian distribution was strongly rejected ($p=0$) for each of the datasets and normality tests.

\begin{figure}[h]%
    \centering
    \setlength\tabcolsep{3pt} % default: 6pt
    \begin{tabular}{c M{0.17\linewidth} M{0.17\linewidth} M{0.17\linewidth} M{0.17\linewidth} M{0.17\linewidth}}
     & \textbf{I3D}  & \textbf{\videomaept} & \textbf{\videomaeft} & \textbf{\vjepapt} & \textbf{\vjepaft}\\
     LDA & \includegraphics[width=\hsize]{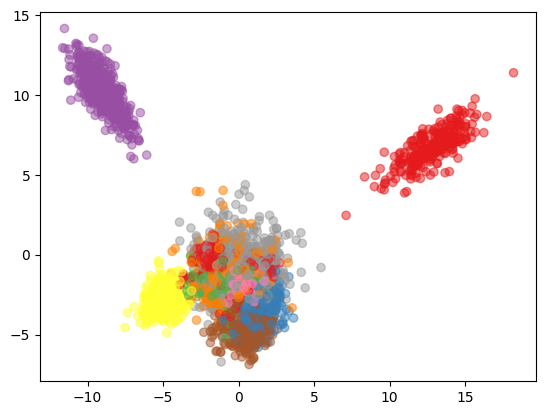}
      & \includegraphics[width=\hsize]{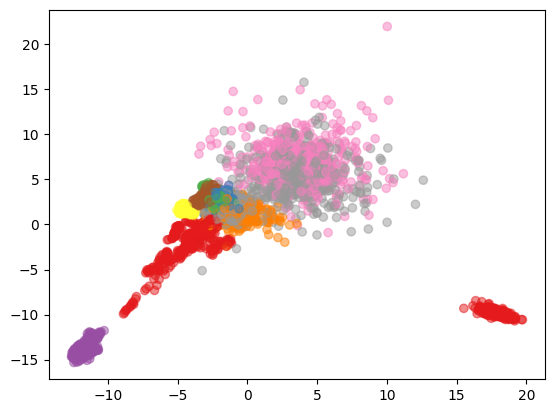}
      & \includegraphics[width=\hsize]{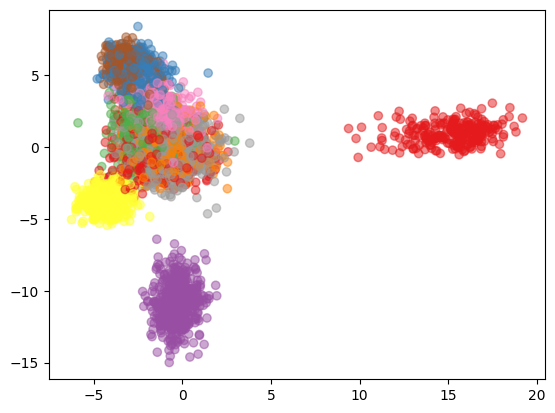}
      & \includegraphics[width=\hsize]{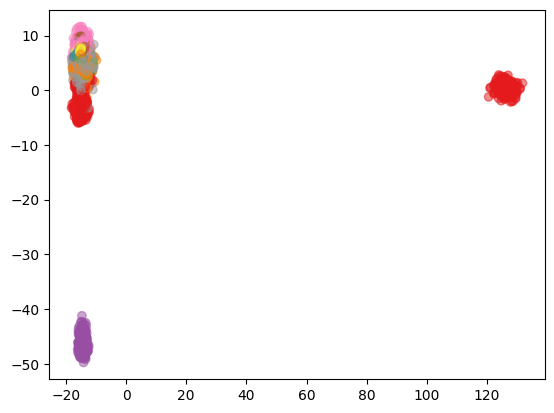}
      & \includegraphics[width=\hsize]{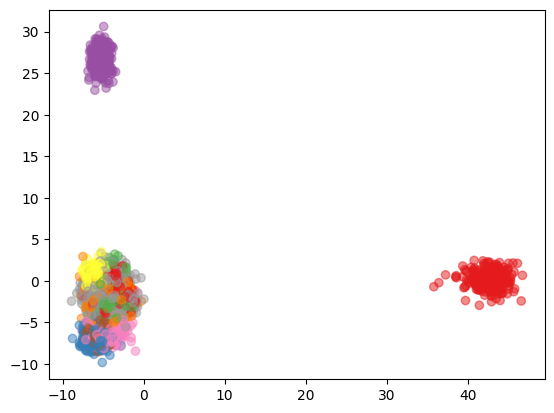}\\
    PCA & \includegraphics[width=\hsize]{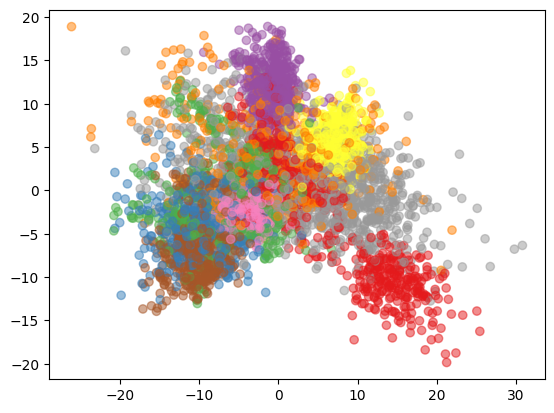}  
      & \includegraphics[width=\hsize]{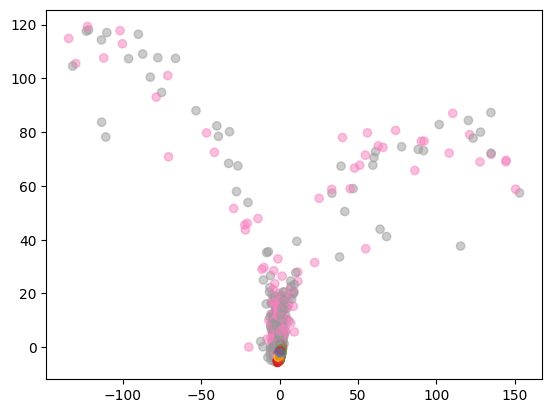}
      & \includegraphics[width=\hsize]{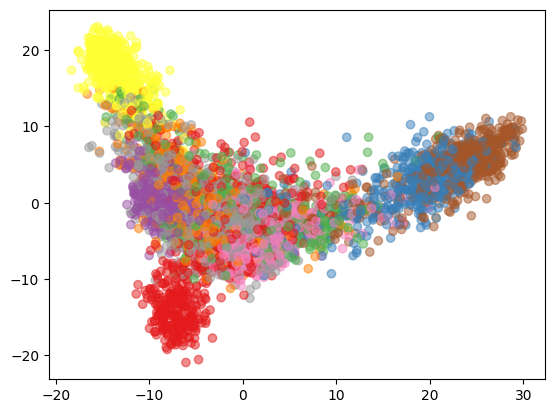}
      & \includegraphics[width=\hsize]{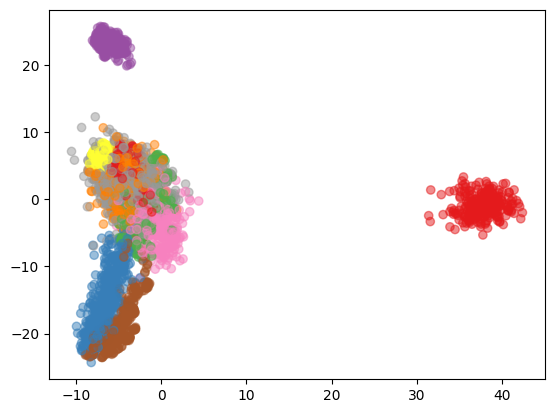}
      & \includegraphics[width=\hsize]{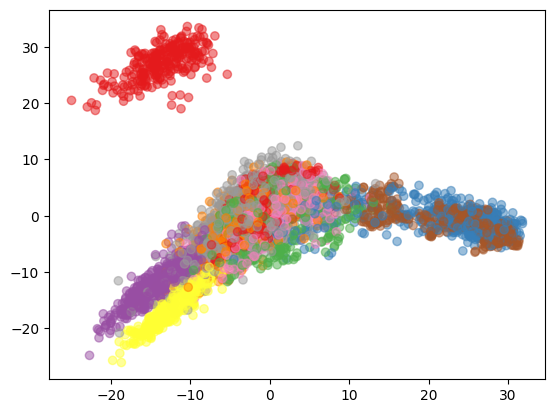}
      
      \\
      & \multicolumn{5}{c}{\includegraphics[width=0.87\hsize]{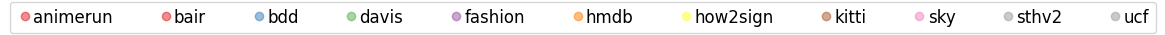}}\\
      \end{tabular}
    \caption{The dimensionally reduced video features of the 11 datasets using LDA and PCA indicate that the video features are non-Gaussian in the combined dataset space. While individual dataset clusters may appear Gaussian in these plots, the low explained variance ratios (0.134-0.231) of the PCA-reduced spaces suggest that 2D projections in these plots may not capture the complexity of higher-dimensional feature distributions within individual datasets. Figures~\ref{fig:pca_individual} and~\ref{fig:pca_individual_vjepa_pt} contain dataset-specific LDA and PCA plots, which reveal non-Gaussian characteristics within the datasets.}%
\label{fig:i3d_pca_lda_all}
\end{figure}

We then normalize the aggregated training-set features and fit a Principal Component Analysis (PCA) model and a Linear Discriminant Analysis (LDA) model using dataset labels as classes. We apply the same pipelines to transform 5,256 I3D features (up to 500 samples from each of the eleven datasets' \emph{testing sets}) into lower-dimensional spaces for visualizations. As demonstrated in Appendix~\ref{sec:linear_transformation_multivariate_gaussian}, applying PCA or LDA transformations to Gaussian-distributed data preserves their Gaussian properties. Figure~\ref{fig:i3d_pca_lda_all} shows that the I3D features don't follow a single multivariate Gaussian distribution; rather, they cluster by dataset.

\begin{comment}
In order to test the assumption of multivariate Gaussian distribution, we utilize PCA and LDA to decrease the dimensionality of the I3D features from 400 to 2. As demonstrated in Appendix~\ref{sec:linear_transformation_multivariate_gaussian}, applying PCA or LDA transformations to Gaussian-distributed data preserves their Gaussian properties. Our objective is to visually assess whether the reduced-dimension features exhibit a multivariate Gaussian distribution in the lower-dimensional space.
\end{comment}

% \begin{figure}[h]%
%     \centering
%     \subfloat[PCA]{{\includegraphics[width=0.26\linewidth]{figs/linear_analysis/i3d_all_pca_2d.png} }}%
%     \subfloat[LDA]{{\includegraphics[width=0.26\linewidth]{figs/linear_analysis/i3d_all_lda_2d.png} }}%
%     \subfloat[PCA's explained variance ratio]{{\includegraphics[width=0.30\linewidth]{figs/linear_analysis/i3d_explained_variance.png} }}%
%     \caption{The first two figures display the dimensionally reduced I3D features of the eight datasets using PCA and LDA. The last diagram shows how the explained variance ratio changes as the number of principal components varies. This represents the amount of the dataset's variance that is captured by each principal component. The total explained variance ratio for the first 2 principal components is 0.5390.}%
% \label{fig:i3d_pca_lda_all}
% \end{figure}
\vspace{-0.3cm}

Moreover, we conduct a PCA on individual datasets and apply a LDA to the HMDB-51 and UCF-101 datasets, incorporating classification labels and varying frame counts per clip. Our analysis shows that I3D clip features deviate from Gaussian distributions within each dataset. More notably, we observe a \emph{positive correlation between clip duration and increase in FVD between the train and test sets from the same dataset}, suggesting that higher-order moments may be essential for accurate characterization.

We replicate our experiments in the VideoMAE and V-JEPA feature spaces using the same datasets. Our results in these feature spaces mirror our findings in the I3D feature space. Additional details on the setup and results are provided in Appendix~\ref{sec:non_gaussian}. \hypertarget{vmae-jpepa-extra-results}{}

\begin{comment}
    Notably, longer clips exhibit a more pronounced departure from Gaussianity, with their feature distributions displaying a greater degree of non-Gaussian characteristics in the 2D space. This phenomenon may be attributed to the I3D network's pre-training on a classification task, which emphasizes discriminative features that are more prominent in longer clips, thereby amplifying the deviation from Gaussianity.

    PCA/LDA visualizations reveal that the duration of video clips significantly influences the shape of their representation distributions. Notably, longer clips' features exhibit more complex distribution shapes, suggesting that higher-order moments may be essential for accurately characterizing these distributions. These findings are further investigated in Appendix~\ref{sec:non_gaussian}, where we observe a positive correlation between clip duration and the increase in FVDs between the training and testing sets. The rest of this section will delve into how the scarcity of data relates to the singularity of multivariate covariance matrices and how limited data can affect the stability of these matrices.
\end{comment}

%% file: sections/convergence_discussion.tex
The following section addresses two pivotal challenges in evaluating video distribution distances:
\begin{enumerate}[leftmargin=*]%,noitemsep,nolistsep]
    \setlength{\itemsep}{1pt} % Adjust this value for desired spacing
    \item \emph{The Dimensionality Problem} (Section \ref{sec:curse_of_dimensionality}): We examine the limitations of metrics relying on distribution assumptions (e.g., Fréchet distance, Mixture Wasserstein distance), highlighting the adverse impact of high dimensionality.
    \item \emph{Sample Efficiency and Convergence} (Section \ref{sec:convergence_and_data}): We discuss the sample efficiency issue affecting all metrics and the necessary sample size for trustworthy measurements.
\end{enumerate}

\subsection{Challenge \#1: The Curse of Dimensionality~\label{sec:curse_of_dimensionality}}
\subsubsection*{Impact of Data Dimension on Fr\'echet Distance Metric}
In the previous section, we have shown that the Fr\'echet Distance (FD) can be used as a metric for comparing the discrepancy between the first two moments of two distributions. Consequently, the accuracy of the mean and covariance estimators is crucial for ensuring the validity of FD as a metric. In the following part, we will explore the impact of data dimensionality and sample size on the quality and precision of these estimators, and examine how this affects the reliability of distribution distance metrics.

The rank of the empirical covariance matrix (\(\hat{\Sigma}\)) is tied to the number of samples (\(n\)) and the dimension (\(k\)). Given a matrix \(\rmX\) containing \(n\) observations, where each column vector represents a $k$-dimensional multivariate sample, the empirical covariance matrix can serve as a reliable estimator for the true covariance matrix. The empirical covariance matrix is calculated using the formula: \(\hat{\Sigma} = \frac{1}{n}\sum^n_{i=1} \big(\rmX_i-\bar{\rmX}\big)\big(\rmX_i-\bar{\rmX}\big)^\intercal\), and it consistently converges to the true $ \Sigma $ at a rate of \(\frac{1}{\sqrt{n}}\).

It is crucial to recognize that when the number of samples is less than the number of variables (\(n<k\)), the covariance matrix becomes singular. 
A good covariance estimator requires a sample size that is sufficiently large, ideally at least several times greater than the data dimension. This is because estimating the covariance matrix involves estimating $k(k+1)/2$ parameters, which requires a sufficiently large number of samples to achieve accurate estimates~\citep{bickel2008regularize, Marcenko1967distribution, wang_ecm, jonsson1982limit}. Unfortunately, the high-dimensional nature of I3D (400), VideoMAE (1408), and V-JEPA (1280) representation spaces exacerbates this issue.

Furthermore, optimal transport methods with complex distributional assumptions require more samples yet. The Mixture Wasserstein ($MW_2$) experiment described in Appendix~\ref{sec:toy_experiment} highlights significant computational and practical limitations of optimal-transport type metrics, making them impractical for this project. See Appendix~\ref{sec:toy_experiment} for further discussion. \hypertarget{toy-experiment}{}

\subsubsection*{Data Transformation: Dimensionality Reduction\label{sec:data_transformation}}

To address the challenges posed by the curse of dimensionality, dimension reduction techniques such as PCA and autoencoders~\citep{lecun1987thesis} can be applied. Our preliminary investigation from Appendix ~\ref{sec:toy_experiment} suggests that decreasing the representation dimension could enable the metric to converge with a smaller number of samples using metrics like Fréchet Distance. We test that hypothesis by training autoencoders in various feature spaces to reduce dimensionality. 

Our autoencoder architectures consist of simple multilayer perceptron networks, which compress feature dimensionality to either \(\frac{1}{6}\) of original size for I3D features or \(\frac{1}{8}\) of original size for VideoMAE and V-JEPA features. Additional information about the autoencoder training is available in Appendix~\ref{sec:ae_configuration}. \hypertarget{autoencoder}{}

Interestingly, dimension reduction significantly enhances the sample efficiency of the Fr\'echet Distance and energy statistic metrics in our experiments with Gaussian data (Figure~\ref{fig:toy_experiment}). However, its benefits are less pronounced for video features, resulting in only marginal improvements (Figures~\ref{fig:train_test_distances} and ~\ref{fig:train_test_convergence_rate}). Nonetheless, we retained autoencoder features in subsequent experiments to investigate other possible benefits.
%\vspace{-0.3cm}
\subsection{Challenge \#2: Sample Efficiency and Data Scarcity~\label{sec:convergence_and_data}}
Building on the insights from Sections \ref{sec:curse_of_dimensionality}, we recognize the critical importance of sufficient sampling for accurate estimation of metrics. This section delves into the relationship between sample size and convergence rate for each metric, exploring its impact across various feature spaces. Here, we define convergence rate as the rate at which the distance between training and testing set feature stabilizes as the number of video sample increases. It is a measure of sample efficiency.
 
Previous studies in the image and audio domains have shown that as the sample sizes $N$ decrease, Fr\'echet Distances increase~\citep{binkowski2021demystifyingmmdgans, gui2024adaptingfad, jayasumana2024rethinkingFID, chong2019unbiasfid}. This sensitivity to sample size is a common phenomenon among distributional distance metrics: As sample sizes increase, distance metrics become more reliable and accurate. However, while all metrics benefit from additional data, some converge to the true underlying distance more quickly than others.

\begin{figure}[ht]%
    \centering
    \setlength\tabcolsep{3pt} % default: 6pt
    \resizebox{\textwidth}{!}{\begin{tabular}{c M{0.3\linewidth} M{0.3\linewidth} M{0.3\linewidth}}
     & \textbf{I3D}  & \textbf{\videomaeft} & \textbf{\vjepaft} \\
    UCF-101 & \includegraphics[width=\hsize]{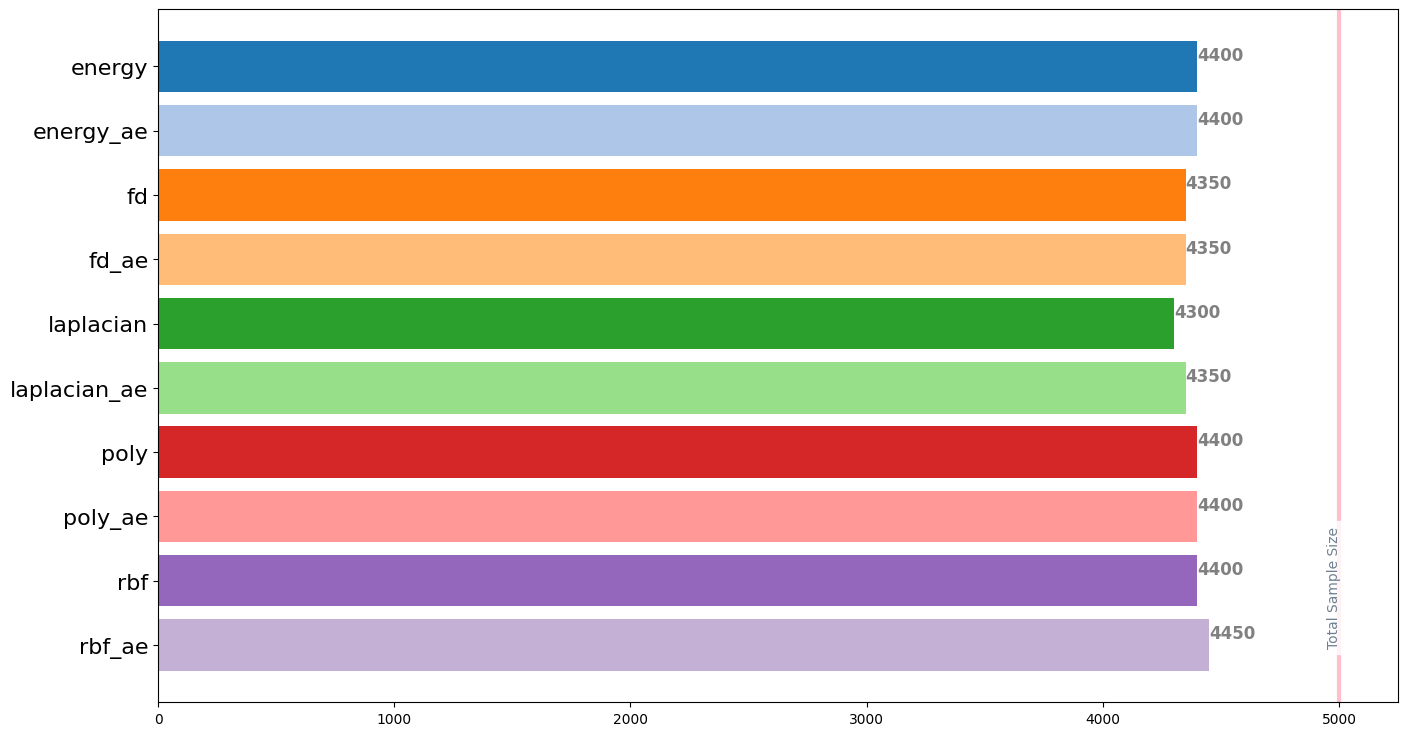}  
      & \includegraphics[width=\hsize]{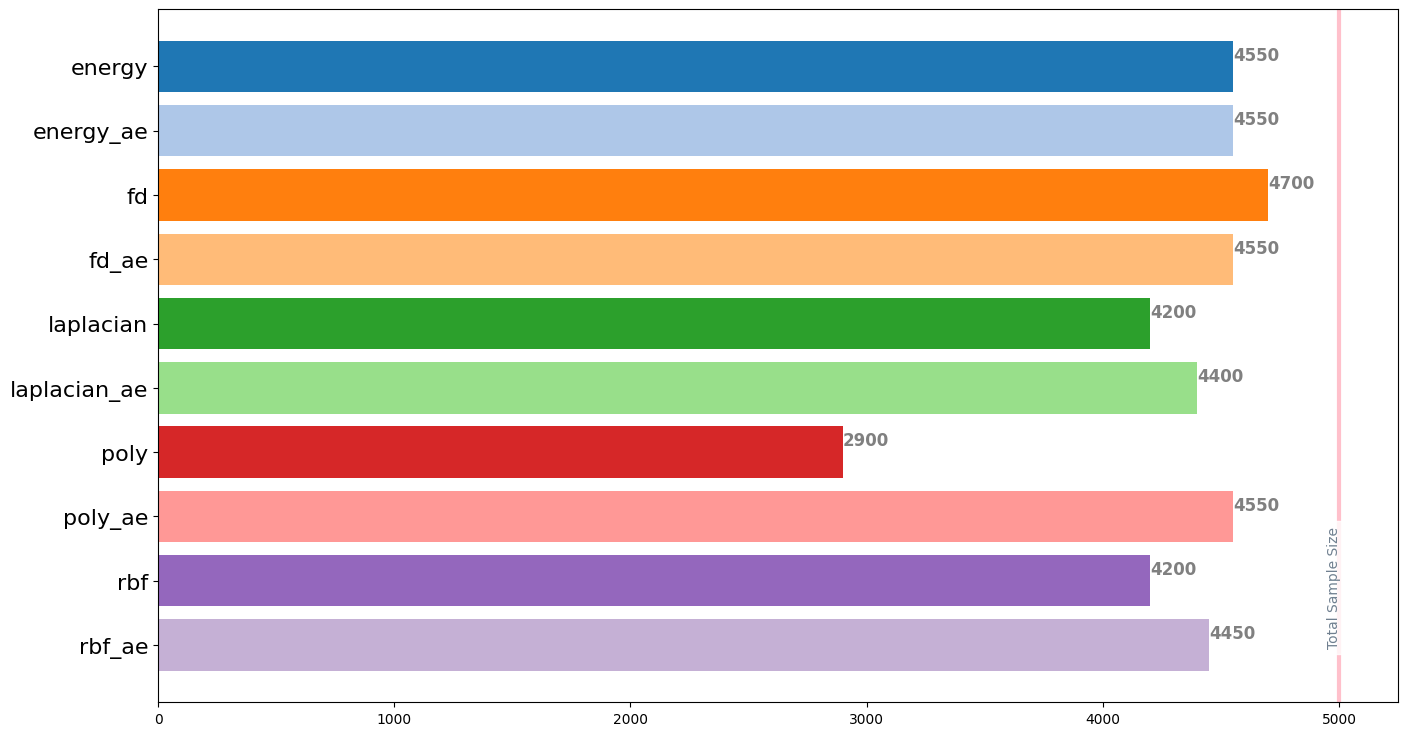}
      & \includegraphics[width=\hsize]{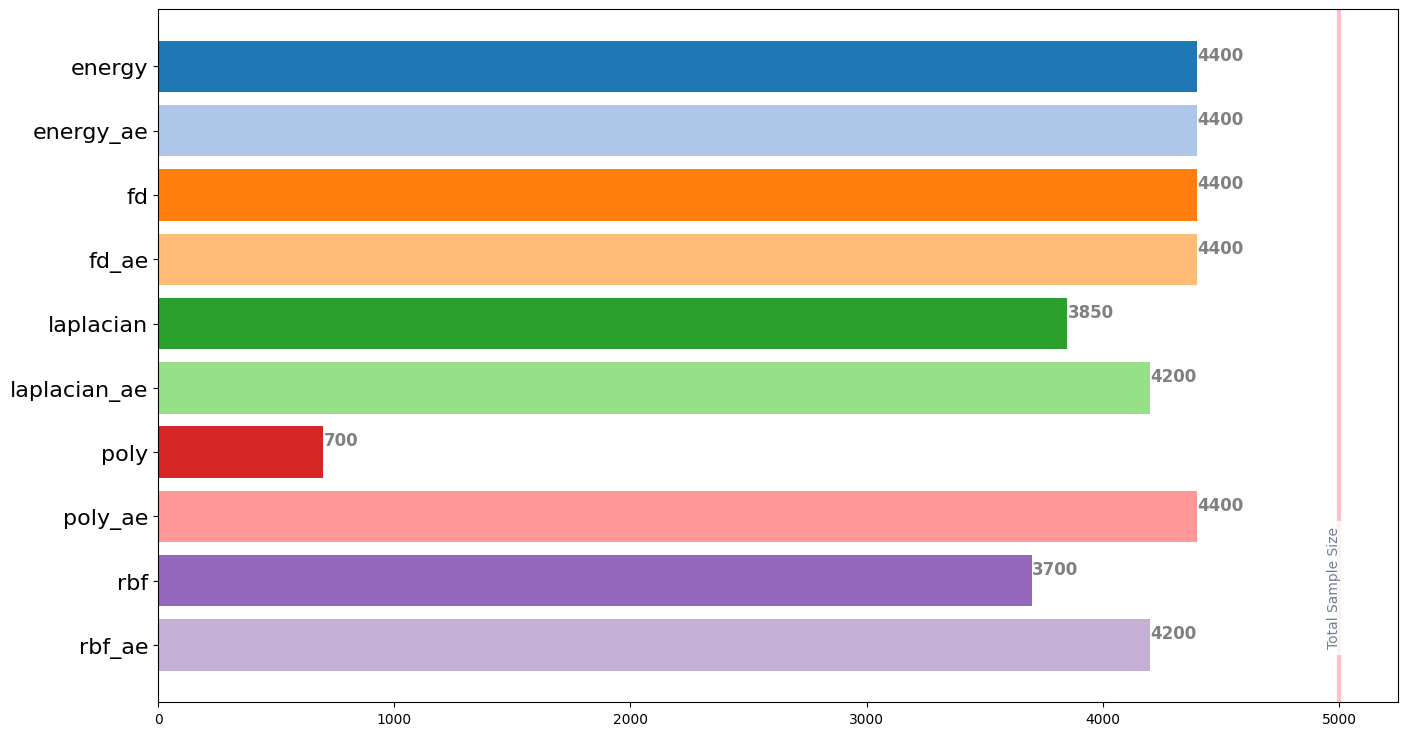}
      \\
      SSv2 & \includegraphics[width=\hsize]{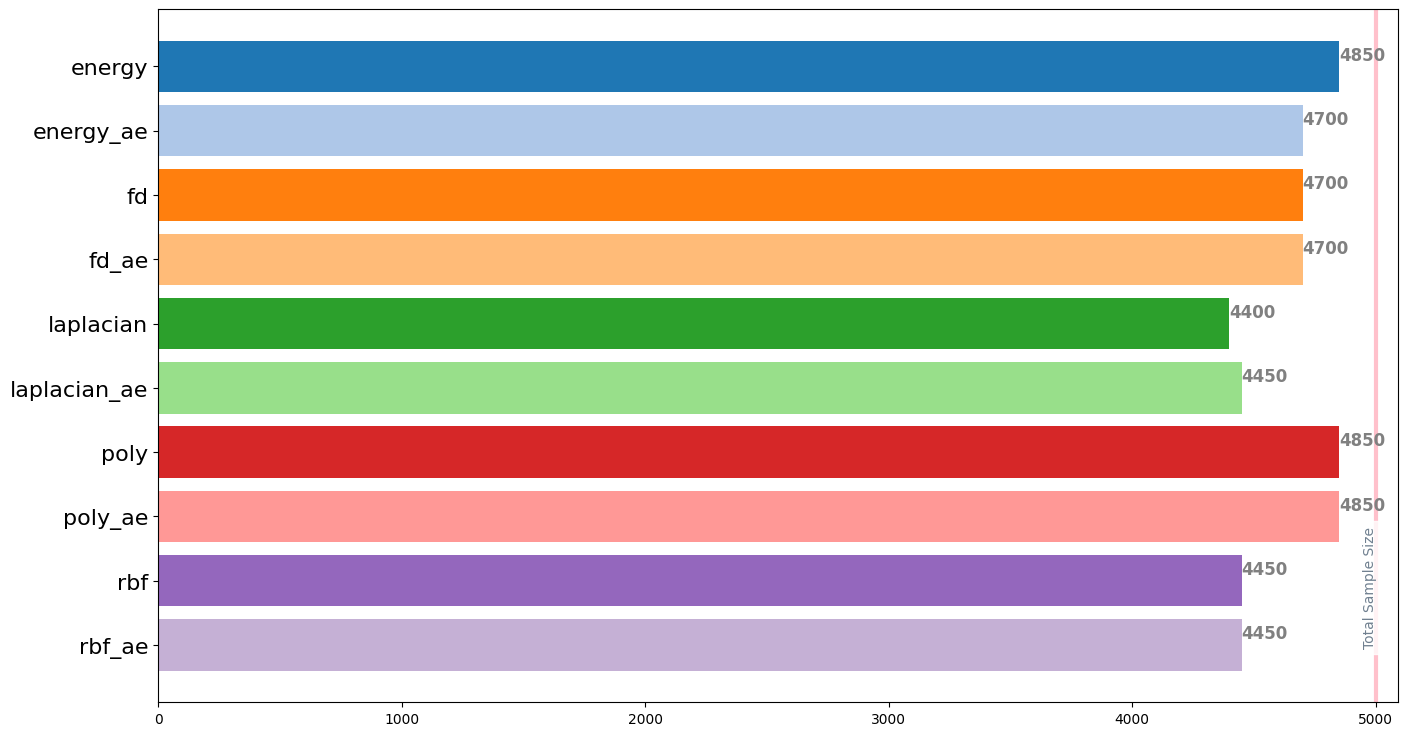}   
      & \includegraphics[width=\hsize]{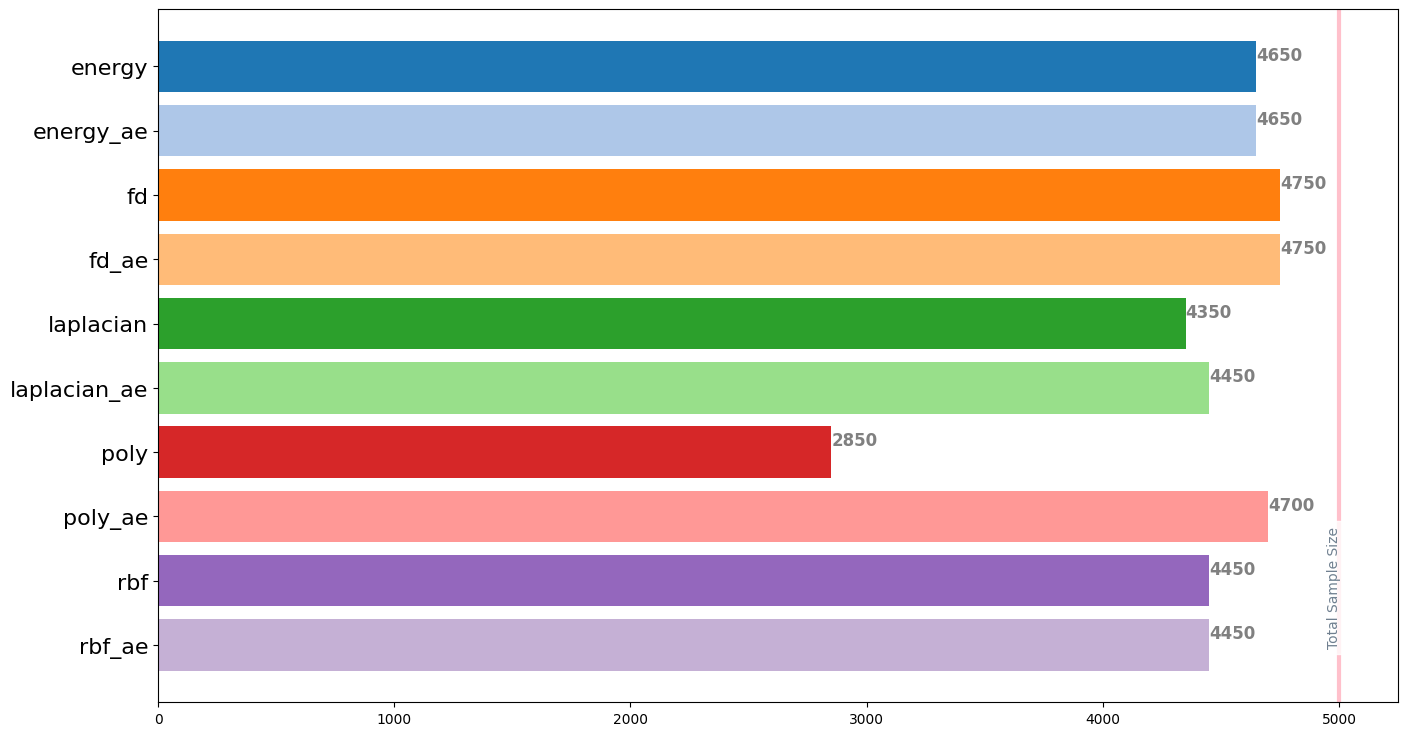}
      & \includegraphics[width=\hsize]{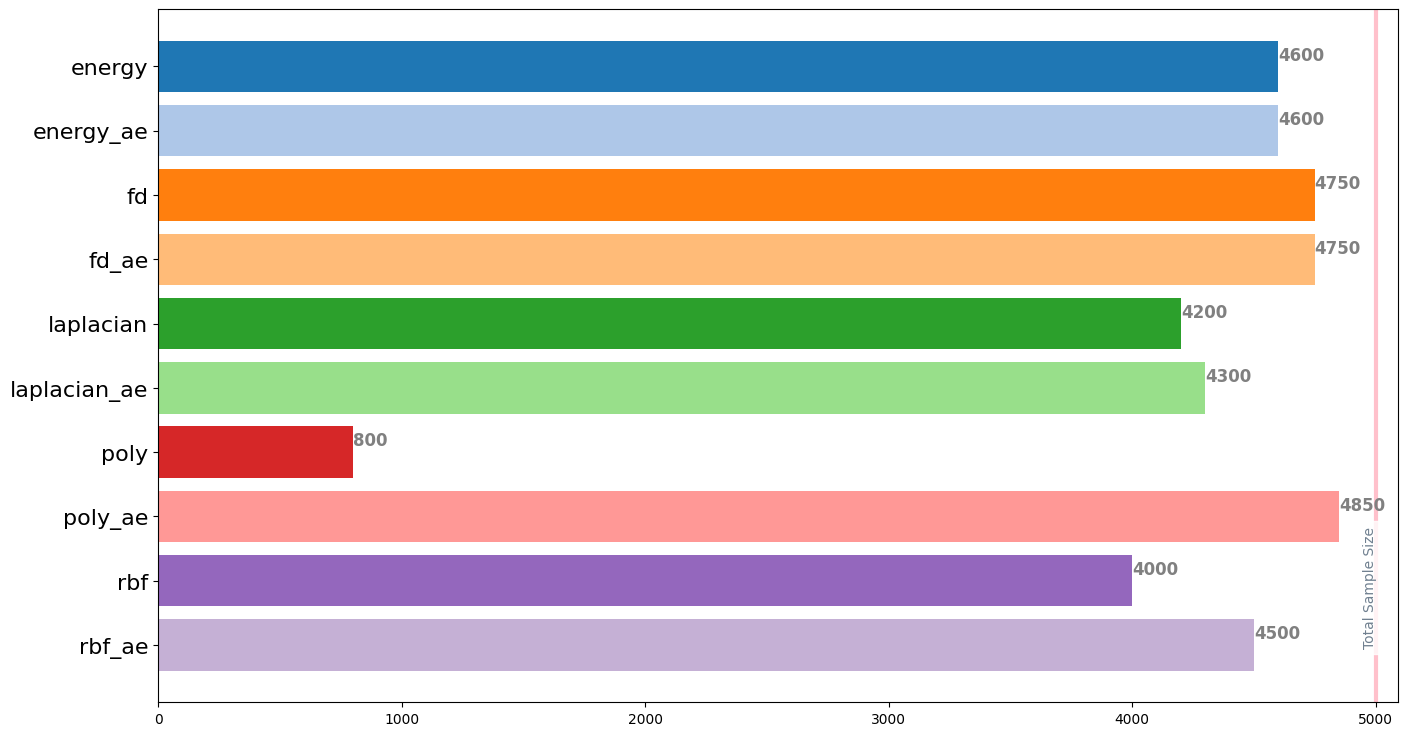}\\
      \end{tabular}}
    \caption{The number of samples needed to achieve a 5\% error margin of the distance measured from 5,000 samples using the training and testing sets of UCF-101. An ``\texttt{\_ae}" suffix indicates that the feature space has been compressed using an autoencoder. We assess the number of samples required for convergence at 100 sample intervals. Convergence at sample size $N$ is achieved if: (1) the average metric value from 5 repeated samplings of $N$ features falls within a 5\% error margin, and (2) all subsequent interval evaluations maintain an average metric value within the 5\% error margin. \videomaept~and \vjepapt~results are in the Appendix (Figure~\ref{fig:number_sample_convergence_others}). We find that Fr\'echet Distance (FD) converges slowest, while $\text{MMD}_{\text{POLY}}$ shows the highest sample efficiency.}%
\label{fig:number_sample_convergence}
\end{figure}

%Figures~\ref{fig:number_sample_convergence}, ~\ref{fig:number_sample_convergence_others}, ~\ref{fig:number_sample_convergence_all} and ~\ref{fig:bair_issue} shows 
We investigate the sample size required to achieve convergence within a 5\% error margin to average metric distance measured at 5,000 samples. Our analysis in Figure~\ref{fig:number_sample_convergence} spans two diverse datasets: UCF-101 (human action recognition) and Something-Something-v2 (SSv2, hand gesture recognition). An extended analysis on more datasets is found in Appendix~\ref{sec:sample_convergence_analysis_appendix}. \hypertarget{sample-efficiency}{}

\newpage
Our key findings include:
\begin{enumerate}[leftmargin=*]%,noitemsep,nolistsep]
    \setlength{\itemsep}{1pt} % Adjust this value for desired spacing
    \item 
    The sample sizes needed for convergence within the same feature space are similar across datasets. For instance, 4,350 samples on UCF-101 and 4,700 samples on SSv2 are required for FVD (I3D+FD) to converge, while 700 samples on UCF-101 and 800 samples on SSv2 are required for JEDi ($\text{MMD}_{\text{POLY}}$+\vjepaft) to converge.
    \item 
    \(\text{MMD}_{\text{POLY}}\) demonstrates the quickest convergence rate across non-I3D feature spaces. For example, it requires only 1,500 and 700 samples to reach a steady metric value on UCF-101 in \vjepapt~and \vjepaft~feature spaces, respectively.
    \item Fr\'echet Distance has the worst sample efficiency among metrics, while I3D features exhibit the worst sample efficiency across feature spaces.
    %\emph{Fr\'echet and energy distances demonstrate the lowest sample efficiency among evaluated metrics.} They require over 4,000 samples to converge on either of the two datasets when used within any of the three feature spaces.
\end{enumerate}

%Moreover, Figures \ref{fig:train_test_distances}, \ref{fig:train_test_convergence_rate}, and \ref{fig:train_test_convergence_rate_other} illustrate the convergence of distance estimates on the UCF-101 dataset. We conclude that Fr\'echet Distance has the worst sample efficiency among metrics, while I3D features exhibit the worst sample efficiency across feature spaces.

%In Figure \ref{fig:train_test_convergence_rate}, our metric exhibits significantly better sample efficiency compared to FVD. 

% Generating video samples is computationally expensive,
\hypertarget{convergence-samples}{}
As shown in Figure \ref{fig:number_sample_convergence}, it often takes thousands of video clips for FVD features to converge; however, many datasets contain insufficient amount of unique videos to reach this convergence, making it challenging to define a robust distribution in such high-dimensional spaces~\citep{ponttuset2017davis, geiger2013kitti, ebert2017bair}. This is often worked around by transforming videos into shorter, partly overlapping clips. This method is problematic and biases the metric due to the repetitiveness of the data. This fault has remained largely challenged. For instance, as shown in Figure~\ref{fig:bair_issue}, the BAIR dataset's size and sample efficiency issues are particularly noteworthy. Despite these limitations, BAIR dataset remains a widely-used benchmark for video generation, with numerous studies reporting FVD results on it~\citep{yu2023magvit, chenfei2021nuwa, voleti2022mcvd}.

We note three key hurdles stemming from this sample efficiency issue in video generation: (1) \emph{Data size:} Limited samples compromise estimate reliability, undermining robust statistical analysis; (2) \emph{Computational resources:} Generating samples is computationally expensive and time-demanding; (3) \emph{Metric convergence speed:} Slow convergence rates hinder accurate assessments. While dataset size and computational resources are largely beyond our control, we can address the third concern by selecting metrics with higher sample efficiency where convergence happens with less samples.

%% file: sections/noise_and_human_study.tex
This section explores the effects of videos distorted with noise and videos generated at varying model training checkpoints on metric reliability, assessing their impact on: (1) metric accuracy, (2) sample efficiency and (3) human metric alignment.
\subsection{Noise \& Generation Models and Their Impacts on Metric Measurement~\label{sec:noise_impact}}

\begin{table}[h]
\centering
\begin{tabular}{cccccc}
\toprule
\textbf{Metric} & \textbf{No Noise} & \textbf{Blur (low)} & \textbf{Blur (medium)} &\textbf{Blur (high)}\\
\midrule
FVD & \textcolor{red}{\textbf{$64.7\pm0.002$}} & \textcolor{red}{\textbf{$60.4\pm0.001$}} & $65.6\pm0.002$ & $86.7\pm0.003$\\
\ourmetric & $0.411\pm0.000$ & $0.627\pm0.000$ & $1.142\pm0.000$ & $1.630\pm0.000$ \\
\bottomrule
\end{tabular}
\caption{The table shows average FVD and \ourmetric~distances between training and testing set feature distributions under various blur distortions. The testing video dataset is subjected to noise distortions, including low blur ($\sigma\sim[0.05, 0.75]$), medium blur ($\sigma \sim [0.1, 1.5]$), and high blur ($\sigma \sim [0.01, 3]$), where $\sigma$ represents the per-frame blur intensity, and a larger range indicates greater temporal inconsistency. The experiment is replicated 10 times to account for variability. Our analysis reveals that FVD fails to detect low blur noise and incorrectly suggests an improvement in video quality. \emph{Note: To improve readability, we standardize \ourmetric~by applying a scaling factor of 100 to the \vjepaft+MMD polynomial distance.}}
\label{tab:noise_inconsistency}
\end{table}
This study investigates metric reliability when presented with videos affected by three noise distortion types (salt and pepper noise, temporal blur and elastic distortion) and two image-to-video generation models (I2V-Stable Video Diffusion and Open-Sora).

Salt and pepper noise, a type of impulsive noise, spatially corrupts visual data by randomly altering pixel values to extreme intensities. Elastic noise distortion from ~\citep{ge2024content} primarily introduces temporal distortions and occasionally deforms object shapes. In addition, we introduce temporal blur noise which involves applying Gaussian kernels of varying strengths to blur frames, preserving appearance and shape integrity while focusing on temporal distortion. The two generative models we used were adopted from open-source repositories, utilizing the provided checkpoints~\citep{zangwei2024opensora, von2022diffusers}. Detailed inference configurations for these models are in Appendix~\ref{sec:generative_model}. \hypertarget{generative-specs}{}

We highlight some of our results in Table~\ref{tab:noise_inconsistency}, and the remaining results are in Figures~\ref{fig:fvd_noise_impact} and ~\ref{fig:FD_noise_impact_others}. The key findings in these experiments include:
\begin{enumerate}[leftmargin=*]%,noitemsep,nolistsep]
    \item 
    Metrics in the I3D feature space are impacted by salt and pepper noise (a spatial distortion) significantly more than by other types of distortions. This aligns with the findings of~\citeauthor{ge2024content}, \emph{demonstrating that the I3D feature space is highly sensitive to spatial distortions but less responsive to temporal distortions}. As shown in Section~\ref{sec:human_study}, I3D does not align with human preferences with respect to distortions.
    \item 
    \emph{I3D and \videomaept~are not ideal feature spaces for building video quality metrics, as they do not capture blur distortion well.} Notably, they perceive a testing distribution with slight artificial blur added to the frames as more similar to the training distribution than the original testing distribution. In fact, they estimate the low-blur distorted videos are 10\%-20\% closer to the ground-truth training distribution.
    \item 
    Our experimental evaluation reveals that the choice of feature space have a greater impact on distance values than the choice of distribution distance metric. Moreover, the metrics exhibit consistent performance across diverse datasets, demonstrating robustness.
\end{enumerate}

\subsection{Metric Robustness Assessment With Progressive Distortion level and Training Duration}

\begin{figure}[h]%
    \centering
    \setlength\tabcolsep{3pt} % default: 6pt
    \subfloat[FVD vs \ourmetric]{{\includegraphics[width=0.32\linewidth]{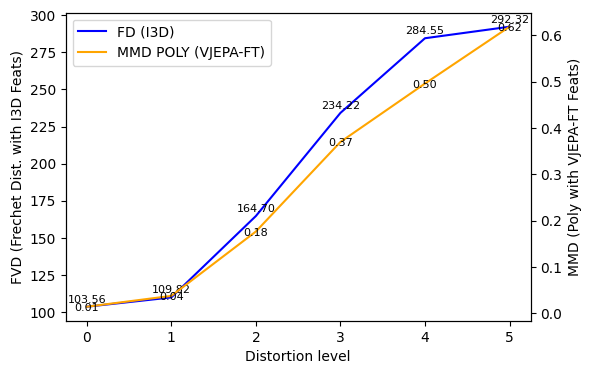} }}%
    \subfloat[FVD vs $\text{VMAE}_\text{SSv2}$+$\text{MMD}_{\text{POLY}}$]{{\includegraphics[width=0.32\linewidth]{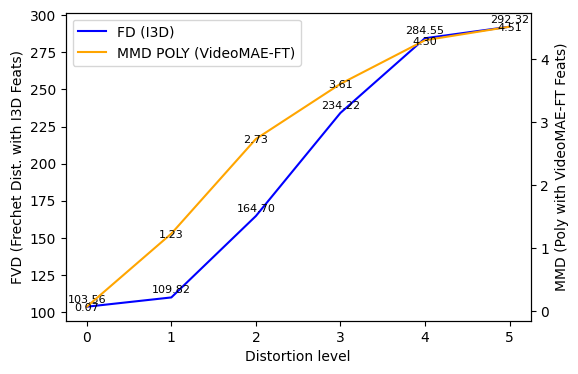} }}%
    \subfloat[FVD vs \vjepapt+$\text{MMD}_{\text{POLY}}$]{{\includegraphics[width=0.32\linewidth]{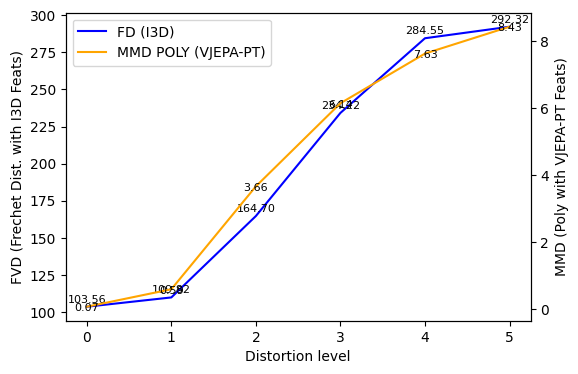} }}%
    \caption{How metric distance changes as temporal blur increases. Specifically, temporal blur distortion is controlled by varying the sigma range (\(\sigma\)) using the distortion level (\(\lambda\)), with $\sigma=[0.1-0.01\lambda, 0.75+0.8\lambda]$. The study is carried out on the UCF-101 dataset.}%
    % [for ICLR] \vspace{-0.3cm}
\label{fig:noise_progression}
\end{figure}

\textbf{Distortion Level}~~~In Figure~\ref{fig:noise_progression}, we conduct a study to evaluate the performance of metrics in assessing various blur noise distortions on the UCF-101 dataset. All metrics identified the decline in video quality as noise levels increased.

\begin{figure}[ht]%
    \centering
    \setlength\tabcolsep{3pt} % default: 6pt
    \subfloat[FVD vs \ourmetric]{{\includegraphics[width=0.32\linewidth]{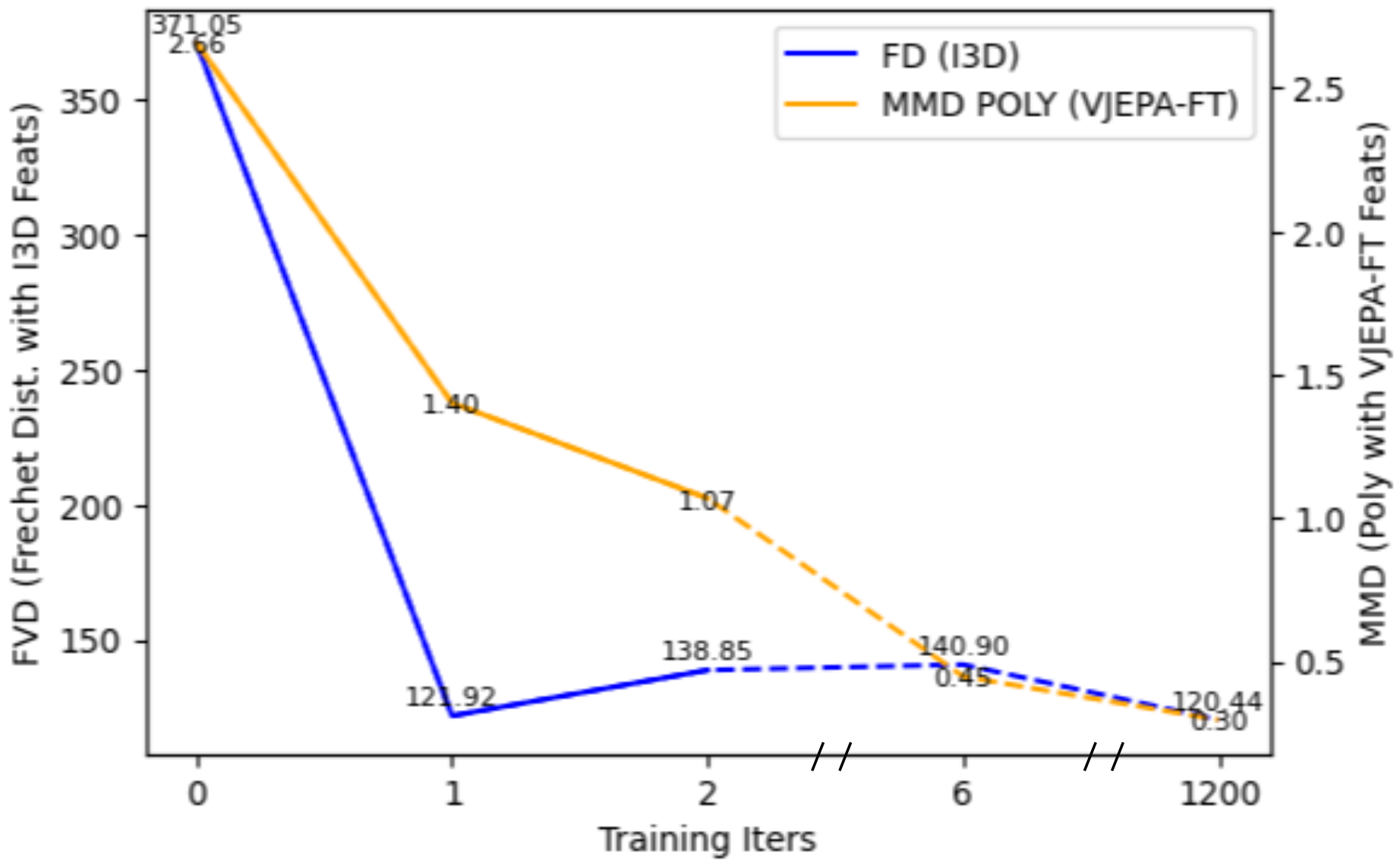} }}%
    \subfloat[FVD vs $\text{VMAE}_\text{SSv2}$+$\text{MMD}_{\text{POLY}}$]{{\includegraphics[width=0.32\linewidth]{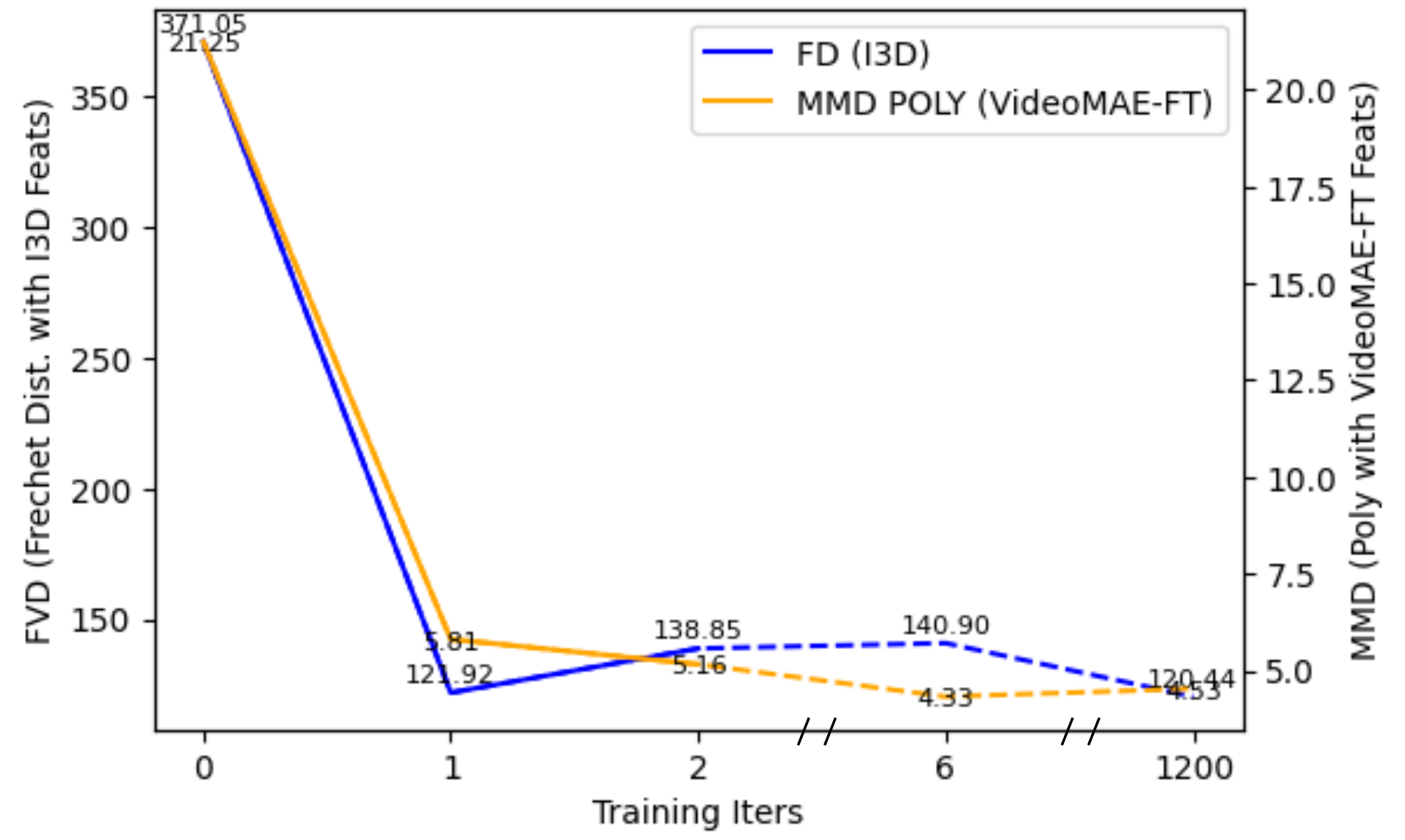} }}%
    \subfloat[FVD vs \vjepapt+$\text{MMD}_{\text{POLY}}$]{{\includegraphics[width=0.32\linewidth]{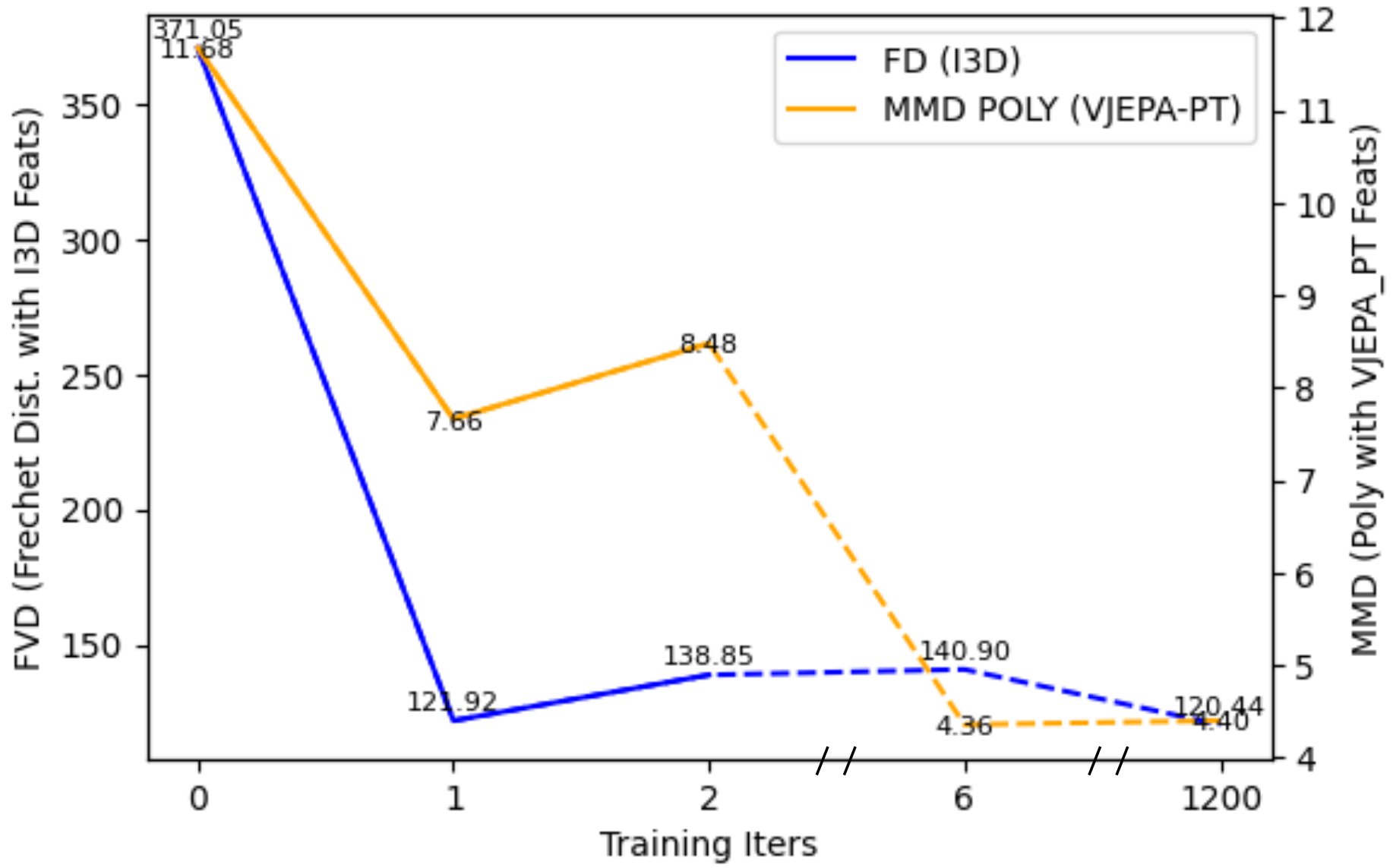} }}%
    \caption{Ctrl-V is fine-tuned on BDD. Visual inspection show incremental improvements in generation quality at each training step. This is captured by, 
    \ourmetric~(\vjepaft+$\text{MMD}_{\text{POLY}}$). However, FVD (I3D+FD), \videomaeft+$\text{MMD}_{\text{POLY}}$ and \vjepapt+$\text{MMD}_{\text{POLY}}$ fail to detect incremental improvements. The Spearman coefficient correlation values for the X and Y axes are -1, -0.6, -0.9 and -0.8 for \ourmetric, FVD, \videomaeft+$\text{MMD}_{\text{POLY}}$ and \vjepapt+$\text{MMD}_{\text{POLY}}$, respectively, with only \ourmetric~showing statistical significance.}
  \label{fig:svd_ft_progression}
\end{figure}

\textbf{Training Duration}~~~We further aim to evaluate metrics for generative models, focusing on their ability to track changes in video quality throughout training (Figure~\ref{fig:svd_ft_progression}). Due to the computational expense of training video generation models from scratch, we fine-tune Stable Video Diffusion's weights on the BDD dataset using Ctrl-V's code~\citep{luo2024ctrlv}. Ctrl-V uses a pre-trained SVD model but modifies the input padding strategy to enable multi-frame conditioning. Initially, the visual quality of generated videos is poor, but it improves over the training time. We utilize these fine-tuning steps to assess our metrics' robustness in evaluating fine-tuned model checkpoints. We expect a good metric to decrease steadily over time and thus have a negative correlation close to 1 in magnitude. We visualize several checkpoint generations in Figure~\ref{fig:ctrlv_training}. \hypertarget{ctrlv-training}{}

\textbf{Results}~~~Only \ourmetric~(\vjepaft+$\text{MMD}_{\text{POLY}}$) successfully tracks incremental gains in all checkpoints, whereas FVD (I3D+FD), \videomaeft+$\text{MMD}_{\text{POLY}}$ and \vjepapt+$\text{MMD}_{\text{POLY}}$ do not.

\subsection{Sample Efficiency Under Noise Distortion} \hypertarget{noise-distortion}{}

Alongside Section~\ref{sec:noise_impact}, we investigate the sample efficiency of various metrics under noisy conditions. Specifically, we measure the number of samples it takes for the distance between the original training distribution and the noise-added testing distribution to stabilize/converge. The noise-added testing set essentially simulates a set of generations.
Our findings indicate that: \ourmetric~remains much more sample efficient compared to FVD in this condition.
%(2) Sampling efficiency is lowered as distortion lessens (Figure~\ref{fig:noise_vs_metric_convergence} shrinks when more blur distortions are being introduced). This suggests more samples are needed to accurately model the generation distribution as generation quality gets better. As such, as future video generation becomes more and more realistic, sample efficiency of the metric used to evaluate these generations also becomes more important.
We present our experiment results in Appendix~\ref{sec:noise_and_convergence}.

\subsection{Human Evaluation\label{sec:human_study}}
\hypertarget{human-evaluation}{}

To investigate human alignment on the perception of video quality degradation under various noise distortions, we conduct a small scale survey. We randomly select 24 videos from each of the UCF-101 and Sky Scene test sets, originally captured at 30 frames per second (fps), and subsample them to 25 frames at 7 fps to generate clips three seconds in length. Four types of noise distortions are systematically applied: blur at two increasing levels, elastic distortion, and salt and pepper noise. The specific parameters for each distortion are found in Appendix~\ref{sec:human_eval}. To mitigate border effects resulting from elastic distortion, all videos are center-cropped to $230\times 310$ pixels. 

Separate surveys for the UCF-101 and Sky Scene datasets are conducted in a randomized order, presenting participants with anonymized video pairs differing only in noise type. Participants evaluate each pair of noise types under four distinct comparisons, rating either one video as superior in quality or indicating no observable difference. Each comparison is rated by 20 independent raters sourced from an academic community. The raters are not aware of the details of the datasets or distortion applied and rate the videos purely on visual quality alone.

Following the Analytic Hierarchy Process (AHP)~\citep{Saaty1987TheAH}, a pairwise comparison matrix is used to aggregate the responses. A priority vector for the noise distortion types is found by normalizing the pairwise comparison matrix by column and then averaging it by row. To align the priority vector with the scale used in distribution distance metrics, where a value of 0 indicates ideal quality, the priority vector is inverted and re-normalized. For each of the distance metric-feature space combinations, its score for the noise types under study are put into a vector, normalized, and then compared to that of the human survey using cosine similarity. Evaluation results are summarized in Figure~\ref{fig:human_evaluation_ucf_sky} and more details can be found in Appendix~\ref{sec:human_eval}. 

\begin{figure}[h!]%
    \centering
    \setlength\tabcolsep{3pt} % default: 6pt
    \subfloat[UCF-101]{{\includegraphics[width=0.5\linewidth]{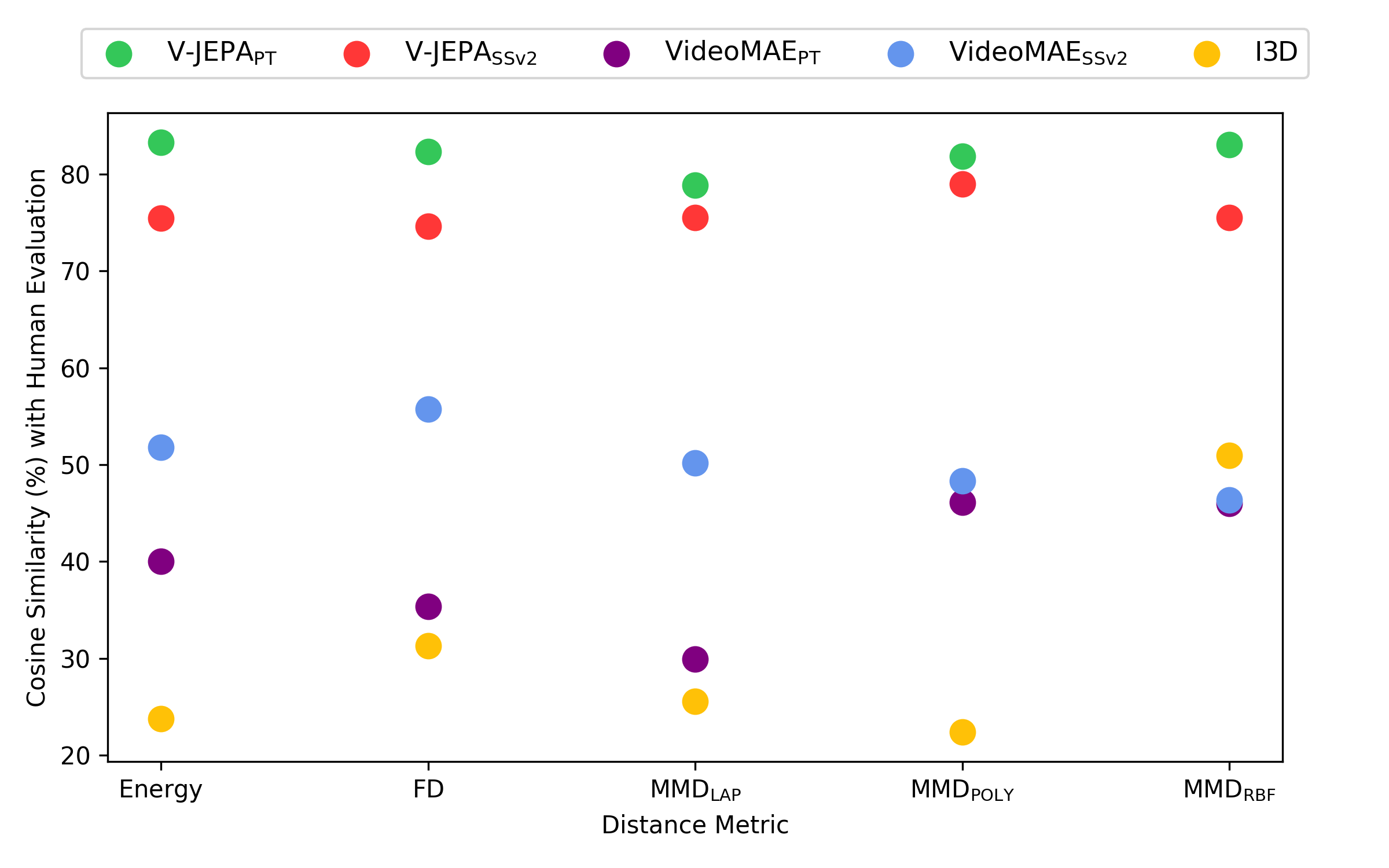} }}%
    \subfloat[Sky Scene]{{\includegraphics[width=0.5\linewidth]{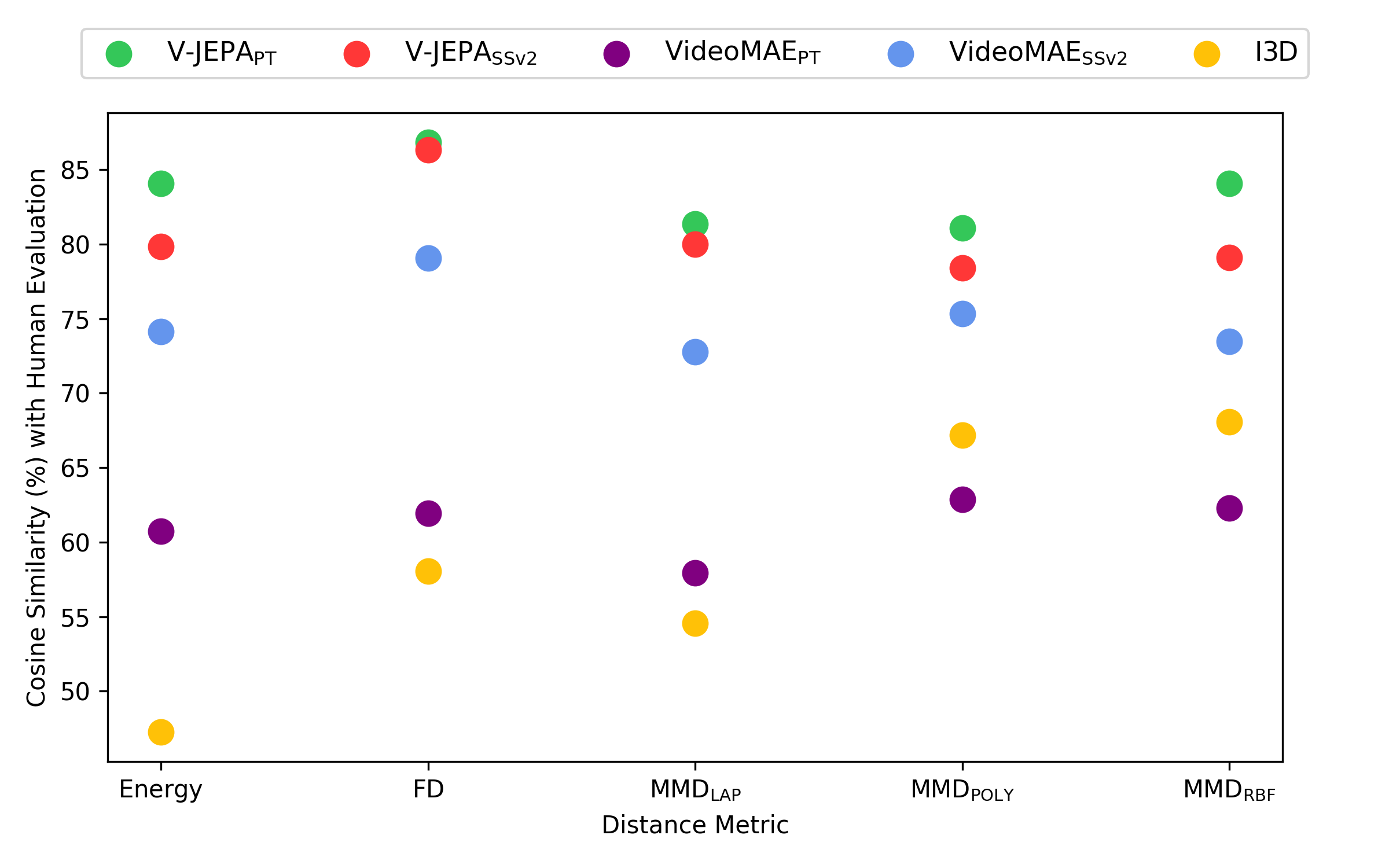} }}%
    \caption{Alignment of human evaluation with distribution distance metrics. Metrics computed in V-JEPA feature spaces surpass that of both I3D and VideoMAE in terms of alignment with human evaluation.}%
\label{fig:human_evaluation_ucf_sky}
\end{figure}

\textbf{Results}~~~While metrics within a feature space generally perform at the same level, distances calculated in the feature space of \vjepaft~or \vjepapt~model resoundingly outperform both I3D and VideoMAE-based metrics in terms of alignment with human evaluation. Among raters, there were agreements of 83.70\% and 53.54\% on the UCF-101 and Sky Scene datasets, respectively. Perhaps intuitively, humans are much more confident assessing content with human activity than with more abstract visuals, where subjective interpretations can vary significantly.

This cross-sectional study reveals that standard and commonly used frameworks are unable to reliably capture the extent to which arbitrary noise distortion degrades video quality. Further, this corroborates our previous insight that the model used to produce the feature representations has a greater impact on video distribution distance calculations than the metric used to calculate the distance itself.

%% file: sections/conclusion.tex
In this study, we carefully look at many aspects of video metrics and find \textbf{\ourmetric} to be the best choice.

First, we show that the normality assumption made by the Fr\'echet Distance (FD) does not hold true in the video feature spaces, and this becomes more evident as the duration of the videos increases.

Second, we discuss two challenges with FD. 1) Estimating the covariance matrix, which is required in FD, is challenging due to the high dimensionality of the latent space. However, we found no enhancement from reducing this dimension using autoencoders. 2) The sample efficiency of FD is low; through extensive comparison, we find that Maximum Mean Discrepancy (MMD) with V-JEPA exhibits much higher sample efficiency across all datasets tested.

Third, we investigate the impact of noise on feature spaces and found that I3D was more sensitive to image quality distortion than temporal distortion, and that I3D and $\text{VideoMAE}_\text{SSv2}$ does not capture blur distortion well. On the other hand, \vjepaft~stands out as the more robust feature space among them.

Fourth, we observe the correlation between metric with distortion level and training duration. We show that while both FVD (FD+I3D) and \ourmetric~ ($\text{MMD}_{\text{POLY}}$+\vjepaft) are positively correlated with distortion level (higher distance with higher distortion), only \ourmetric~ is highly negatively correlated to training duration (lower distance with more training).

Finally, we complete a human alignment study on the perception of video quality under various noise distortions. The results of the study suggest that among all the metrics investigated, \ourmetric~ has the highest similarity to human preferences.

Based on our comprehensive analysis, \ourmetric~emerges as the most effective and practical metric for guiding the current surge in video generation research. To facilitate the usage of \ourmetric, we provide simple and easy-to-use code that, given its striking benefits, hope the community will embrace.

\paragraph{Limitations} While we performed extensive study of different feature space with different distance metrics on different datasets, there are more datasets to be tested on and more types of noise distortions can be investigated.
% A portion of participants in our survey are familiar with or part of the machine learning community, this may be a source of bias. 
While we choose \ourmetric~($\text{MMD}_{\text{POLY}}$+\vjepaft) as our main proposed method, other choices (see Figure \ref{tab:human_evaluation_results} for the list) such as the Energy distance with \vjepapt~has slightly higher alignment with human evaluation. It is a decision we made considering the large gain in sample efficiency from \ourmetric. %Currently, we have no explanation to the better convergence rate when using $\text{MMD}_{\text{POLY}}$ with \vjepaft, more investigation could be done.

%% file: sections/appendix.tex
\def\house{\hbox{\kern2pt \vbox to8pt{}% 
   \pdfliteral{q 0 0 m 0 3 l 3 6 l 6 3 l 6 0 l 4 0 l 4 3 l 2 3 l 2 0 l f
               1 j 1 J -1.2 3 m 3 7.2 l 7.2 3 l S Q }%
   \kern 8pt}}

\input{sections/appendix/stats_background}

% [IMPORTANT] TODO: OLGA WILL DO THIS PART.
\input{sections/appendix/metric_configuration}

\input{sections/appendix/frechet_semimetric_proof}

\input{sections/appendix/non_gaussianity_evidences}

\input{sections/appendix/toy_experiment_and_convergence}

\input{sections/appendix/noise_human_study}
\clearpage

\input{sections/appendix/generative_models}

% will add these guidelines to our webpate
% \input{sections/appendix/metric_guidelines}

%% file: sections/appendix/stats_background.tex
\section{Statistical Distributions}
\subsection{Multivariate Gaussian Distribution}
The probability density function of a $k$-dimensional multivariate Gaussian is given by:
\begin{equation}
    P(\rx) = \frac{1}{(2\pi)^{k/2}|\Sigma|^{1/2}}\exp{\bigg(-\frac{1}{2}\big(\rx-\mu\big)^{\intercal}\Sigma^{-1}\big(\rx-\mu\big)\bigg)}
\end{equation}
where \(\rx\in\sR^k\) is the $k$-dimensional random sample, \(\mu\in\sR^k\) is the mean vector, \(\Sigma\in\sR^{k\times k}\) is the covariance matrix which is symmetric and positive definite, \(|\Sigma|\) denotes the determinant of the covariance matrix, and \(\Sigma^{-1}\) denotes the inverse of the covariance matrix.

\subsection{Gaussian Mixture Models}
 A Gaussian Mixture Models (GMM) with $c$ clusters is a probabilistic model that assumes the data is generated from a mixture of $c$ Gaussian distributions, each with its own mean and covariance. The probability density function of the GMM is given by: \(P(\rx) = \sum^c_{i=1}\pi_i\gN(\rx|\mu_i,\Sigma_i)\) where the cluster weights \(\pi_i\) sum up to 1. The parameters of GMMs are often estimated using iterative algorithms, such as Expectation-Maximization (EM) algorithm~\citep{dempster1977em}, as there is no closed-form solution to maximize its likelihood function.

\subsection{Preliminary: Linear Transformation of Multivariate Gaussian Distribution and Linear Dimensionality Reduction Methods\label{sec:linear_transformation_multivariate_gaussian}}

\hyperlink{Examining-FVD}{\house} Back to paper

Let \(\rx\sim\gN(\mu, \Sigma)\) follow a multivariate Gaussian distribution with mean vector \(\mu\) and covariance matrix \(\Sigma\). Let \(\mA\) be a matrix and \(\vb\) be a vector. We are interested in the linear transformation \(\ry = \mA\rx+\vb\). The following is the proof of \(\ry\) is also a multivariate Gaussian distribution. Specifically, \(\ry \sim\gN(\mA\mu+b, \mA\Sigma\mA^\intercal)\).

The moment generating function for $\rx$ is 
\begin{equation}
        M_\rx(t)=\E(\exp[t^T\rx])=\exp\Big[t^T\mu+\frac{1}{2}t^T\Sigma t\Big]
\end{equation}
and the moment generating function for $\ry$ is given by
\begin{equation}
\begin{split}
    M_\ry(t) &= \E \Big(\exp\big[t^T(\mA\rx+\vb)\big]\Big)\\
    &=\exp[t^T\vb]\E\big(\exp[t^T\mA\rx]\big)\\
    &=\exp[t^T\vb]M_\rx(\mA^Tt)\\
    &=\exp\Big[t^T(\mA\mu+\vb)+\frac{1}{2}t^T\big(\mA\Sigma\mA^T\big)t\Big]
\end{split}
\end{equation}
This indicates that the moment generating function of \( \ry \) aligns with the moment generating function of the multivariate Gaussian distribution. Therefore, \( \ry \) is a random variable that follows a multivariate Gaussian distribution (detailed proof can be found in \citet{Soch2024stat_proof}).

Principal Component Analysis (PCA) and Linear Discriminant Analysis (LDA) are statistical methods used to reduce the number of variable dimensions in a dataset. PCA is a process of linear transformation that involves mapping data from a higher dimensional space to a lower dimensional space by identifying the directions in which the data varies the most~\citep{pearson1901pca}. LDA entails a linear transformation that maps data from a higher-dimensional space to a lower-dimensional space in order to effectively separate multi-class objects~\citep{Martnez2001PCAvsLDA}. Because PCA and LDA are linear transformations, and we've shown that data from a multivariate Gaussian distribution remains Gaussian after linear transformations, we can conclude that applying PCA or LDA to data from a multivariate Gaussian distribution preserves its Gaussian properties.

\subsection{Distribution Distance Metrics Overview\label{sec:distribution_distance_overview}}
\hyperlink{other-distribution-metrics}{\house} Back to paper

\emph{Mixture Wasserstein ($\mw$) \citep{delon2020gmmot}: }Optimal transport is a mathematical field that deals with finding a transport plan that minimizes the total cost of moving the mass from the source distribution to the target distribution~\citep{monge1781mémoire, montesuma2023recentadvancesoptimaltransport}. Recently, \citet{delon2020gmmot} proposed a Wasserstein-type distance within a novel optimal transport framework for Gaussian Mixture Models (GMMs) with restricted couplings. By confining the set of possible coupling measures to GMMs, they derive a simple, discrete formulation of the distance metric, making it computationally efficient for problems with high dimensions. The distance is called Mixture Wasserstein and is denoted as $\mw$. The $\mw$ distance is always upper bounded by Wasserstein distance ($W_2$) plus the variances of the Gaussian components. In addition, its computational complexity is solely determined by the number of clusters.

\emph{Energy Statistic~\citep{baringhaus2004energy, szekely2004energy}: }Energy statistic measures the difference between distributions based on pairwise distances between points. Given \(\{\rx_1, \dots, \rx_m\}\) are random samples generated from distribution \(P\) and \(\{\ry_1, \dots, \ry_n\}\) are random samples generated from distribution \(Q\), the energy distance $\gE(P,Q)$ is given by:
\begin{equation}
    \gE(P,Q)=\frac{2}{mn}\sum_{i=1}^m\sum_{j=1}^n\Vert\rx_i-\ry_j\Vert-\frac{1}{m^2}\sum_{i=1}^m\sum_{j=1}^m\Vert\rx_i-\rx_j\Vert-\frac{1}{n^2}\sum_{i=1}^n\sum_{j=1}^n\Vert\ry_i-\ry_j\Vert.
\end{equation}
The energy statistic is appropriate for comparing complex distributions without making assumptions about a particular underlying distribution.

\emph{Maximum Mean Discrepancy (MMDs)~\citep{Gretton2012MMD}: }MMD is a general class of kernel-based sample tests that maximize the mean difference between samples from two distributions by optimizing over all data transformations \(f\) within a function space \(\gF\). Some popular kernel functions used for MMD include: linear, polynomial, sigmoid, Laplace and RBF (Gaussian) kernels.

Given two sets of features, \( X = \{ \mathbf{x}_1, \mathbf{x}_2, \ldots, \mathbf{x}_m \} \) and \( Y = \{ \mathbf{y}_1, \mathbf{y}_2, \ldots, \mathbf{y}_n \} \), sampled from \( P \) and \( Q \), \( d^2_{MMD}(P, Q) \) with a given kernel, \( k \), is given by~\citep{jayasumana2024rethinkingFID}:

\begin{equation}
    \hat{d}^2_{MMD}(X, Y) = \frac{1}{m(m - 1)} \sum_{i=1}^{m} \sum_{\substack{j=1 \\ j \neq i}}^{m} k(\mathbf{x}_i, \mathbf{x}_j) + \frac{1}{n(n - 1)} \sum_{i=1}^{n} \sum_{\substack{j=1 \\ j \neq i}}^{n} k(\mathbf{y}_i, \mathbf{y}_j) - \frac{2}{mn} \sum_{i=1}^{m} \sum_{j=1}^{n} k(\mathbf{x}_i, \mathbf{y}_j).
\end{equation}

Like the energy statistic, MMD is distribution-free, requiring no assumptions about the underlying distributions of $P$ or $Q$.

%% file: sections/appendix/metric_configuration.tex
\section{Computation Configurations}
\subsection{Feature Extractors}
Feature extraction is performed on a \emph{single NVIDIA RTX 4080 GPU with float32 precision}. However, VideoMAE-v2 features require a more specialized setup: \emph{a single NVIDIA RTX A100 GPU with 80G memory}. Notably, VideoMAE-v2 precision varies by clip length: \emph{float32 for clips under 64 frames and float16 for longer clips}. We use batch sizes of 10 (clips $<$ 64 frames) and 2 (clips $\ge$ 64 frames) for feature extraction.
\begin{description}
    
    \item[I3D Configuration\footnotemark] \footnotetext{Feature extractor for FVD} We adopt the recommended feature extractor from the FVD paper~\citep{unterthiner2019fvd}: I3D logits features pre-trained on Kinetics-400.

    \item[$\text{VideoMAE}_\text{PT}$ and $\text{VideoMAE}_\text{SSv2}$] We compute VideoMAE features using the official PyTorch implementation~\citep{wang2023videomae}, following the guidelines outlined in \citeauthor{ge2024content}.
    
    \item[\vjepapt~Configuration]
    We compute V-JEPA features using the official PyTorch implementation~\citep{bardes2024revisiting}. Specifically, our V-JEPA pretrained model, referred to as \vjepapt, consists solely of the V-JPEA encoder. Consistent with previous research \citep{ge2024content}, we extract features from the encoder and average them across all feature tokens.
    
    \item[\vjepaft~Configuration\footnotemark] \footnotetext{Feature extractor for \ourmetric}
    We compute V-JEPA features using the official PyTorch implementation~\citep{bardes2024revisiting}. Specifically, our V-JEPA classifier model, referred to as V-JEPA SSv2 fine-tuned model or \vjepaft, consists of two components: (1) a V-JEPA encoder, and (2) an adaptive probe with attentive pooler. Consistent with previous research \citep{ge2024content}, we exploit the pre-logit features generated by the attentive pooler.
\end{description}

\subsection{Distribution Metric Configurations}
\begin{description}
    \item[Fr\'echet Distance\footnotemark]\footnotetext{Distribution metric for FVD} We calculate the Fréchet Distance using torchaudio's functional API, passing mean and covariance statistics to \texttt{frechet\_distance}.
    \item[Energy statistic] We utilize the \url{https://github.com/josipd/torch-two-sample/} repository to calculate the energy statistic.
    \item[Mixture Wasserstein] We utilized the official implementation from  \citeauthor{delon2020gmmot}'s repository to calculate the Mixture Wasserstein distance.
    \item[Mean Maximum Discrepancy] We compute MMD distances using code from \citep{repo_transferlearning}, with kernel-specific parameters:
    \begin{itemize}
        \item \textbf{RBF/Laplacian:} $\gamma=1/\text{ndim}$
        \item \textbf{Polynomial\footnote{Distribution metric for \ourmetric}:} degree = 2, $\gamma=1$, coef = 0.
    \end{itemize}
\end{description}

%% file: sections/appendix/frechet_semimetric_proof.tex
\section{Fr\'echet Video Distance: A Semi-Metric}
A distance metric on a set \(\sX\) is a function \(d: \sX\times \sX \rightarrow \sR\) that satisfies the following properties for all points \(\rx, \ry, \rz \in \sX \):
\begin{enumerate*}[font=\bfseries]
    \item 
    \emph{Non-negativity: }\(d(\rx, \ry) \ge 0\);
    \item 
    \emph{Identity of indiscernible: }\(d(\rx, \ry) = 0 \iff \rx=\ry\);
    \item 
    \emph{Symmetry: }\(d(\rx,\ry)=d(\ry, \rx)\); and
    \item 
    \emph{Triangle inequality: }\(d(\rx, \rz)\le d(\rx, \ry)+d(\ry+\rz)\).
\end{enumerate*}

Fréchet Distance (FD) satisfies all the properties of a metric, except for the triangle inequality, which establishes it as a semi-metric. The triangle inequality is a crucial property that underpins the linearity of a metric and offers valuable interpretability. We will explore this aspect in greater detail later. For now, we will demonstrate the proof for FD being a semi-metric.

The FD of multivariate Gaussian distributions, represented by Equation ~\ref{eq:squared_FD_gaussian_distance}, is composed of two terms: the squared-Euclidean distance between the mean vectors and a term involving the covariance matrices. In order to demonstrate that FD is a semi-metric, we will examine these two components individually.

The first-term, \((\mu_P-\mu_Q)^2\): By definition, the squared Euclidean distance between the mean vectors is non-negative and equals zero if and only if the mean vectors are identical, i.e., \((\mu_P-\mu_Q)^2\ge 0\text{ with equality iff } \mu_P=\mu_Q\). Also, the squared Euclidean distance is a symmetric operation. Thus, it satisfies the first three properties. However, the triangle inequality is not satisfied. A counter-example to prove this is: let \( \mu _A= [ 0, 0 ] ^ \intercal , \mu _B= [ 1, 1 ] ^ \intercal , \mu _C= [ 5, 5 ] ^ \intercal \) , and \( ( \mu _A- \mu _C)^2=50 \ge ( \mu _A- \mu _B)^2+( \mu _B- \mu _C)^2=34 \).

The second-term, \(\Tr\big(\Sigma_P+\Sigma_Q-2(\Sigma_P\Sigma_Q)^{\frac{1}{2}}\big)\): According to \citet{dowson1982frechet_gaussian}, the square root of the second term is considered a natural metric on the space of real covariance matrices of a given order. This implies that the first three properties should hold. Below are the corresponding proofs:
\begin{small}
\begin{equation}
%\tiny
    \begin{split}
        \Tr\bigg(\frac{\Sigma_P+\Sigma_Q}{2}\bigg) \ge \Tr\Big(\sqrt{\Sigma_P\Sigma_Q}\Big)\quad &\text{Arithmetic mean is greater than or equal to geometric mean.}\\
        \Sigma_P = \Sigma_Q \Rightarrow \Tr\bigg(\frac{\Sigma_P+\Sigma_P}{2}\bigg) = \Tr\Big(\sqrt{\Sigma_P\Sigma_P}\Big)= \Tr(\Sigma_P)\quad & \text{If the covariance matrices are the same, the 2nd-term becomes 0.}\\
        \Tr\bigg(\frac{\Sigma_P+\Sigma_Q}{2}\bigg) = \Tr\Big(\sqrt{\Sigma_P\Sigma_Q}\Big) \Rightarrow \Sigma_P = \Sigma_Q\quad & \parbox[t]{0.5\textwidth}{Arithemetic mean equals geometric mean when covariance matrices are identical.}
    \end{split}
\end{equation}
\end{small}
To illustrate the symmetric property of the second term, let's assume \( \mu _P = \mu _Q = \mathbf { 0 } \) for simplicity. Let \( \tX \) represent column vectors sampled from a normal distribution with mean \( \mu _P \) and covariance \( \Sigma _P \), and \( \tY \) represent column vectors sampled from a normal distribution with mean \( \mu _Q \) and covariance \( \Sigma _Q \). There exists a linear transformation \( \Gamma \) such that \( \tY = \Gamma \tX \). According to \citet{chafai2010blog}, \( D_{ \text{Fr\' echet}}^2(P,Q) \) can be derived as:
\begin{equation}
    D_{ \text{Fr\' echet}}^2(P,Q) = \Tr(\Sigma_P)+\Tr(\Sigma_Q)-\E(\langle\tX, \Gamma\tX\rangle)
\end{equation}
To prove the symmetric property, we only need to demonstrate that \( \E ( \langle \tX , \Gamma \tX \rangle ) = \E ( \langle \Gamma \tX , \tX \rangle ) \) because the remaining terms are symmetric. Below is the corresponding proof:
\begin{equation}
%\small
\begin{split}
    &\E ( \langle \tX , \Gamma \tX \rangle ) = \E(\tX(\Gamma\tX)^\intercal) = \E(\tX\tX^\intercal\Gamma^\intercal) = \Sigma_P\Gamma^\intercal,\quad\quad\E ( \langle \Gamma \tX , \tX \rangle ) = \E(\Gamma\tX\tX^\intercal) = \Gamma\Sigma_P\\
     &\Sigma_P\Gamma^\intercal = \Sigma_P^\intercal\Gamma^\intercal = \Gamma\Sigma_P\Rightarrow \E ( \langle \tX , \Gamma \tX \rangle ) = \E ( \langle \Gamma \tX , \tX \rangle ) \quad \text{Covariance matrices are symmetric: } \Sigma_P=\Sigma_P^\intercal
\end{split} 
\end{equation}

Lastly, the triangle inequality is not satisfied. A counter-example to prove this is: let \(\Sigma_A = \big(\begin{smallmatrix}1 & 0 \\ 0 & 1\end{smallmatrix}\big)\), \(\Sigma_B = \big(\begin{smallmatrix}4 & 0 \\ 0 & 4\end{smallmatrix}\big)\), \(\Sigma_C = \big(\begin{smallmatrix}9 & 0 \\ 0 & 9\end{smallmatrix}\big)\),

\begin{equation*}
    \Tr(\Sigma_A+\Sigma_C-2(\Sigma_A\Sigma_C)^{\frac{1}{2}}) = 8 \ge \Tr(\Sigma_A+\Sigma_B-2(\Sigma_A\Sigma_B)^{\frac{1}{2}}) + \Tr(\Sigma_B+\Sigma_C-2(\Sigma_B\Sigma_C)^{\frac{1}{2}}) = 4
\end{equation*}

When combining the first and second terms with addition, their mathematical properties still hold because their input parameters are different and do not affect each other.

The triangle inequality property is crucial, as it ensures that the distance between two points remains consistent and intuitive. In video generation, models aim to learn the underlying patterns of a real data distribution (\(R\)) by training on an empirical dataset (\(R_{\text{empirical}}\)). This involves understanding a probability distribution over potential videos, allowing newly generated videos to closely resemble the structure and content of the observed data. Note that the empirical distribution is an approximation, potentially biased towards the specific sample. The discrepancy between the true distribution and empirical distribution can be quantified using metrics, such as:
\begin{equation}
    R_{\text{empirical}}\xrightarrow{N\rightarrow\infty} R\implies D(R_{\text{empirical}}, R)\xrightarrow{N\rightarrow\infty}0
\end{equation}
where \(N\) is the number of samples. The equation demonstrates that as the sample size approaches infinity, the discrepancy between the empirical and true distributions converges to zero. However, in practice, most datasets are finite, and training is typically done on a limited number of samples. As a result, the difference between the empirical and true distributions is bounded by a specific value, rather than reaching zero. Triangle inequality provides insight into the upper-bound of the model generation quality based on the true distribution, expressed as \( D(R_{ \text{empirical}} , R) + D(G, R_{\text{empirical}} ) \ge D(G, R) \).

On the other hand, much of the video generation work involves using pre-trained models to perform zero-shot inference on different datasets in order to test the models' performance and domain adaptation abilities~\citep{hong2022cogvideo, singer2022makeavideo, blattmann2023stable, zhou2023magicvideo}. However, measuring a model’s generation quality on another dataset requires setting up the model for the new dataset and generating numerous samples, which demands significant time, effort, and computational resources. If a distribution distance metric follows the triangle inequality and one only wants to compute an upper bound to validate a model’s generation quality on another dataset, it is sufficient to compute: \( D(R_ { \text {empirical} }^{\text{dataA}} , G) + D(R_ { \text {empirical} }^{\text{dataA}}, R_ { \text {empirical} }^{\text{dataB}}) \ge D(G, R_ { \text {empirical} }^{\text{dataB}}) \).

Finally, and most importantly, triangle inequality provides valuable insights into the magnitude of metric distances. Consider the example where \(D(G_{\text{modelA}}, R_{\text{empirical}})=2D(G_{\text{modelB}}, R_{\text{empirical}})\). With the triangle inequality constraint, we can deduce that model A's performance is at least twice as suboptimal as model B's. Without this constraint, we could only conclude that model A underperforms model B, but without a precise measure of the extent of this underperformance.

%% file: sections/appendix/non_gaussianity_evidences.tex
\section{Non-Gaussian Characteristics of I3D Features\label{sec:non_gaussian}}
\hyperlink{vmae-jpepa-extra-results}{\house} Back to paper
\subsection{Visualization in Lower Dimensionality}
\begin{figure}[ht]
    \setlength\tabcolsep{3pt} % default: 6pt
    \centering
    \begin{tabular}{@{} c M{0.15\linewidth} M{0.15\linewidth} M{0.15\linewidth} M{0.15\linewidth} | M{0.15\linewidth} M{0.15\linewidth}}
    \textbf{\# F.} & \textbf{BDD} & \textbf{HMDB} & \textbf{Sky} & \textbf{UCF} & \textbf{HMDB} & \textbf{UCF}\\
    16 & \includegraphics[width=\hsize]{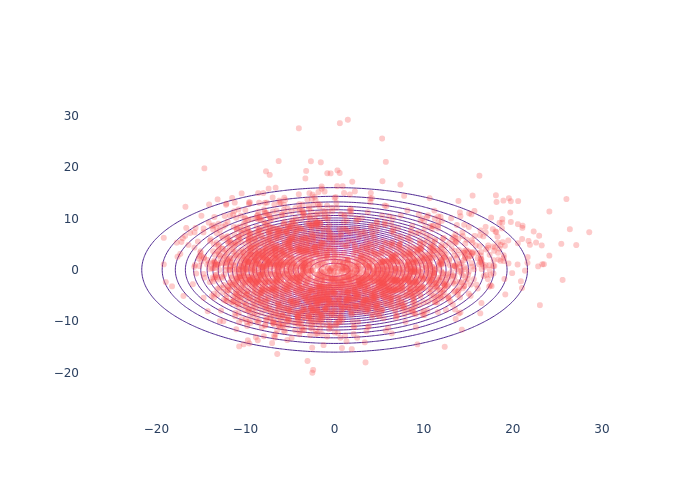}   
      & \includegraphics[width=\hsize]{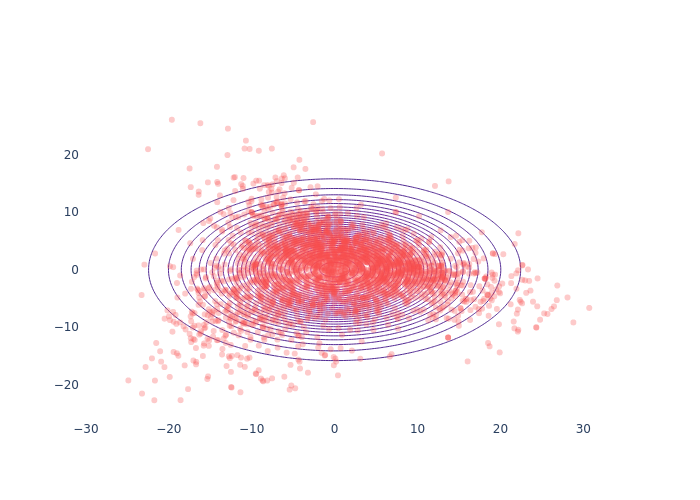}
      & \includegraphics[width=\hsize]{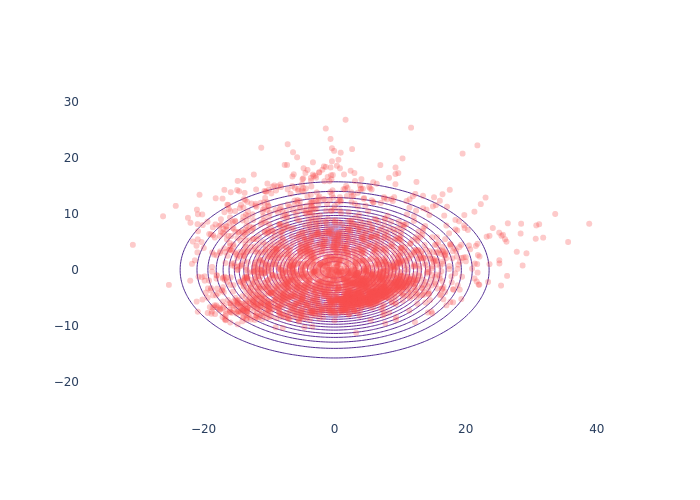}
      & \includegraphics[width=\hsize]{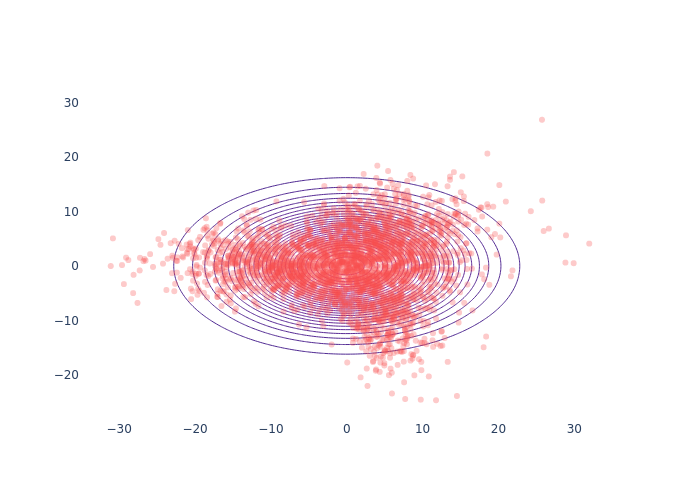}
      & \includegraphics[width=\hsize]{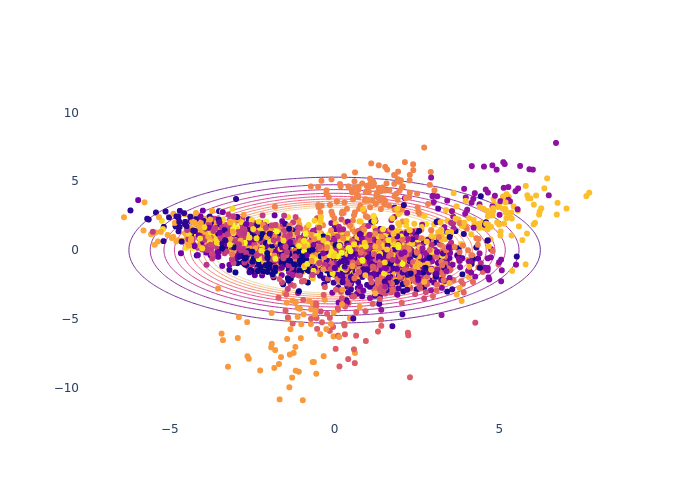}
      & \includegraphics[width=\hsize]{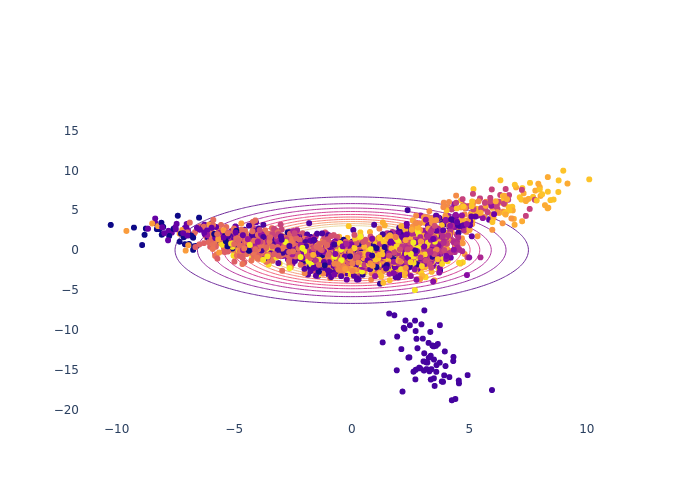}\\
    32 & \includegraphics[width=\hsize]{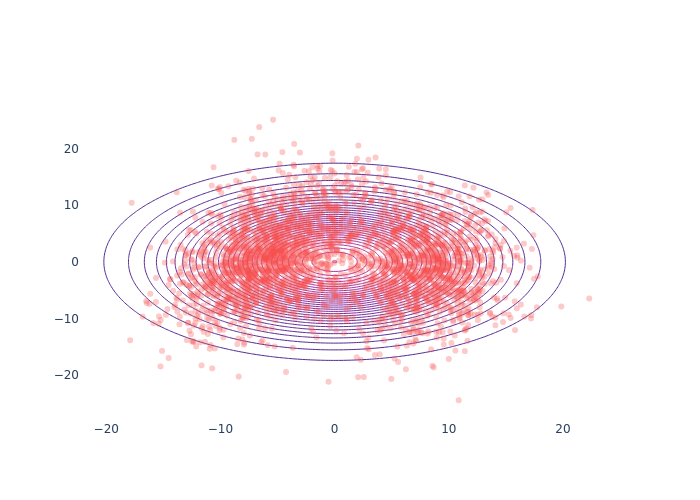}    
      & \includegraphics[width=\hsize]{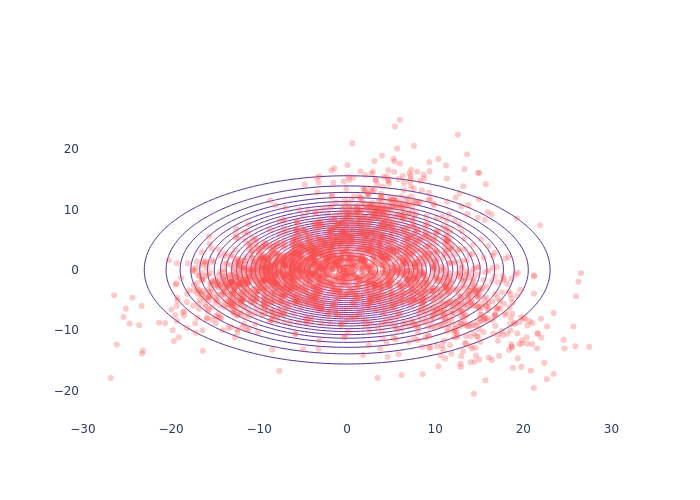}
      & \includegraphics[width=\hsize]{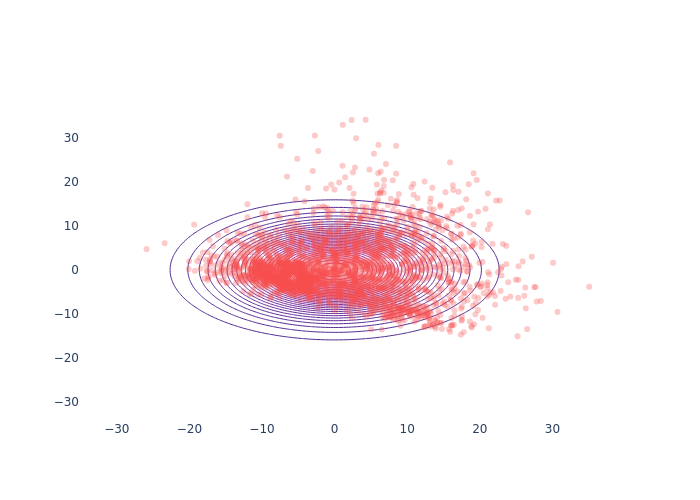}
      & \includegraphics[width=\hsize]{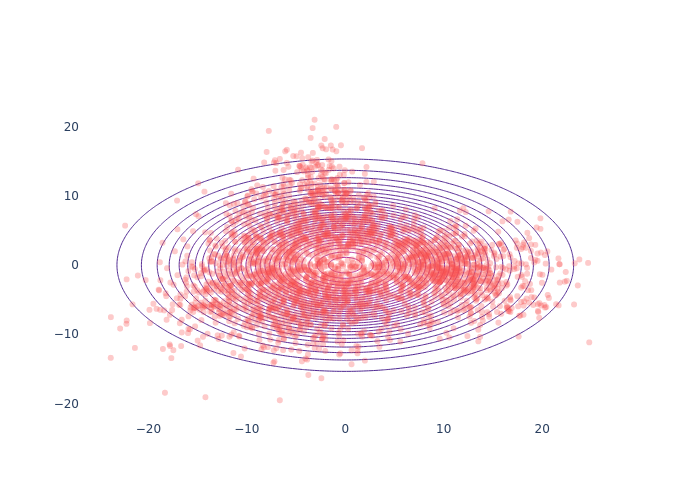}
      & \includegraphics[width=\hsize]{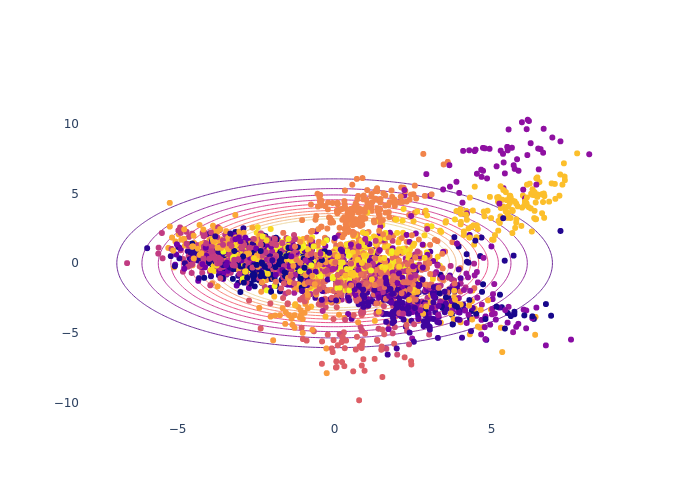}
      & \includegraphics[width=\hsize]{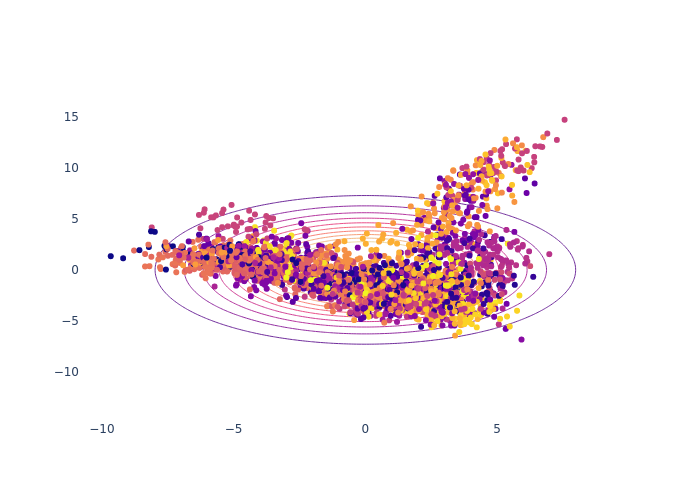}\\
    64 & \includegraphics[width=\hsize]{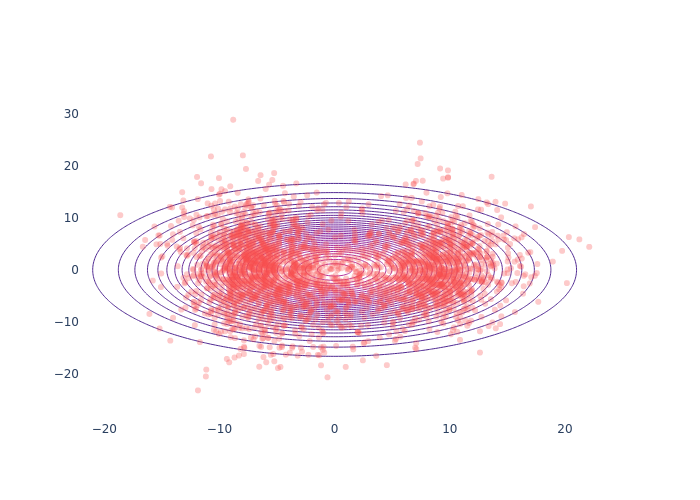}    
      & \includegraphics[width=\hsize]{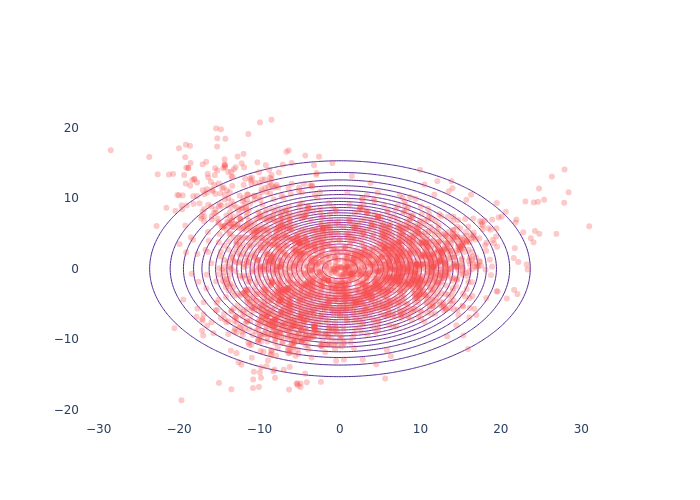}
      & \includegraphics[width=\hsize]{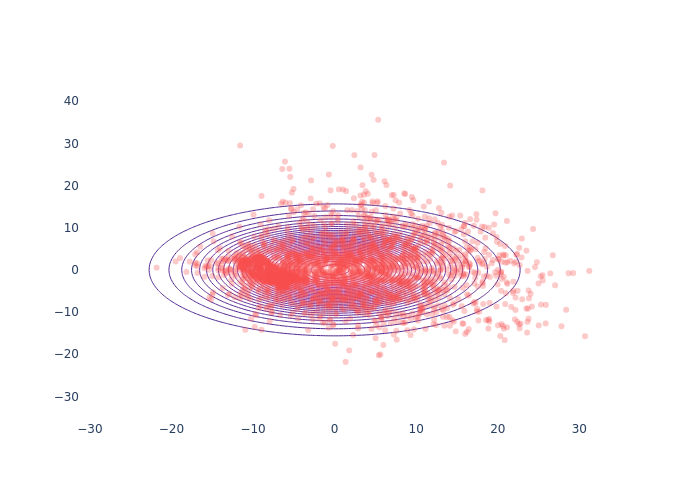}
      & \includegraphics[width=\hsize]{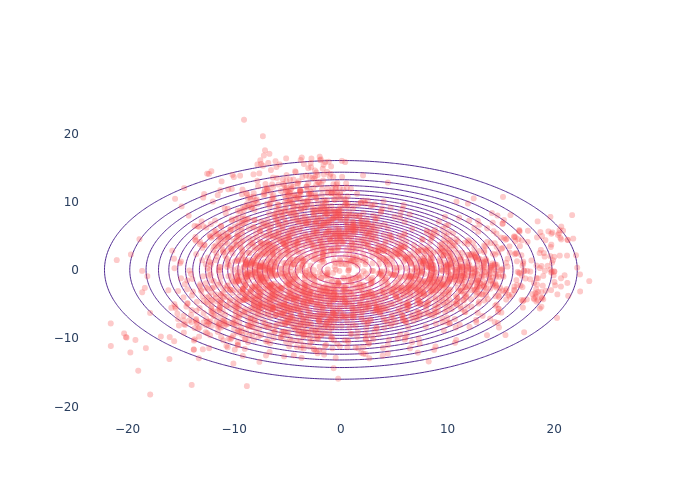}
      & \includegraphics[width=\hsize]{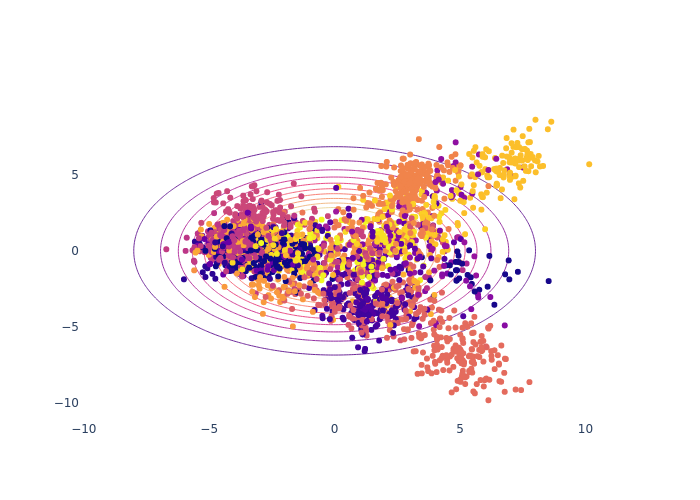}
      & \includegraphics[width=\hsize]{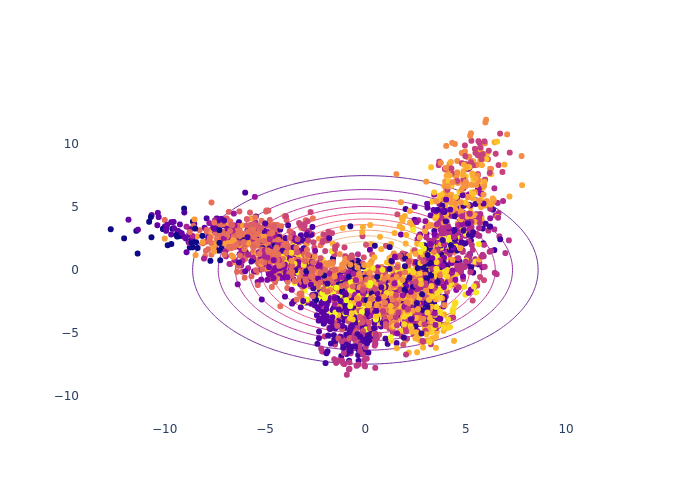}\\
    128 & \includegraphics[width=\hsize]{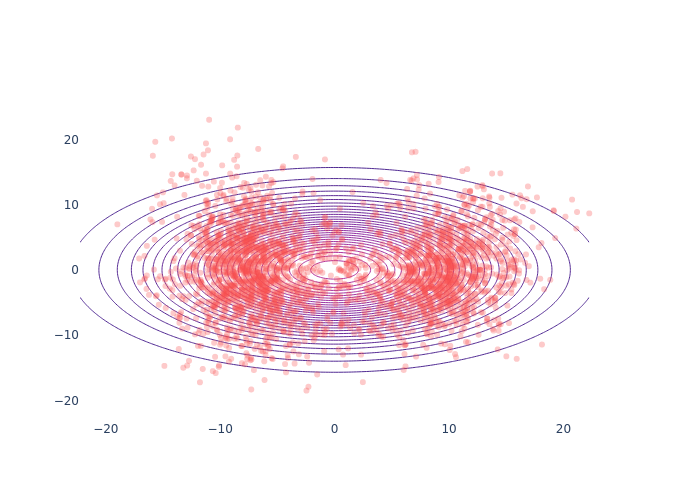}   
      & \includegraphics[width=\hsize]{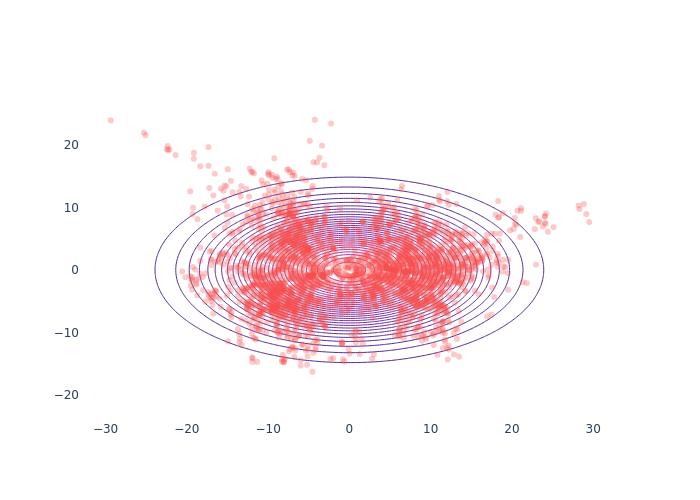}
      & \includegraphics[width=\hsize]{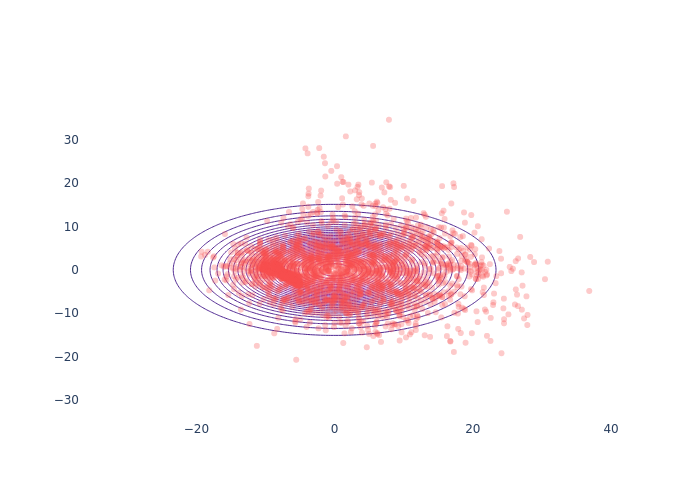}
      & \includegraphics[width=\hsize]{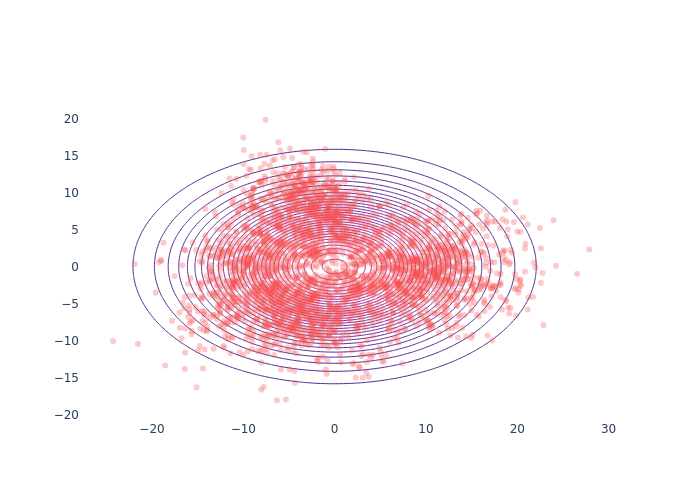}
      & \includegraphics[width=\hsize]{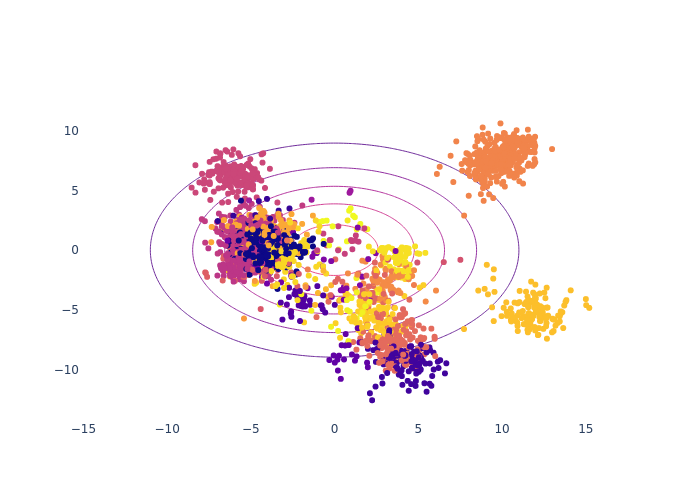}
      & \includegraphics[width=\hsize]{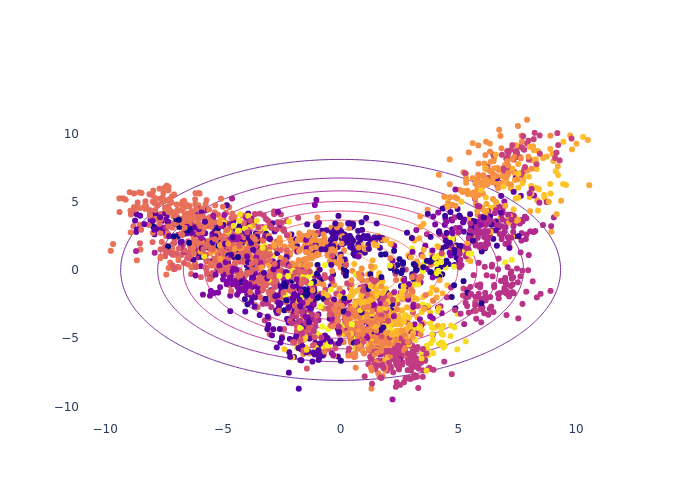}\\
    \end{tabular}
      \caption{This figure provides empirical evidence of non-Gaussianity in I3D feature space across individual datasets.}
      \label{fig:pca_individual}
      \raggedright \hyperlink{vmae-jpepa-extra-results}{\house} Back to paper
\end{figure}

Figure~\ref{fig:pca_individual} displays 2D PCA and LDA projections of I3D features, highlighting non-Gaussian characteristics within individual datasets. The numbers listed in the first column represent the quantity of frames per clip utilized for the experiments in the respective row. In each experiment, we randomly selected 5,000 video clips from each dataset and obtained their I3D features. We performed principal component analysis on each set of 5,000 samples, and the red point clouds in columns 2-5 represent the I3D features projected onto their first 2 principal components. 

\begin{figure}[ht]
    \setlength\tabcolsep{3pt} % default: 6pt
    \centering
    \begin{tabular}{@{} c M{0.15\linewidth} M{0.15\linewidth} M{0.15\linewidth} M{0.15\linewidth} | M{0.15\linewidth} M{0.15\linewidth}}
    \textbf{\# F.} & \textbf{BDD} & \textbf{HMDB} & \textbf{Sky} & \textbf{UCF} & \textbf{HMDB} & \textbf{UCF}\\
    16 & \includegraphics[width=\hsize]{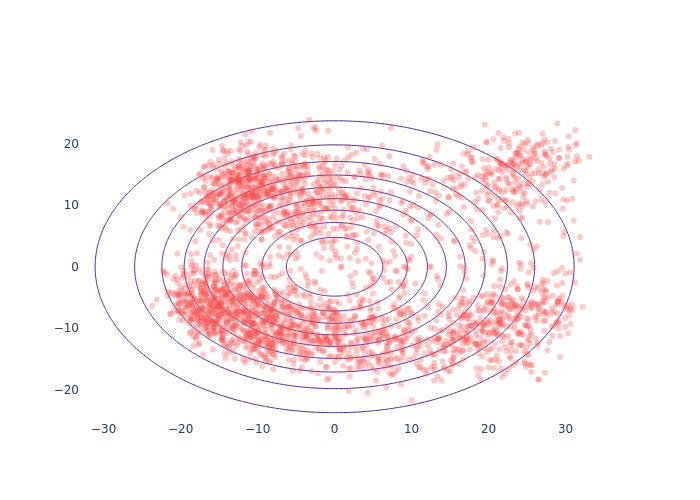}   
      & \includegraphics[width=\hsize]{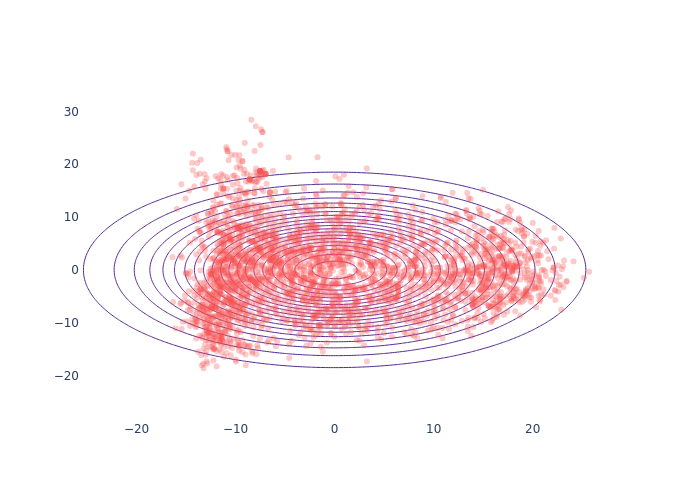}
      & \includegraphics[width=\hsize]{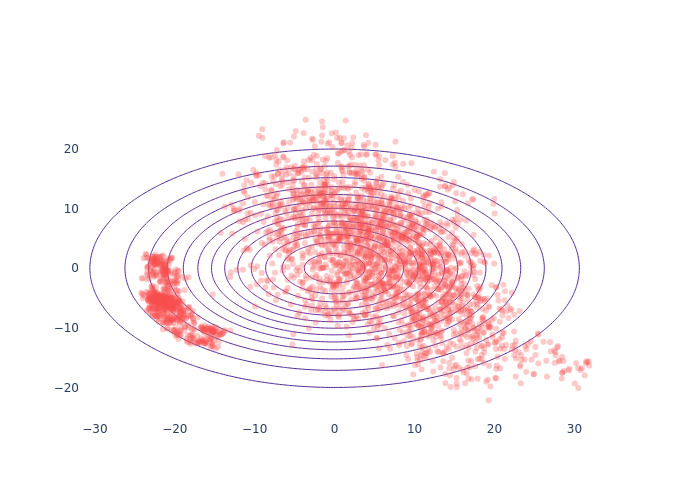}
      & \includegraphics[width=\hsize]{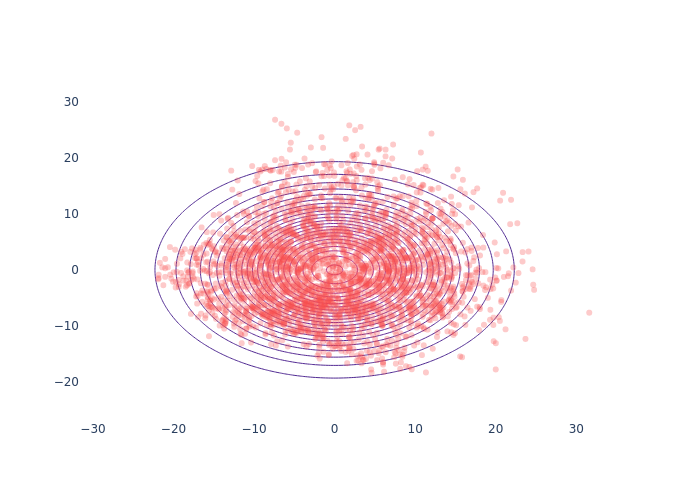}
      & \includegraphics[width=\hsize]{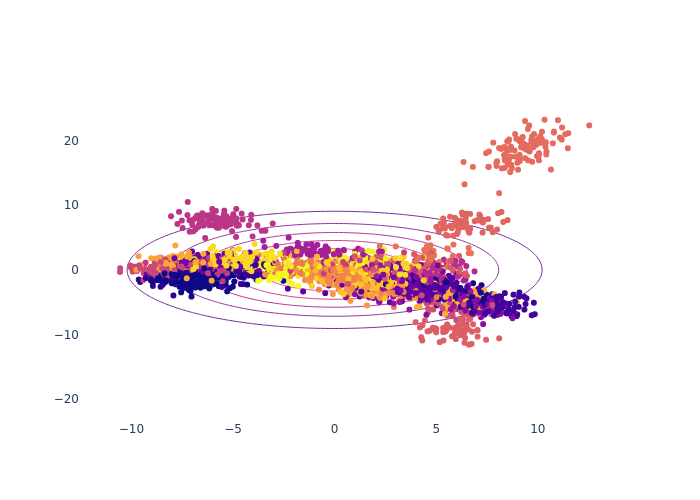}
      & \includegraphics[width=\hsize]{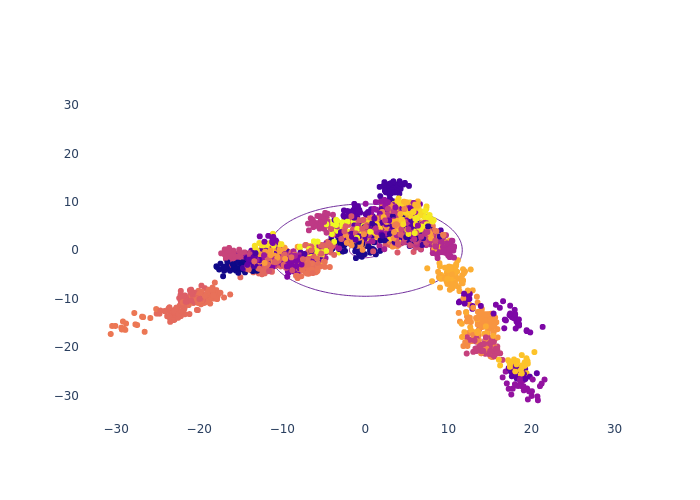}\\
      32 & \includegraphics[width=\hsize]{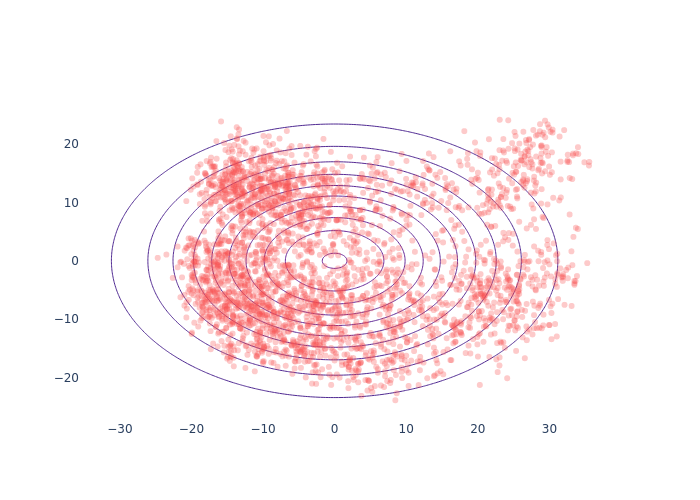}   
      & \includegraphics[width=\hsize]{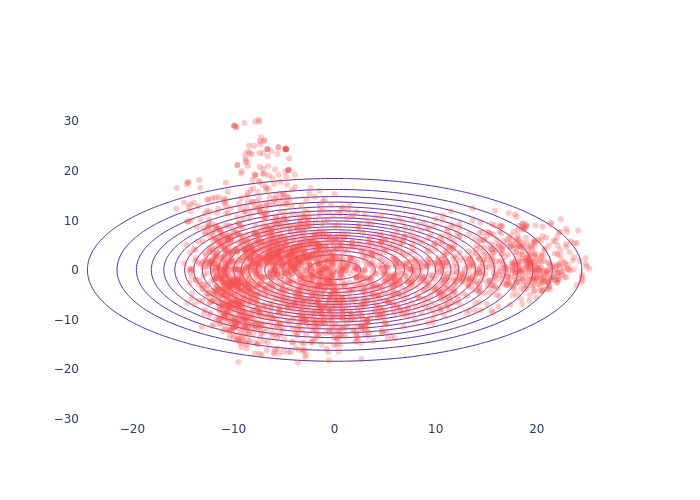}
      & \includegraphics[width=\hsize]{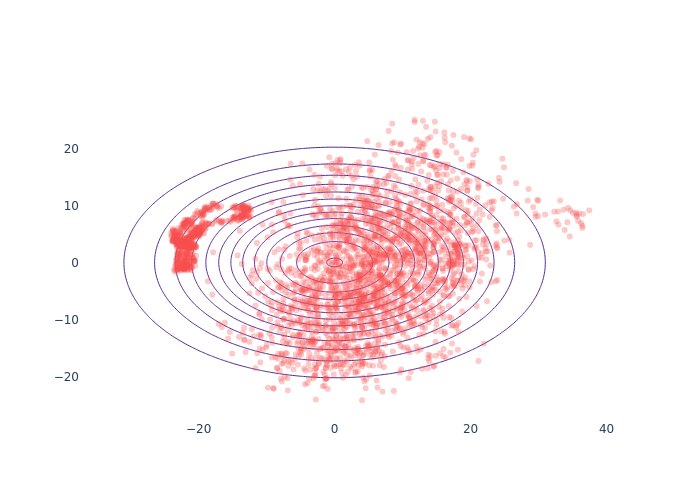}
      & \includegraphics[width=\hsize]{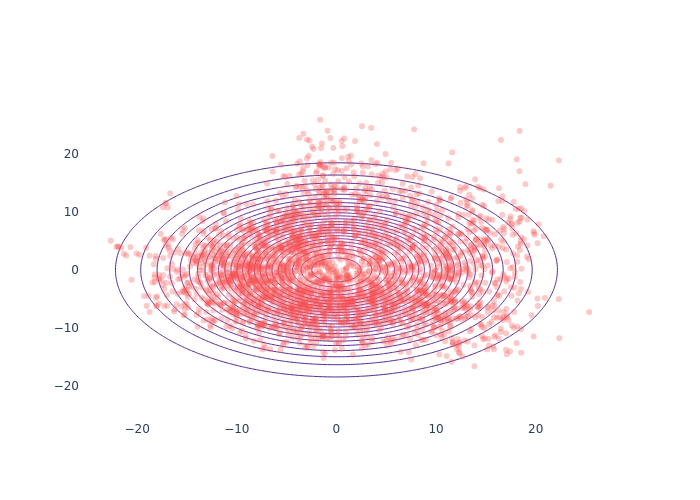}
      & \includegraphics[width=\hsize]{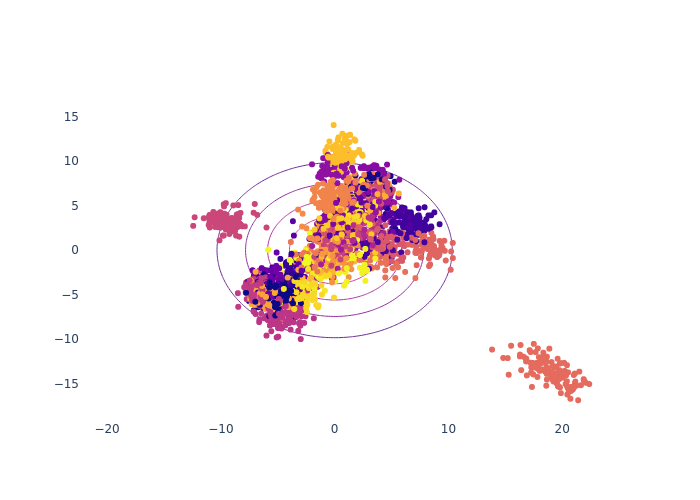}
      & \includegraphics[width=\hsize]{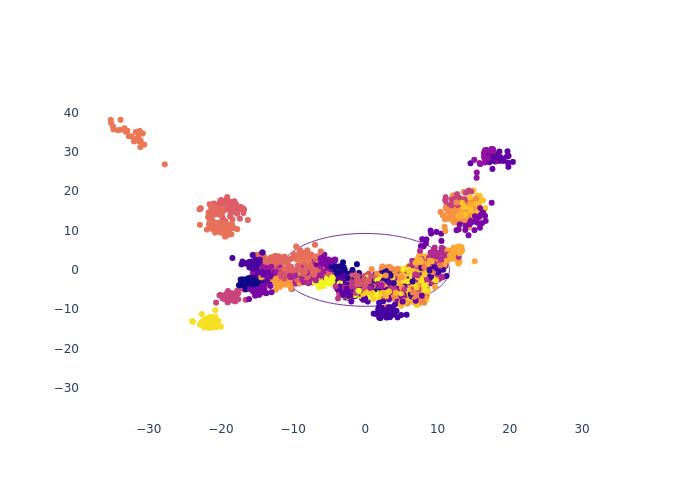}\\
      \end{tabular}
      \caption{This figure provides empirical evidence of non-Gaussianity in \vjepapt~ feature space across individual datasets.}
      \label{fig:pca_individual_vjepa_pt}
      \raggedright \hyperlink{vmae-jpepa-extra-results}{\house} Back to paper
\end{figure}

\begin{figure}[ht]
    \setlength\tabcolsep{3pt} % default: 6pt
    \centering
    \begin{tabular}{@{} c M{0.15\linewidth} M{0.15\linewidth} M{0.15\linewidth} M{0.15\linewidth} | M{0.15\linewidth} M{0.15\linewidth}}
    \textbf{\# F.} & \textbf{BDD} & \textbf{HMDB} & \textbf{Sky} & \textbf{UCF} & \textbf{HMDB} & \textbf{UCF}\\
    16 & \includegraphics[width=\hsize]{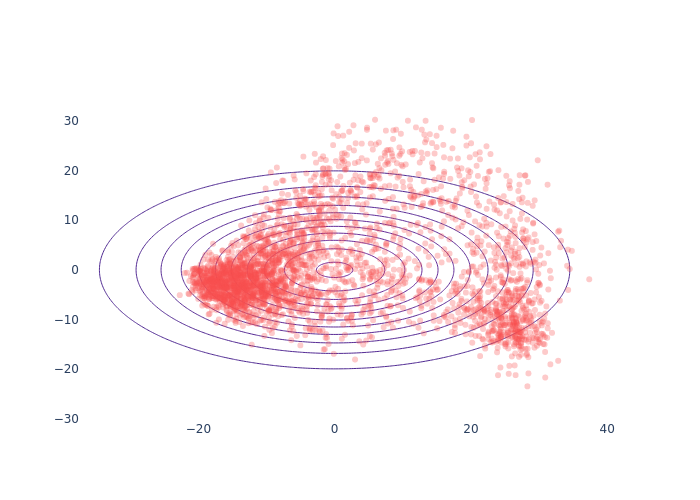}   
      & \includegraphics[width=\hsize]{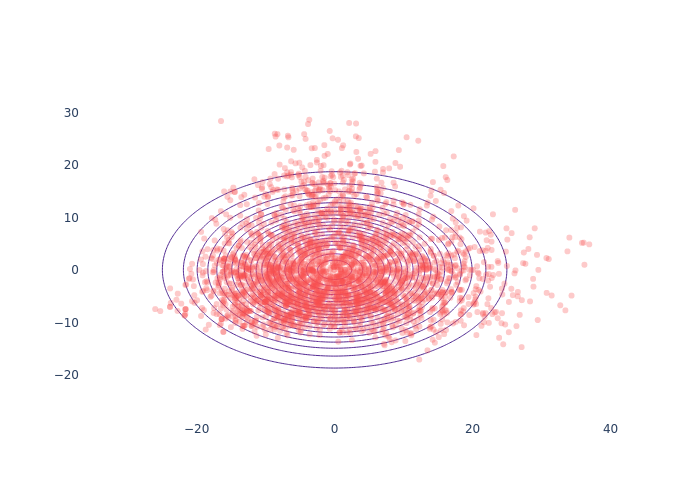}
      & \includegraphics[width=\hsize]{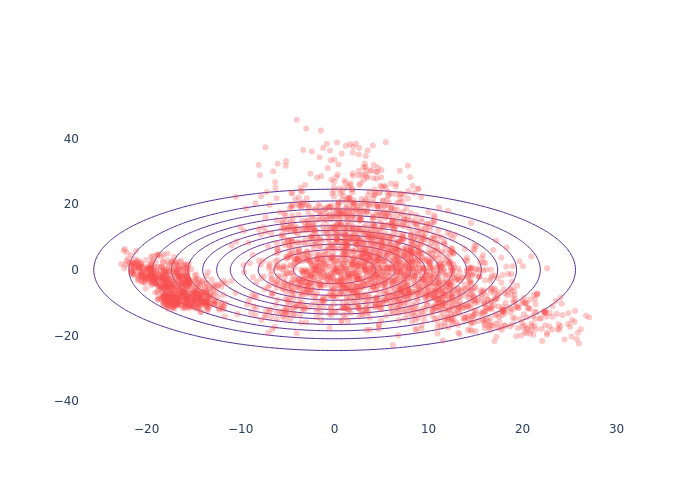}
      & \includegraphics[width=\hsize]{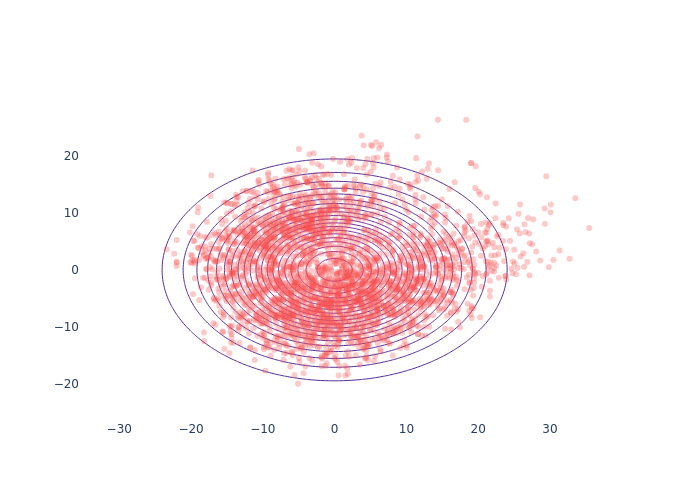}
      & \includegraphics[width=\hsize]{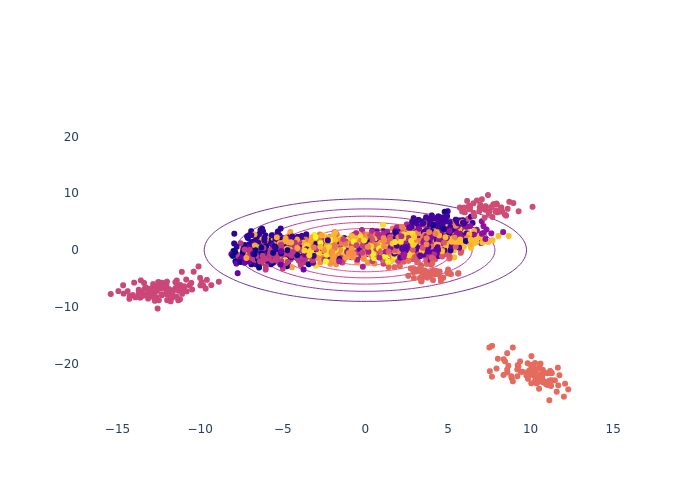}
      & \includegraphics[width=\hsize]{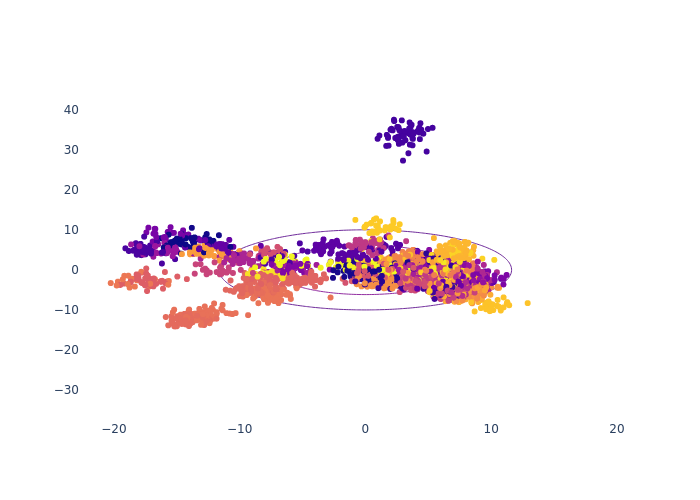}\\
      32 & \includegraphics[width=\hsize]{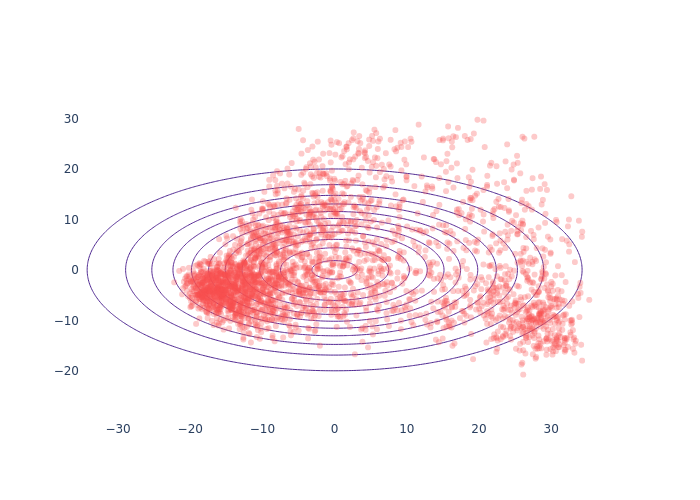}   
      & \includegraphics[width=\hsize]{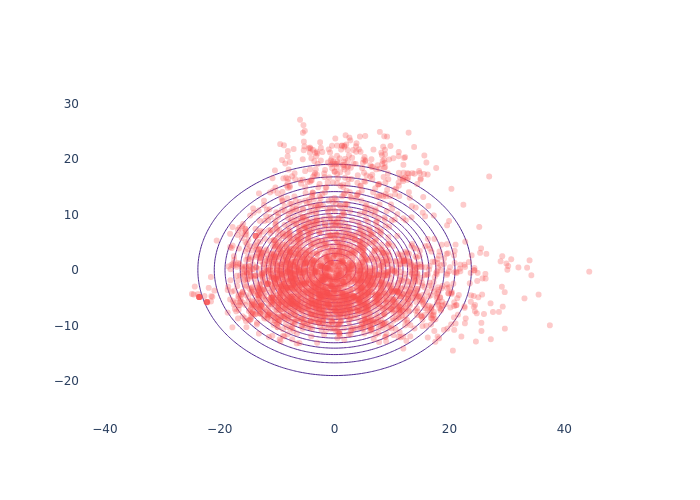}
      & \includegraphics[width=\hsize]{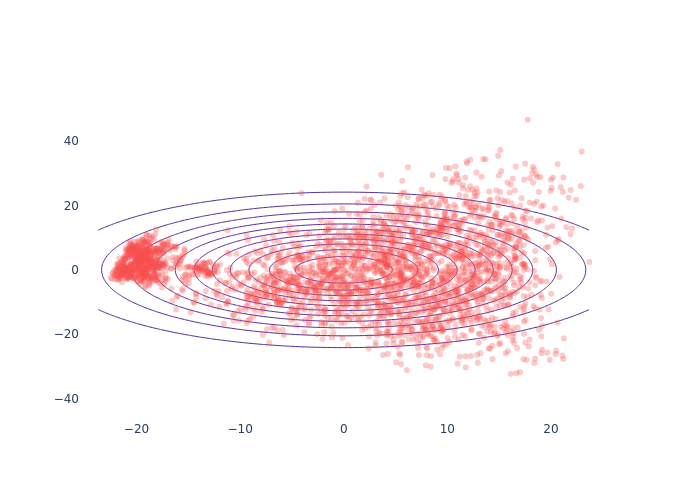}
      & \includegraphics[width=\hsize]{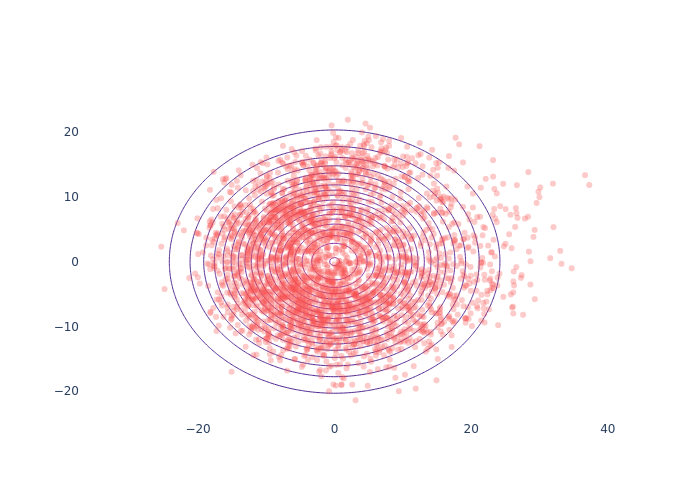}
      & \includegraphics[width=\hsize]{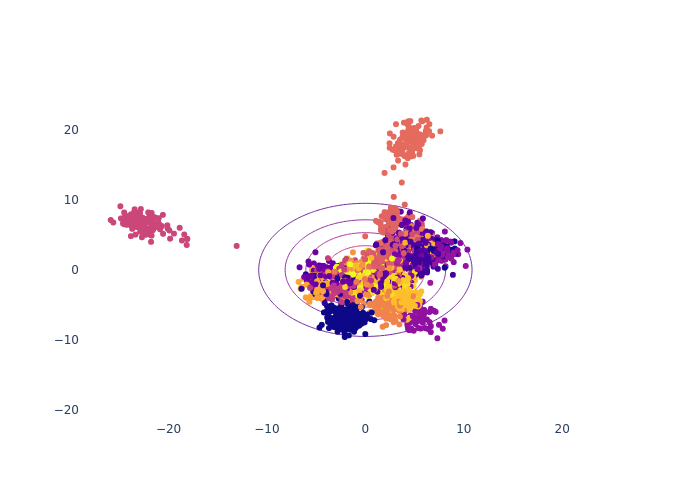}
      & \includegraphics[width=\hsize]{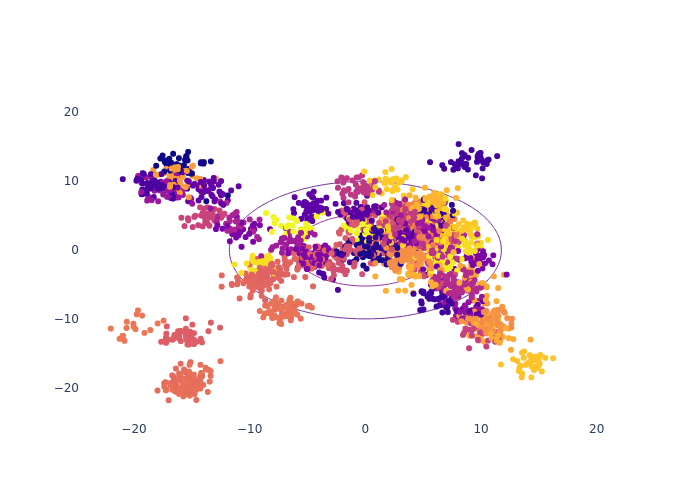}\\
      \end{tabular}
      \caption{This figure provides empirical evidence of non-Gaussianity in \vjepaft~ feature space across individual datasets.}
      \label{fig:pca_individual_vjepa_ft}
\end{figure}

The point clouds displayed in the last two columns depict the LDA dimensionally-reduced I3D features using the classification labels from the dataset. The markers' colors represent the class labels in the dataset. In each figure, we superimpose blue contour lines on the plot, which delineate the ellipsoidal contours of a 2D Gaussian distribution. Specifically, these contours are computed directly from the mean and covariance matrices of the 2D point clouds, providing a visual representation of the multivariate Gaussian distribution suggested by the Fr\'echet Video Distance (FVD) metric. We conducted a similar analysis for \vjepapt~and \vjepaft~features, as shown in Figure \ref{fig:pca_individual_vjepa_pt} and Figure \ref{fig:pca_individual_vjepa_ft}.

\subsection{Multivariate Normality Tests}
The notion that I3D video features do not follow multivariate Gaussian distributions was further investigated using Mardia's Skewness~\citep{mardia_skewness}, Mardia's Kurtosis~\citep{mardia_skewness}, and the Henze-Zirkler~\citep{hz_test} normality tests, following~\citep{jayasumana2024rethinkingFID}. The analysis was done on 16-, 32-, and 128-frame versions of the Anime-Run-v2, BAIR, BDD100k, DAVIS, Fashion Modeling, HMDB-51, How2Sign, KITTI, Something-Something-v2, Sky Scene~\citep{xiong2018skyscene}, and UCF-101 datasets, as well as on a baseline dataset constructed by sampling from a multivariate Gaussian distribution with 100 features.

The null hypothesis that the distributions of I3D features were drawn from a multivariate Gaussian distribution was rejected for each of the datasets and normality tests. All three tests accepted the null hypothesis for the multivariate Gaussian baseline dataset.

%% file: sections/appendix/toy_experiment_and_convergence.tex
\section{Experimental Evaluation: Convergence Rates of Distributional Metrics and Dimensionality Reduction Methods}

\begin{figure}[!ht]%
    \setlength\tabcolsep{3pt} % default: 6pt
    \centering
    \begin{tabular}{c M{0.14\linewidth} M{0.14\linewidth} M{0.14\linewidth}M{0.14\linewidth} M{0.14\linewidth} M{0.14\linewidth}}
     & \textbf{FD} & \textbf{Energy} & \textbf{$\mw$} & \textbf{$\text{MMD}_{\text{RBF}}$} & \textbf{$\text{MMD}_{\text{Polynomial}}$ }& \textbf{$\text{MMD}_{\text{Laplacian}}$}\\
    MG & \includegraphics[width=\hsize]{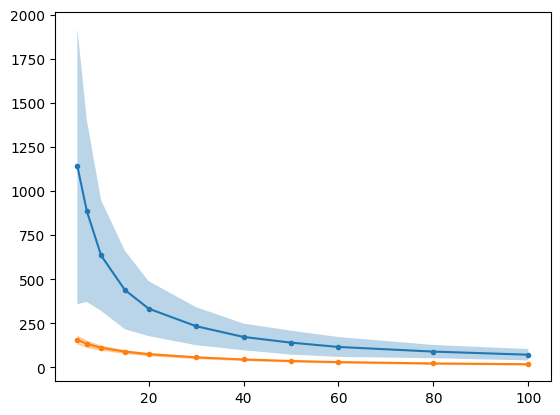}   
      & \includegraphics[width=\hsize]{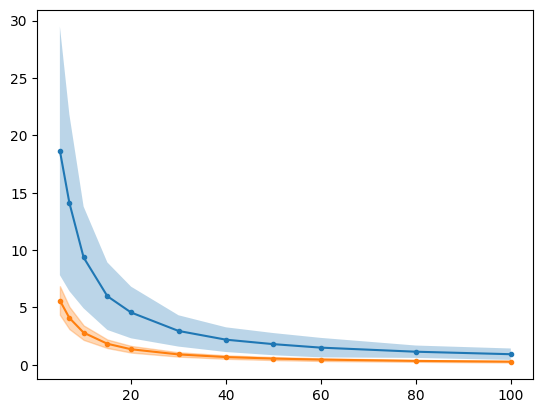}
      & \includegraphics[width=\hsize]{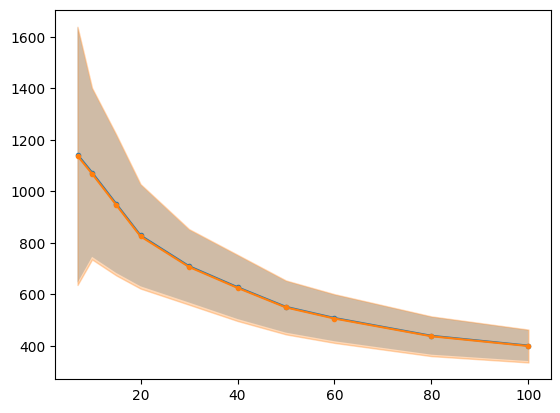}& \includegraphics[width=\hsize]{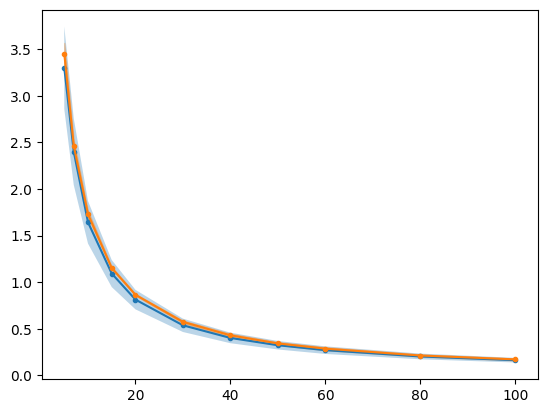}   
      & \includegraphics[width=\hsize]{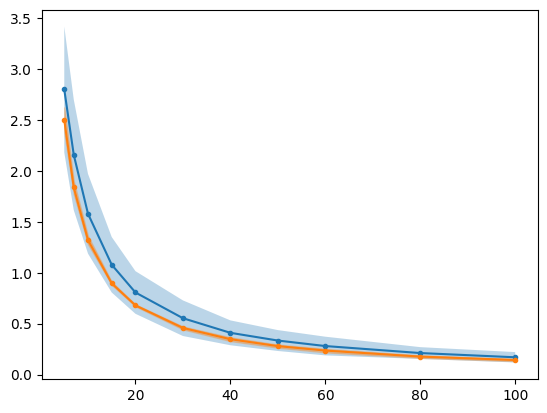}
      & \includegraphics[width=\hsize]{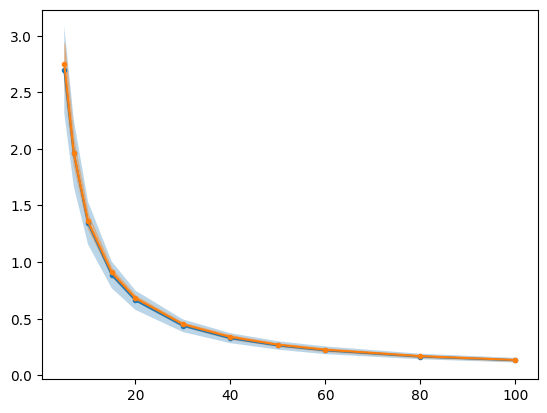}\\
      GMM & \includegraphics[width=\hsize]{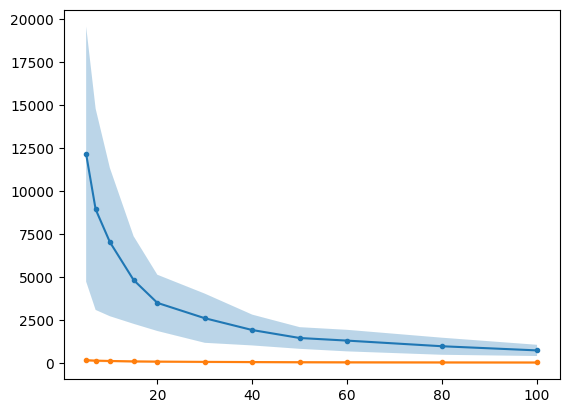}   
      & \includegraphics[width=\hsize]{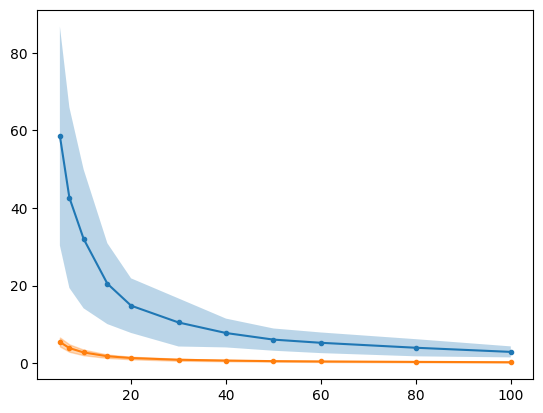}
      & \includegraphics[width=\hsize]{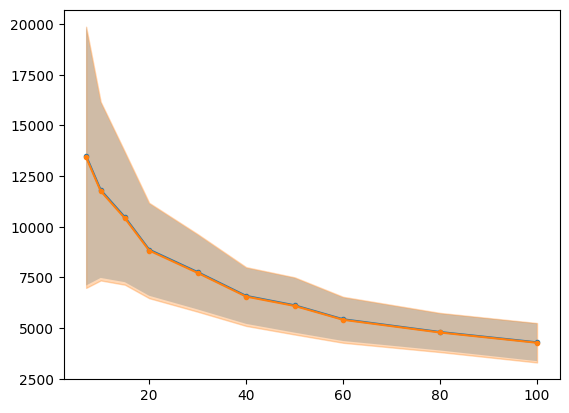}& \includegraphics[width=\hsize]{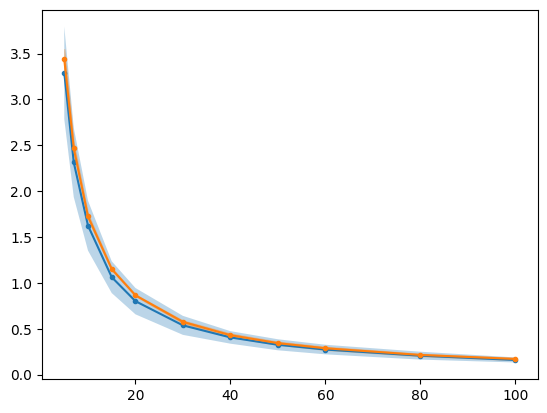}   
      & \includegraphics[width=\hsize]{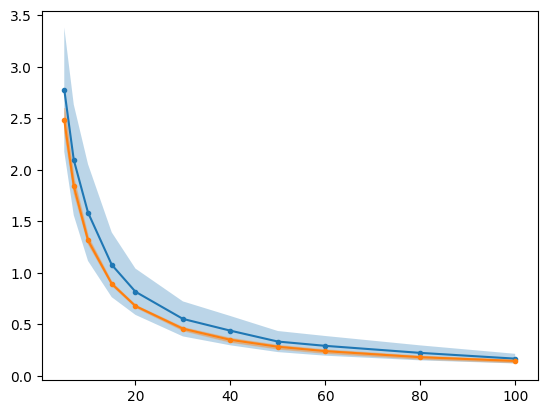}
      & \includegraphics[width=\hsize]{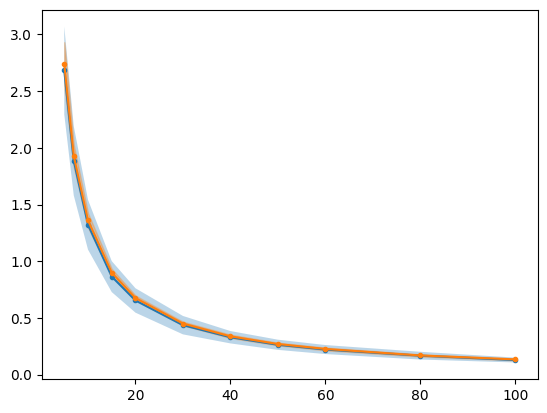}\\
      \end{tabular}
      \caption{The figures illustrate the evolution of distance estimates between two identical distributions as a function of sample size. The x-axes represent the number of samples drawn from each distribution, while the y-axes corresponds to the distance measurements. The plots in the top-row experiments employ a 100-dimensional multivariate Gaussian distribution, denoted as MG. In this distribution, the first 50 dimensions follow a standard normal distribution (\( \mu = \bm { 0 } , \Sigma =I \)), while the remaining 50 dimensions are generated as the cumulative sum of the first 50 dimensions, resulting in a structured correlation pattern. In contrast, the bottom-row plots utilize a 100-dimensional Gaussian mixture model comprising 5 clusters, mirroring the multivariate Gaussian setup, with the first 50 dimensions drawn from a GMM and the last 50 dimensions representing the cumulative sum of the first 50 dimensions. The \emph{blue} lines represent the metrics calculated using samples directly drawn from the original distributions, while the \emph{orange} lines represent the metrics computed using samples that have undergone PCA-based dimensionality reduction to 50 principal components. Since the compared distributions are identical, the ideal estimated distance between them should be zero.}
      \label{fig:toy_experiment}
      \raggedright \hyperlink{autoencoder}{\house} Back to paper
\end{figure}

\subsection{Toy Experiment: Metric Convergence Rate\label{sec:toy_experiment}}
\hyperlink{toy-experiment}{\house} Back to paper

This section reports on simulated experiments analyzing two key questions: (1) How do different metrics converge? and (2) What role does dimensionality play in shaping convergence rates?

Figure~\ref{fig:toy_experiment} illustrates the findings of a series of simulated experiments examining the empirical distribution distances across various metrics, uncovering the intricate interplay between sample sizes, dimensionality-reduced features, and metric convergence rates. The experimental results shown in the figures demonstrate two key findings: (1) the accuracy of the metrics improves consistently as the empirical sample size increases, validating the proof of concept; and (2) dimension-reduced data can significantly enhance the convergence rate and accuracy of certain metrics (notably FD and energy), as evidenced by the faster convergence of the orange curves relative to the blue curves.

Our simulation experiments also yielded the notable observation that kernel-based distance metrics for distributions exhibit reduced sensitivity to data dimensionality, attributable to their inherent feature mapping, reliance on pairwise distance calculations, and regularization properties inherent in kernel selection. This diminished sensitivity confers a significant advantage in mitigating the curse of dimensionality, thereby enhancing the robustness and reliability of distance metric estimates in higher dimensional space.

In contrast, GMMOT involves fitting separate Gaussian Mixture Models (GMMs) to the sample sets and then computing the optimal transport distance between the fitted GMMs. In the GMM fitting process, the Expectation-Maximization (EM) algorithm updates the parameters of the Gaussian mixtures. During the maximization step, cluster means and covariances are computed using the sample responsibilities calculated in the expectation step. Notably, the sum of the weighted responsibilities for each cluster is less than the total number of samples, but the number of parameters for each cluster is the same as a single multivariate Gaussian distribution. As a result, because of the smaller effective sample size, it is necessary to obtain more samples in order to accurately fit a GMM. Therefore, Mixture Wasserstein ($\mw$) has the slowest rate of convergence compared to all other metrics, leading us to exclude it from our subsequent video analysis.

\subsection{Autoencoders: Models and Training Specification\label{sec:ae_configuration}}
\hyperlink{autoencoder}{\house} Back to paper

In our study, we trained specialized autoencoders for each video representation space to account for variations in video length. Notably, we observed a representation shift with differing video lengths: as shown in Figures~\ref{fig:pca_individual} and ~\ref{fig:pca_individual_vjepa_pt}, features extracted from videos with different lengths have visible differences; features from 16f and 32f videos are more similar to each other while those from 64f and 128f videos are similar to each other. To address this,we divide the videos from 11 different datasets (UCF, HMDB, etc) into two groups: short clips (16–32 frames) and long clips (64–128 frames). For each group, we extract up to 10,000 features using five different feature extractors (e.g., I3D, \vjepapt). These results in separate collections of features for short and long clips across all datasets. For instance, I3D yields up to 110,000 short-clip features and up to 110,000 long-clip features from the 11 datasets. We then train two autoencoders for each feature extractor: one using the short-clip features and one using the long-clip features. This gives us a total of 10 autoencoders. The autoencoders' architectures are specified in Algorithm ~\ref{alg:ae_algorithm1} and Algorithm ~\ref{alg:ae_algorithm2}.
% \newpage
\RestyleAlgo{ruled}
\begin{minipage}{0.46\textwidth}
\IncMargin{1em}
\begin{algorithm}[H]
    \caption{I3D Autoencoder Configuration}\label{alg:ae_algorithm1}
    \footnotesize
    \DontPrintSemicolon
\SetKwInOut{Input}{input}
\Input{in\_dim=400}
\BlankLine
encoder = Sequential({\;
                \hskip1.0em Linear(in\_dim, in\_dim//2),\;
                \hskip1.0em ReLU(),\;
                \hskip1.0em Linear(in\_dim//2, in\_dim//4),\;
                \hskip1.0em ReLU(),\;
                \hskip1.0em Linear(in\_dim//4, in\_dim//6),\;}
            )
            
decoder = Sequential({\;
                \hskip1.0em Linear(in\_dim//6, in\_dim//4),\;
                \hskip1.0em ReLU(),\;
                \hskip1.0em Linear(in\_dim//4, in\_dim//2),\;
                \hskip1.0em ReLU(),\;
                \hskip1.0em Linear(in\_dim//2, in\_dim),\;}
            )
\end{algorithm}
\end{minipage}
\hfill
\begin{minipage}{0.46\textwidth}
\IncMargin{1em}
\begin{algorithm}[H]
    \caption{VideoMAE and V-JEPA Autoencoder Configuration}\label{alg:ae_algorithm2}
    \footnotesize
    \DontPrintSemicolon
\SetKwInOut{Input}{input}
\Input{in\_dim=1408 if VideoMAE else 1280}
\BlankLine
encoder = Sequential({\;
                \hskip1.0em Linear(in\_dim, in\_dim//3),\;
                \hskip1.0em ReLU(),\;
                \hskip1.0em Linear(in\_dim//3, in\_dim//4),\;
                \hskip1.0em ReLU(),\;
                \hskip1.0em Linear(in\_dim//4, in\_dim//8),\;}
            )
            
decoder = Sequential({\;
                \hskip1.0em Linear(in\_dim//8, in\_dim//4),\;
                \hskip1.0em ReLU(),\;
                \hskip1.0em Linear(in\_dim//4, in\_dim//3),\;
                \hskip1.0em ReLU(),\;
                \hskip1.0em Linear(in\_dim//3, in\_dim),\;}
            )
\end{algorithm}
\end{minipage}

\subsection{Sample Convergence Analysis~\label{sec:sample_convergence_analysis_appendix}}
\hyperlink{sample-efficiency}{\house} Back to paper

\begin{figure}[h!]%
    \setlength\tabcolsep{3pt} % default: 6pt
    \centering
    \begin{tabular}{c M{0.16\linewidth} M{0.16\linewidth} M{0.16\linewidth}M{0.16\linewidth} M{0.16\linewidth} }
     & \textbf{FD} & \textbf{Energy} & \textbf{$\text{MMD}_{\text{RBF}}$} & \textbf{$\text{MMD}_{\text{Polynomial}}$} & \textbf{$\text{MMD}_{\text{Laplacian}}$}\\
    I3D & \includegraphics[width=\hsize]{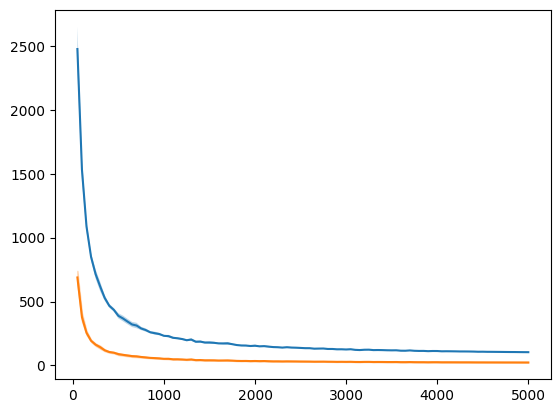}   
      & \includegraphics[width=\hsize]{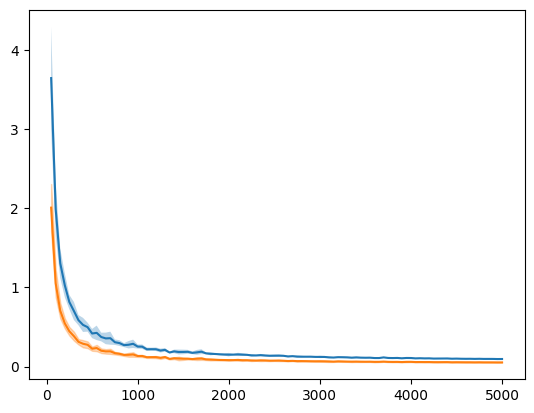}
      & \includegraphics[width=\hsize]{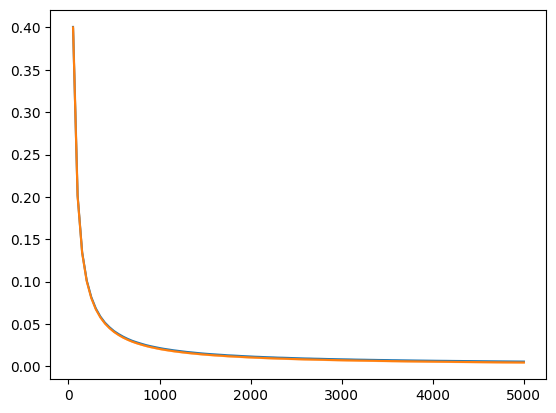}& \includegraphics[width=\hsize]{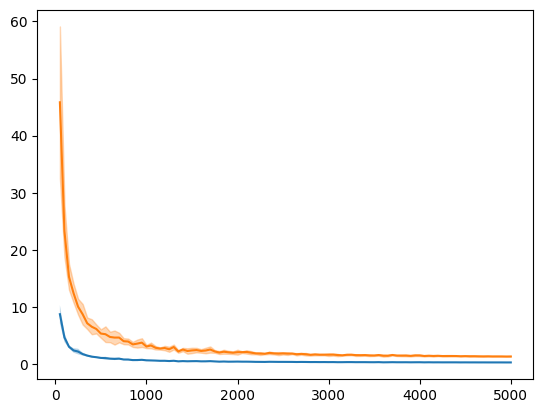}   
      & \includegraphics[width=\hsize]{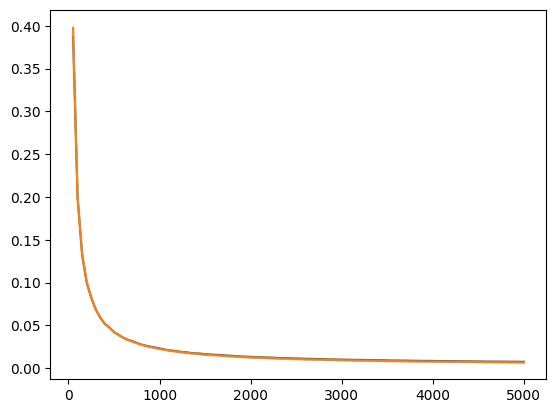}\\
      $\text{VMAE}_{\text{PT}}$ & \includegraphics[width=\hsize]{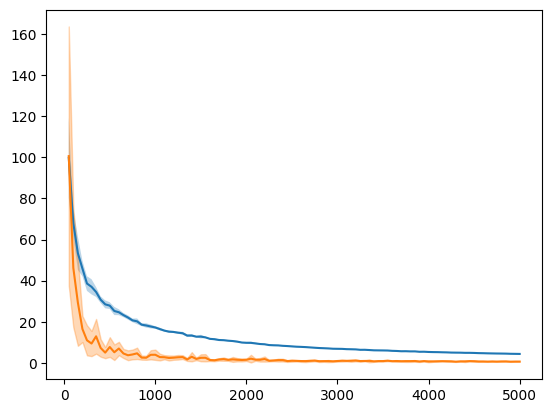}   
      & \includegraphics[width=\hsize]{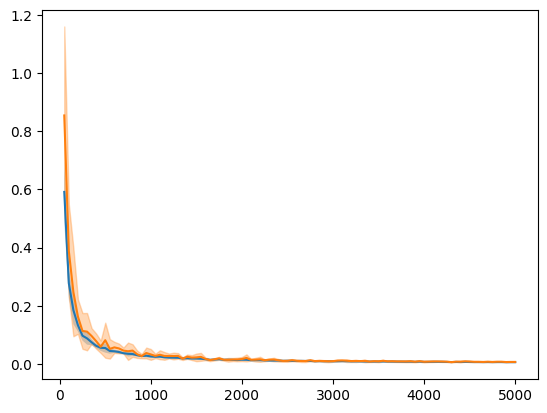}
      & \includegraphics[width=\hsize]{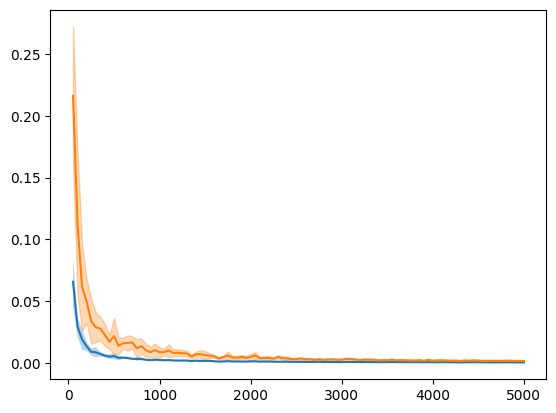}& \includegraphics[width=\hsize]{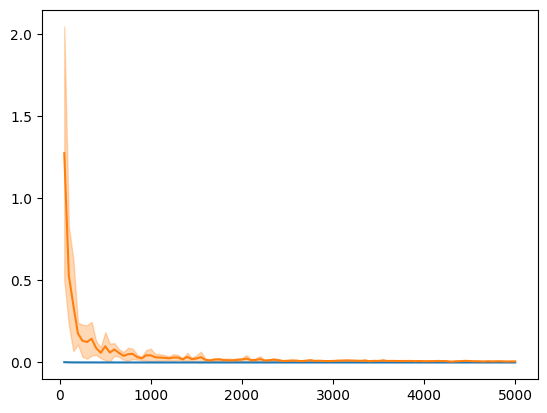}   
      & \includegraphics[width=\hsize]{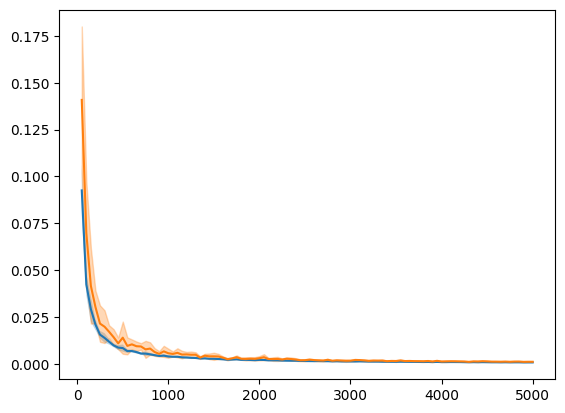}\\
      $\text{VMAE}_{\text{SSv2}}$ & \includegraphics[width=\hsize]{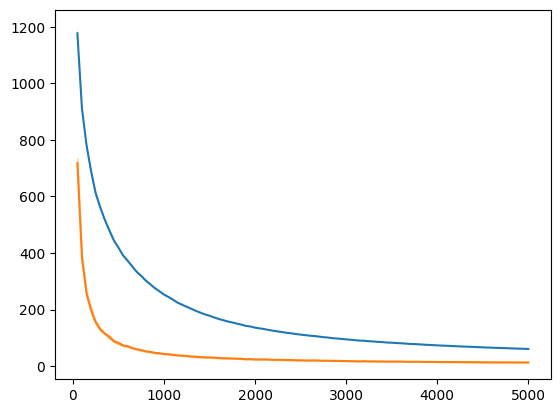}   
      & \includegraphics[width=\hsize]{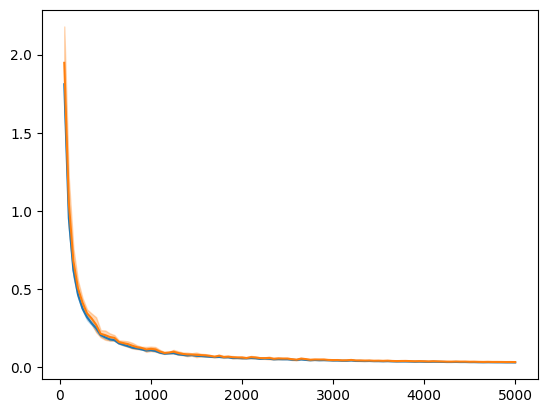}
      & \includegraphics[width=\hsize]{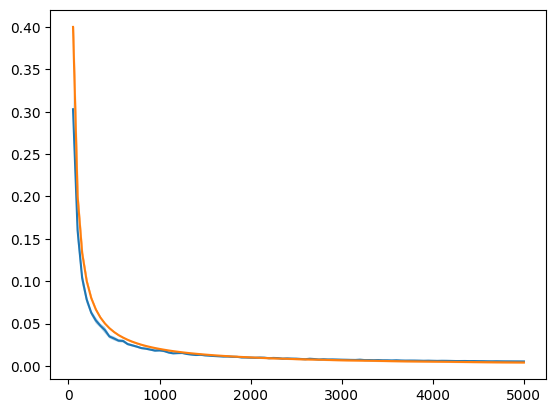}& \includegraphics[width=\hsize]{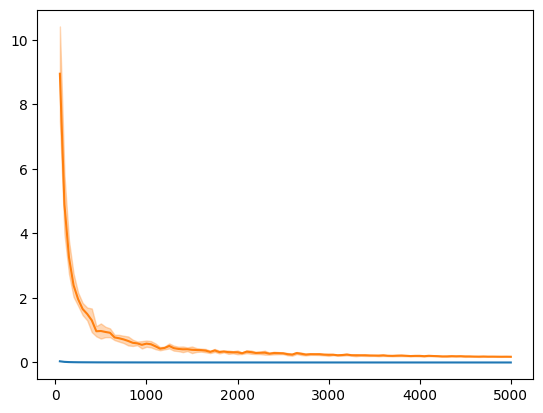}   
      & \includegraphics[width=\hsize]{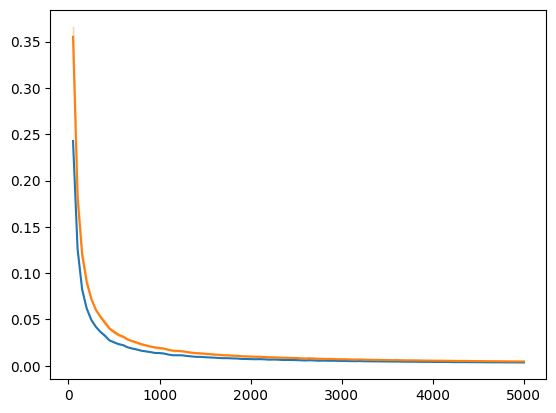}\\
       \vjepapt & \includegraphics[width=\hsize]{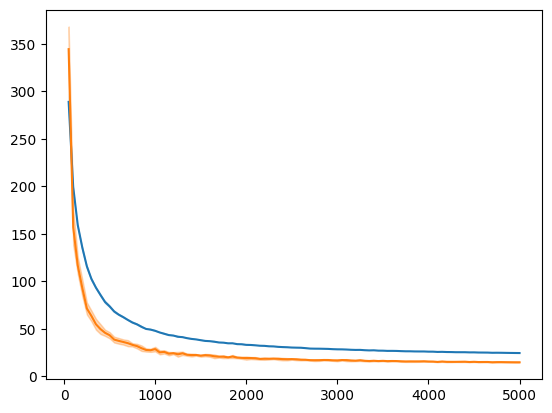}   
      & \includegraphics[width=\hsize]{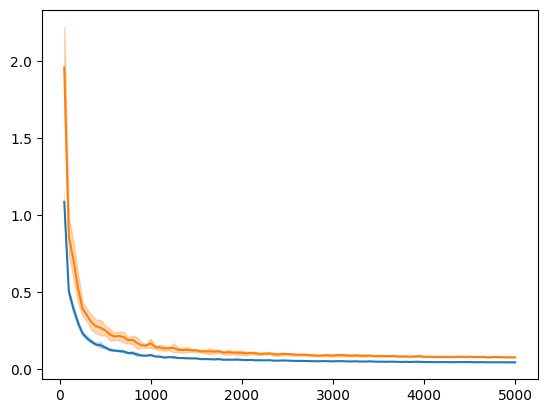}
      & \includegraphics[width=\hsize]{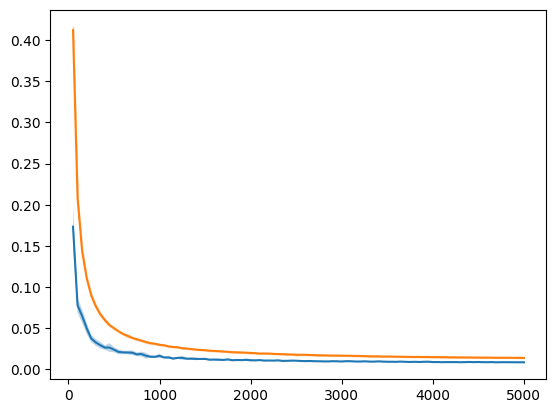}& \includegraphics[width=\hsize]{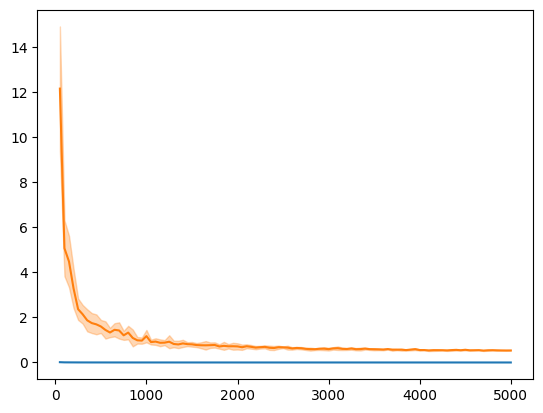}   
      & \includegraphics[width=\hsize]{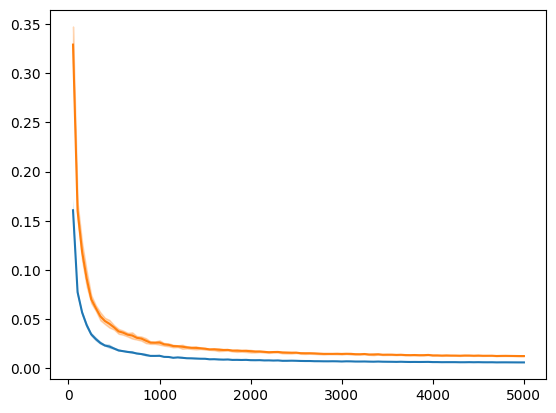}\\
      \vjepaft & \includegraphics[width=\hsize]{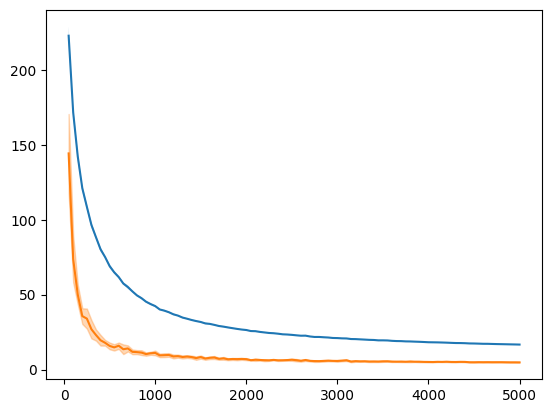}   
      & \includegraphics[width=\hsize]{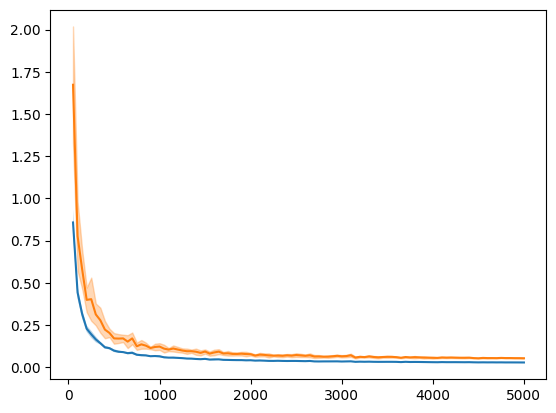}
      & \includegraphics[width=\hsize]{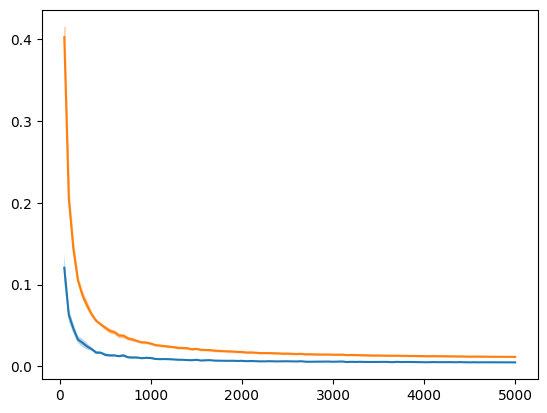}& \includegraphics[width=\hsize]{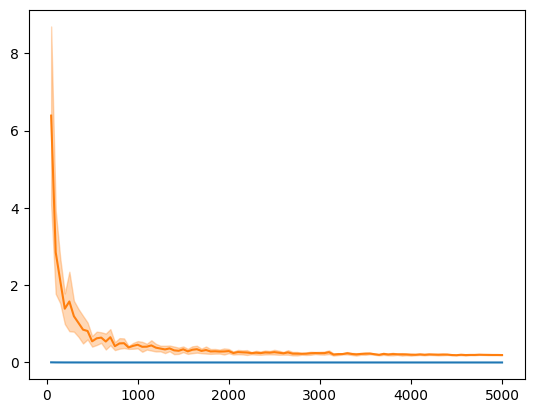}   
      & \includegraphics[width=\hsize]{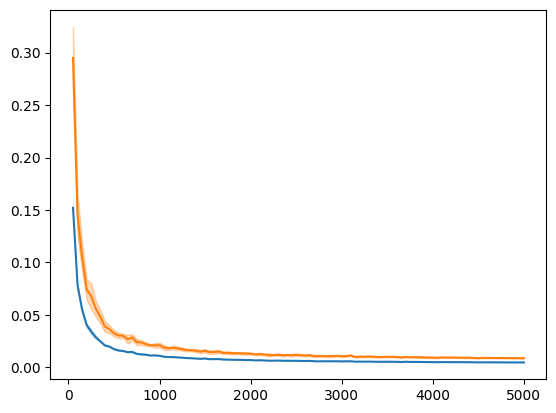}\\
      \end{tabular}
      \caption{The figures illustrate the evolution of distance estimates between the training and testing sets of the UCF-101 dataset. We extract features from video clips of 32-frame duration using the models indicated in the left-most column. The x-axes represent the number of samples drawn from each distribution, while the y-axes corresponds to the distance measurements. We repeat each experiment 10 times. The lighter shaded area on the plots indicate the variance across these 10 runs. The \emph{blue} lines represent the metrics calculated using features directly extracted from the models, while the \emph{orange} lines represent the metrics computed using features that have been compressed by the autoencoder's encoder. The autoencoder's structure is described in Appendix~\ref{sec:ae_configuration}.}
      \label{fig:train_test_distances}
      \raggedright \hyperlink{autoencoder}{\house} Back to paper
\end{figure}

\begin{figure}[h!]%
    \centering
    \vspace{-.5cm}
    \subfloat[UCF-101]{{\includegraphics[width=0.35\linewidth]{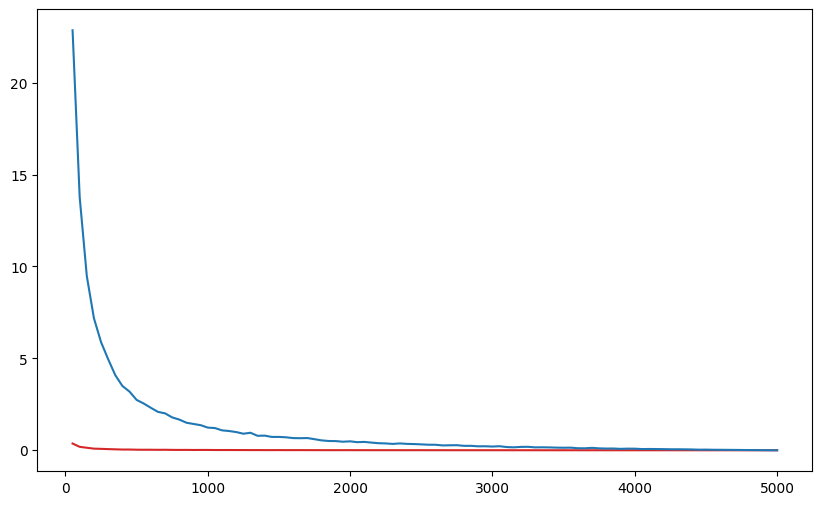} }}%
     \subfloat[SSv2]{{\includegraphics[width=0.35\linewidth]{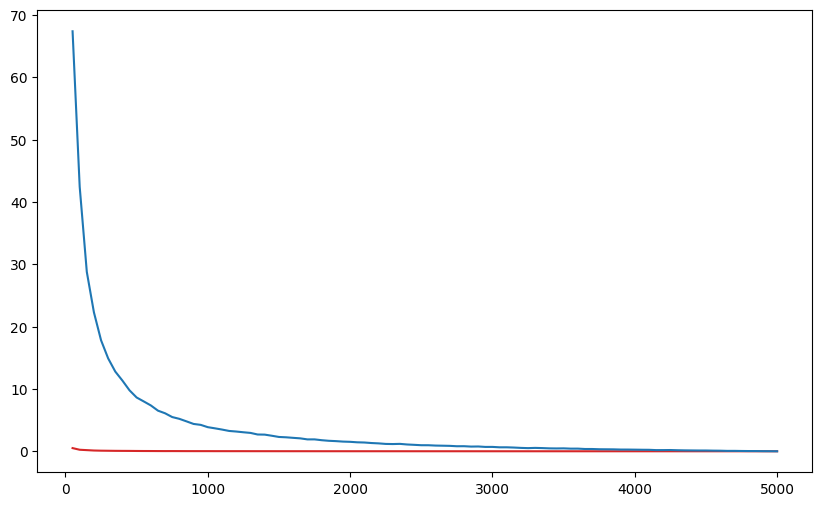} }}%
    \subfloat{\includegraphics[width=0.12\linewidth]{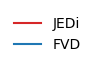}}%
    
    \caption{A comparison of convergence rates of FVD and \ourmetric, comparing training and testing sets on UCF101 and Something-Somethingv2. We evaluate convergence rate at 100 intervals, from 50 to 5,000 samples, with 50-sample increments. The x-axes represent the number of samples drawn from the training and test distributions, while the y-axes show the convergence rate, calculated as \(\frac{\bar{D}_{\text{m}}(n)-\bar{D}_{\text{m}}(5000)}{\bar{D}_{\text{m}}(5000)+\epsilon}\) where \(\epsilon\) is an arbitrarily small number and m is drawn from a set of metrics $\mathcal{M}$. Our methodology involves sampling \(n\) samples from the training and testing sets 10 times, computing the distance 10 times for each sampled sets, and calculating \(\bar{D}(n)\) as the mean distance across the 10 runs.}
    \label{fig:train_test_convergence_rate}
    
    \raggedright \hyperlink{autoencoder}{\house} Back to paper
\end{figure}

\begin{figure}[h!]%
    \centering
    \subfloat[I3D]{{\includegraphics[width=0.32\linewidth]{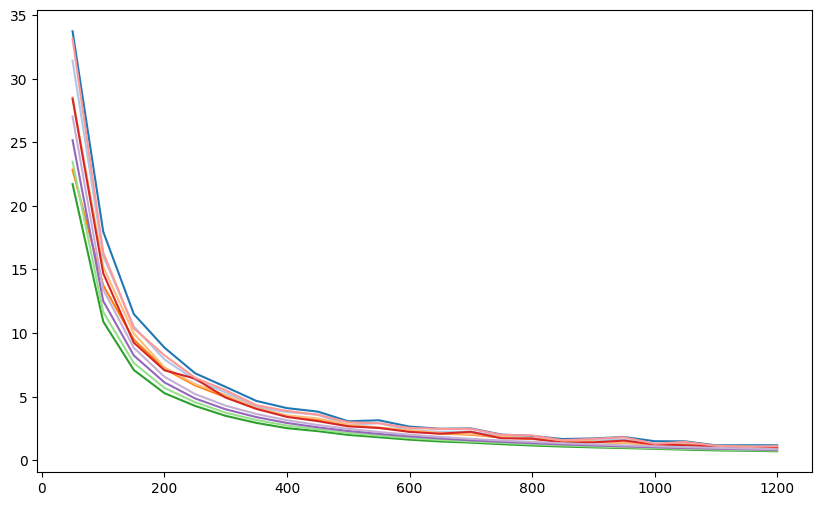} }}%
    \subfloat[\vjepaft]{{\includegraphics[width=0.32\linewidth]{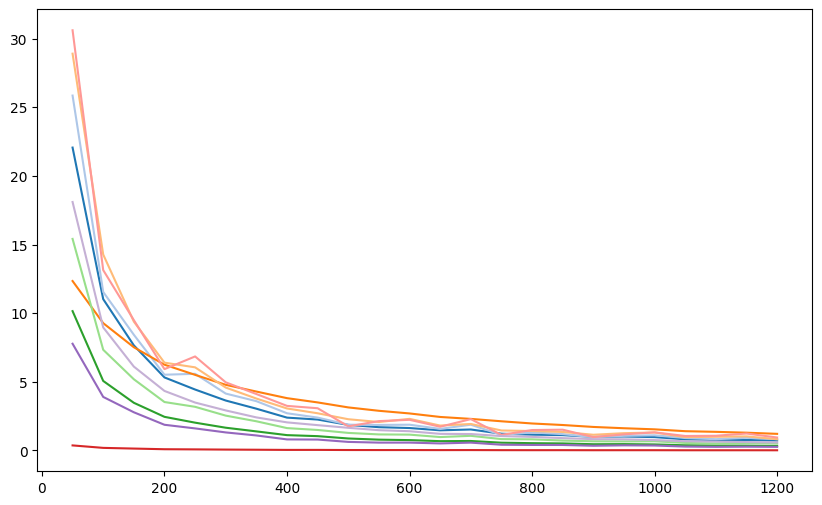} }}%
    \subfloat{\includegraphics[width=0.1\linewidth]{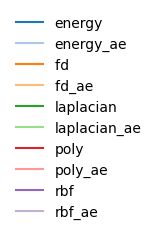}}%

    \subfloat[\videomaeft]{{\includegraphics[width=0.32\linewidth]{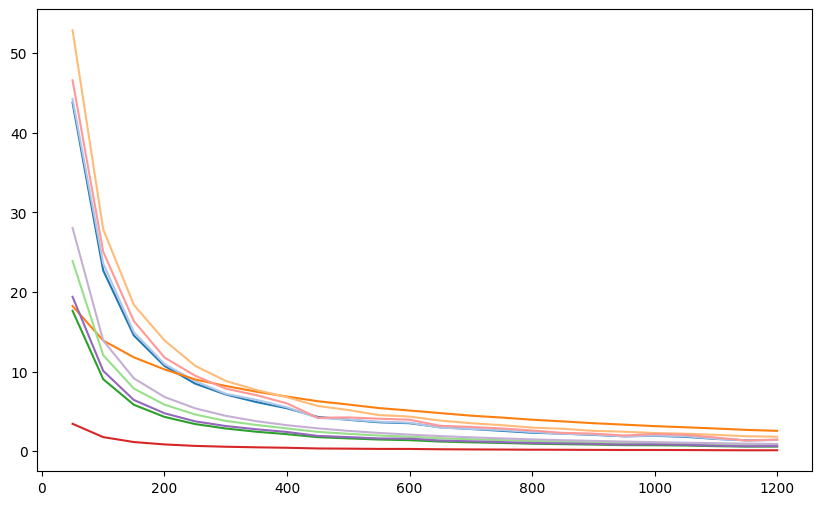} }}%
    \subfloat[\videomaept]{{\includegraphics[width=0.32\linewidth]{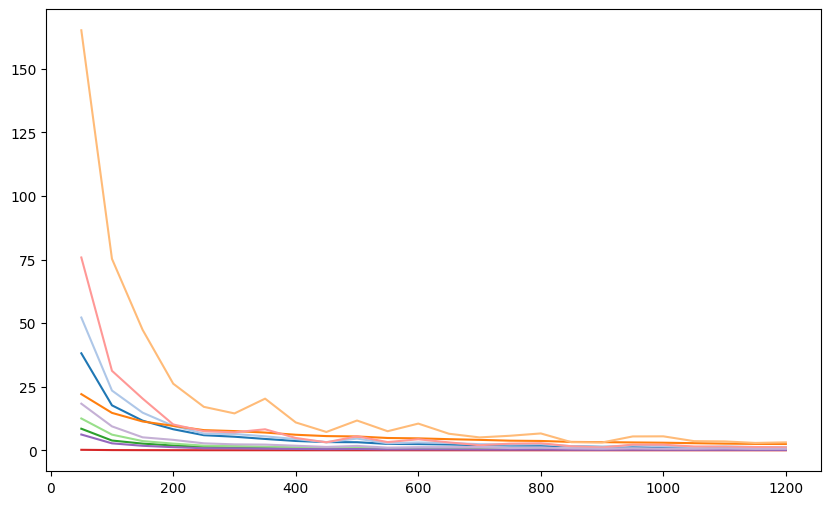} }}%
    \subfloat[\vjepapt]{{\includegraphics[width=0.32\linewidth]{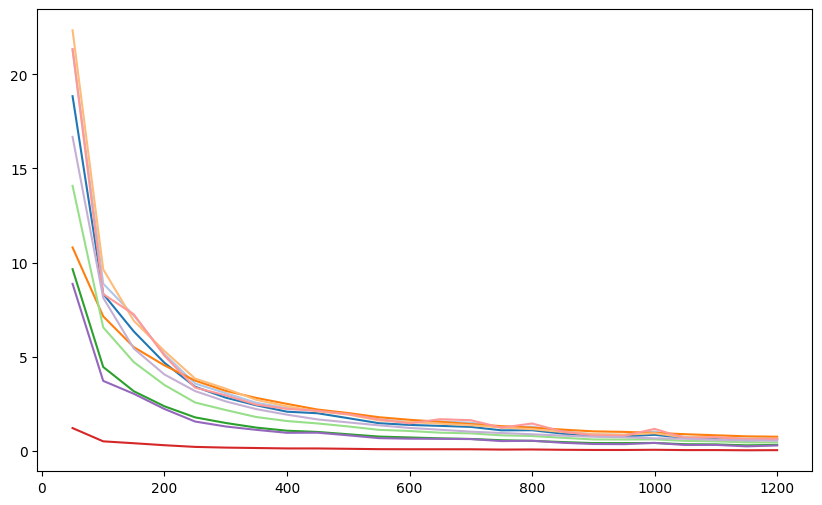} }}%
    
    \caption{Convergence rates of distributional metrics on UCF-101, comparing training and testing sets in various feature spaces. The convergence rate computation in these figures follows the same configuration as Figure~\ref{fig:train_test_convergence_rate}.}%
\label{fig:train_test_convergence_rate_other}
\end{figure}

\begin{figure}[h!]%
    \centering
    \setlength\tabcolsep{3pt} % default: 6pt
    \centering
    \begin{tabular}{c M{0.4\linewidth} M{0.4\linewidth}}
     & \textbf{\videomaept}  & \textbf{\vjepapt} \\
    UCF-101 & \includegraphics[width=\hsize]{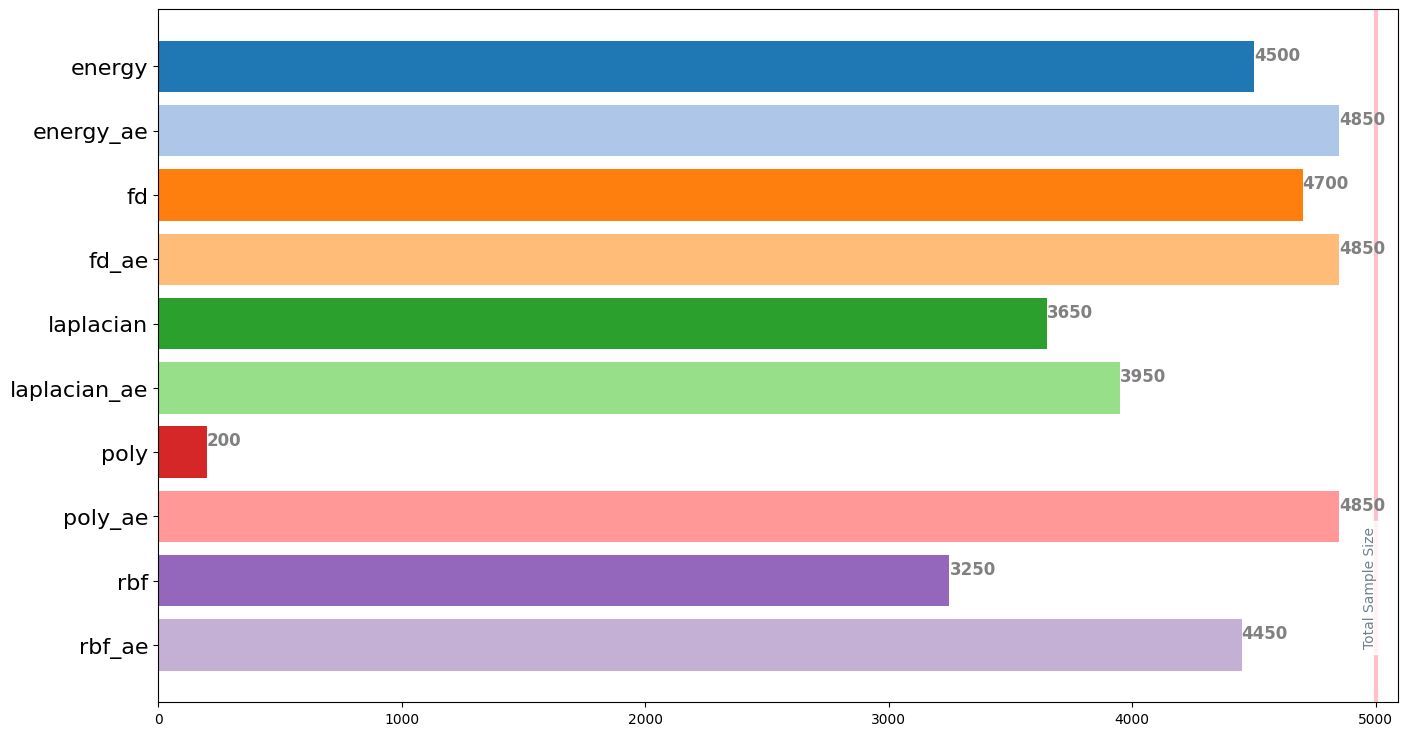}  
      & \includegraphics[width=\hsize]{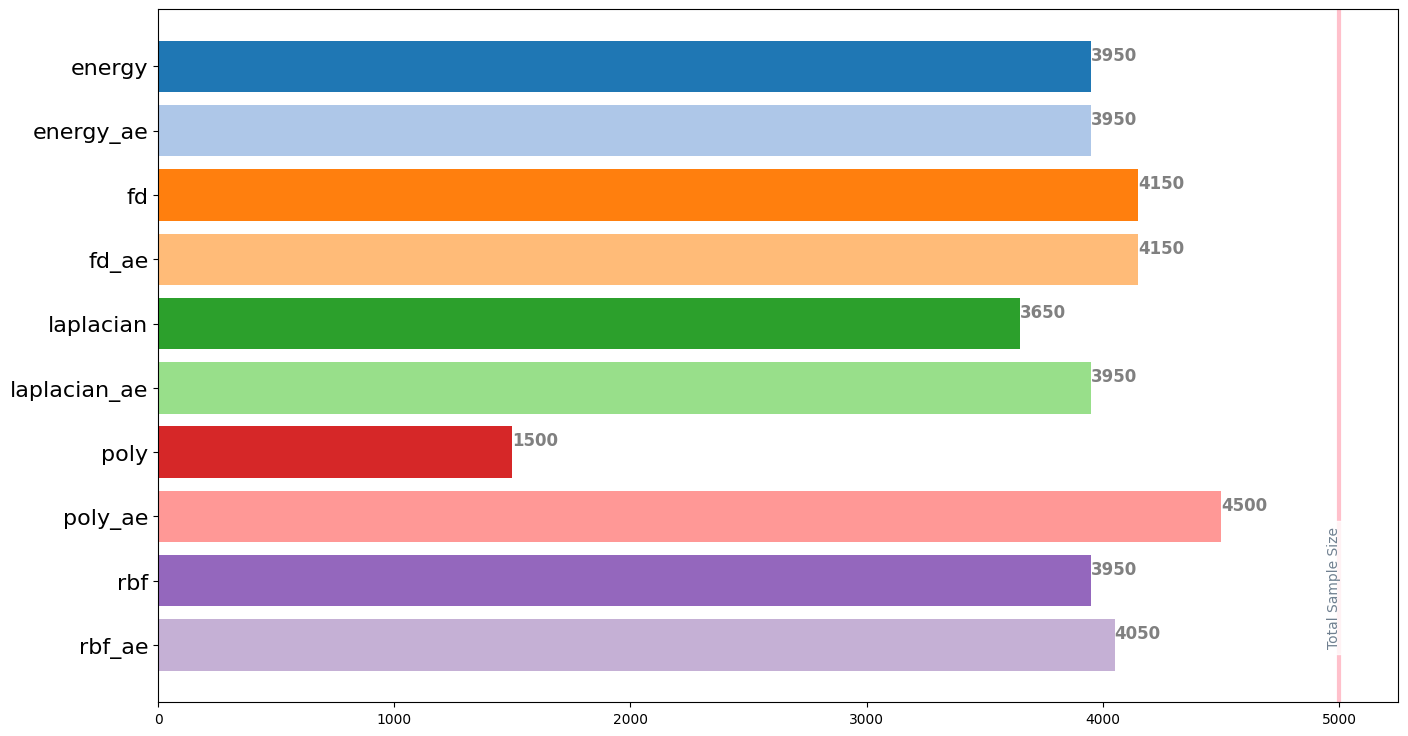}
      \\
      SSv2 & \includegraphics[width=\hsize]{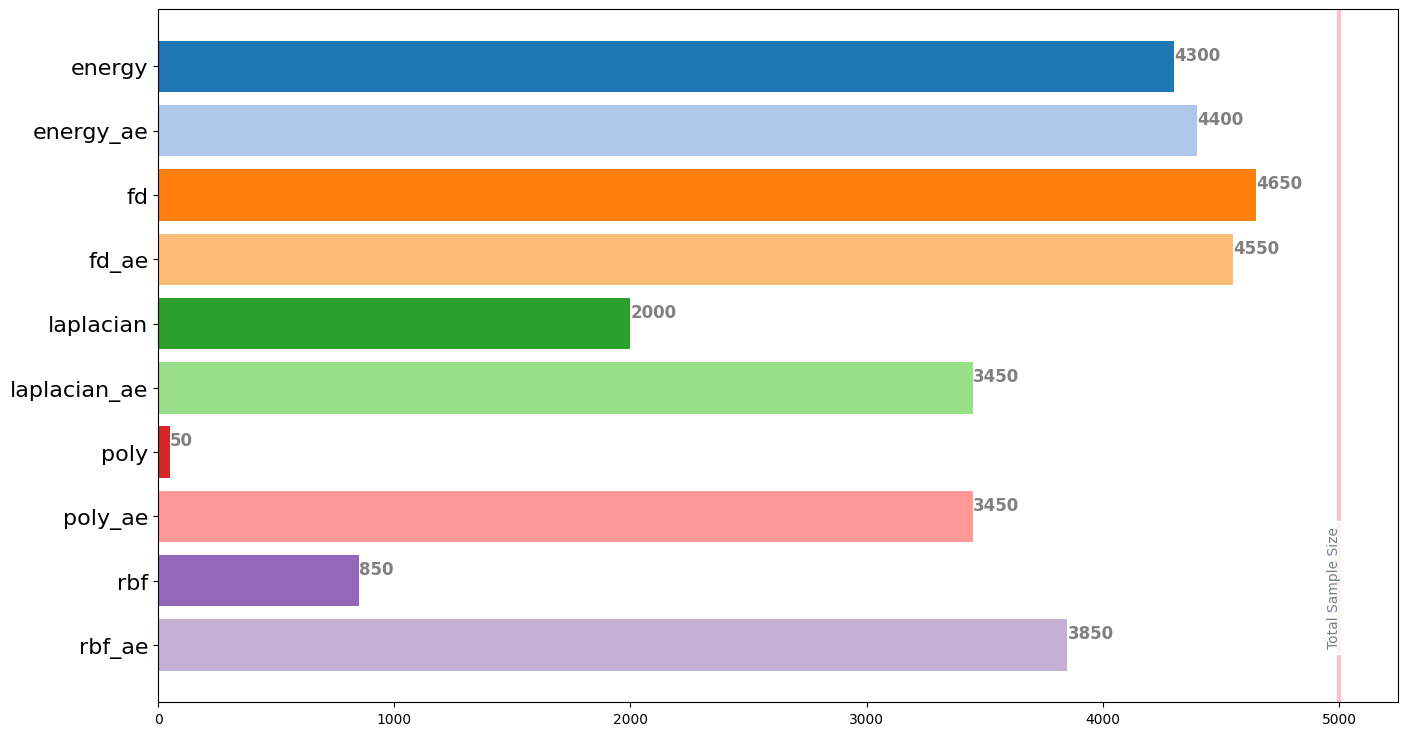}   
      & \includegraphics[width=\hsize]{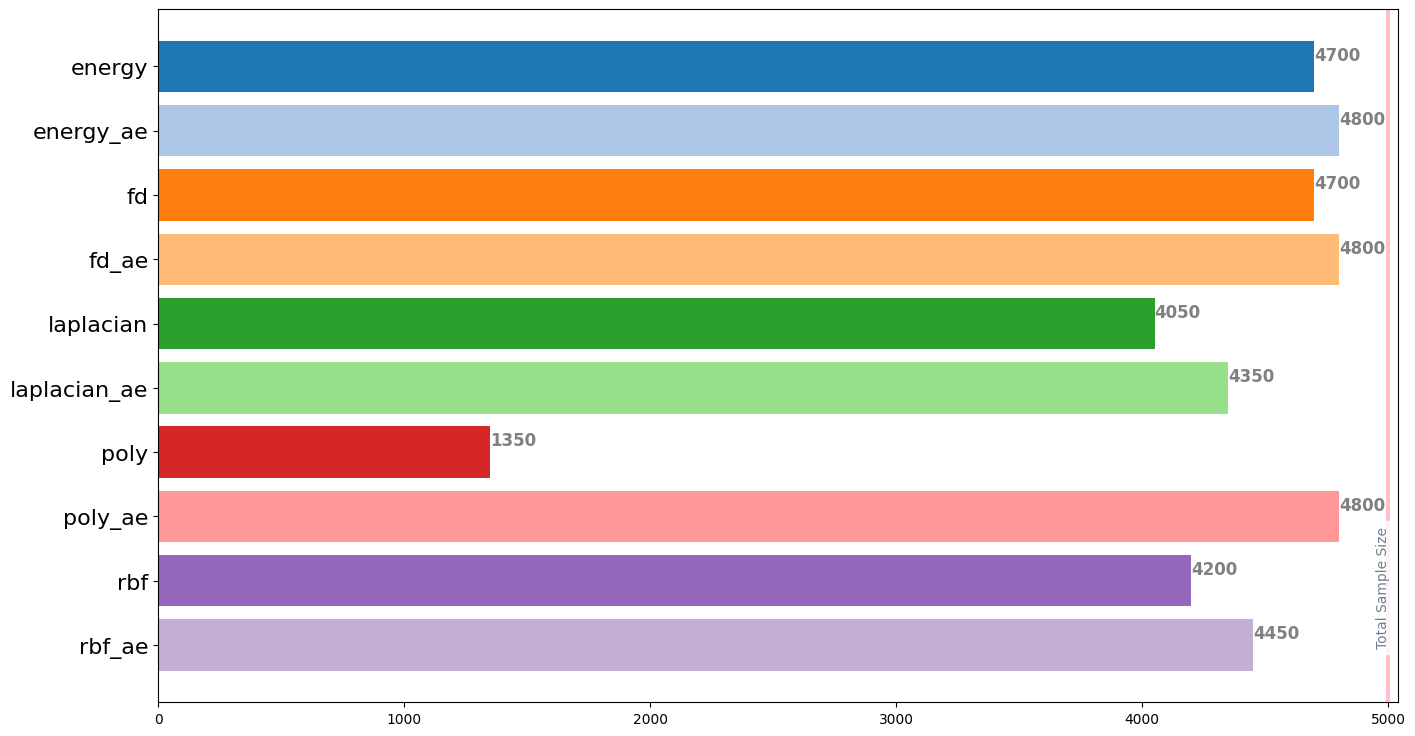}\\
      \end{tabular}
    \caption{This figure shows the number of samples needed for \videomaept~and \vjepapt~to achieve a 5\% error margin of the distance measured from 5,000 samples using the training and testing sets. The convergence requirement is stated in Figure~\ref{fig:number_sample_convergence}.}%
    \label{fig:number_sample_convergence_others}
    \raggedright \hyperlink{convergence-samples}{\house} Back to paper
\end{figure}

\begin{figure}[h]%
    \centering
    \subfloat[I3D]{{\includegraphics[width=0.18\linewidth]{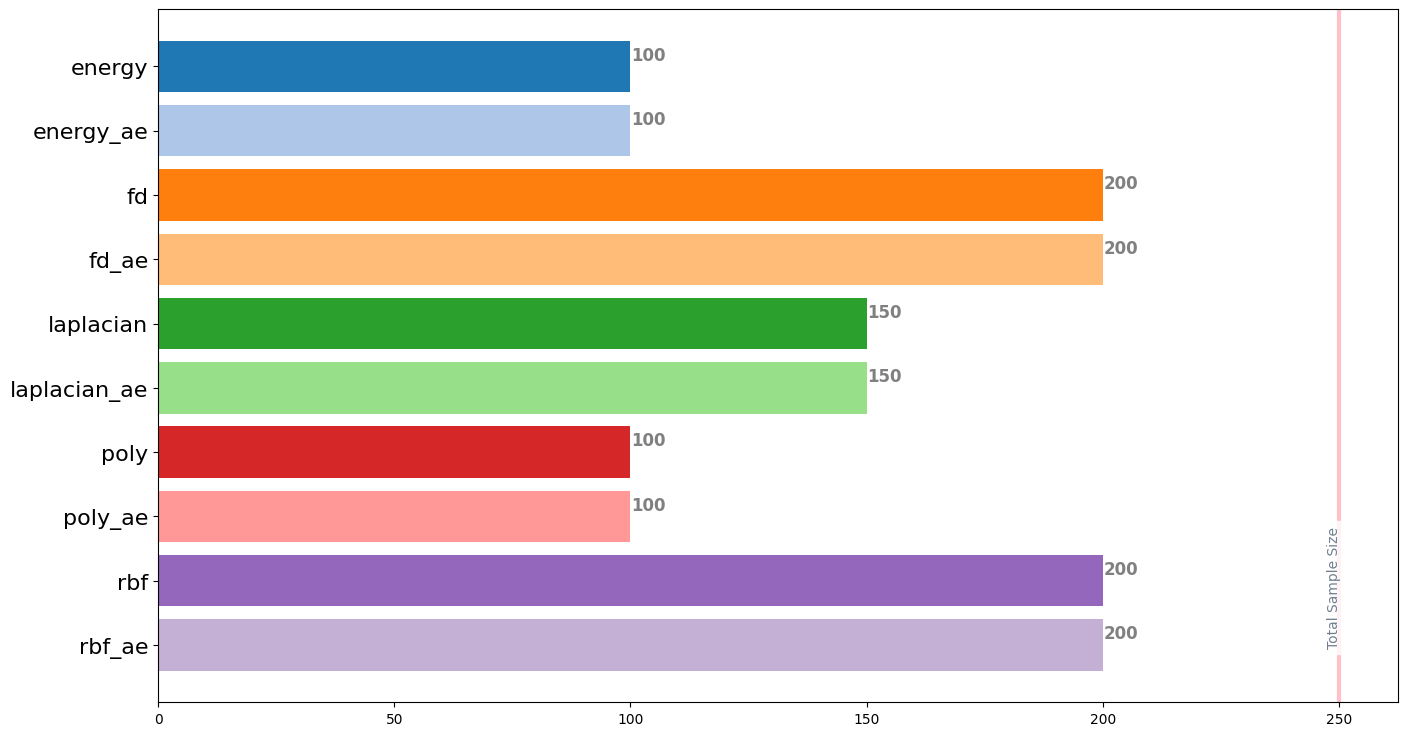} }}%
    \subfloat[\videomaept]{{\includegraphics[width=0.18\linewidth]{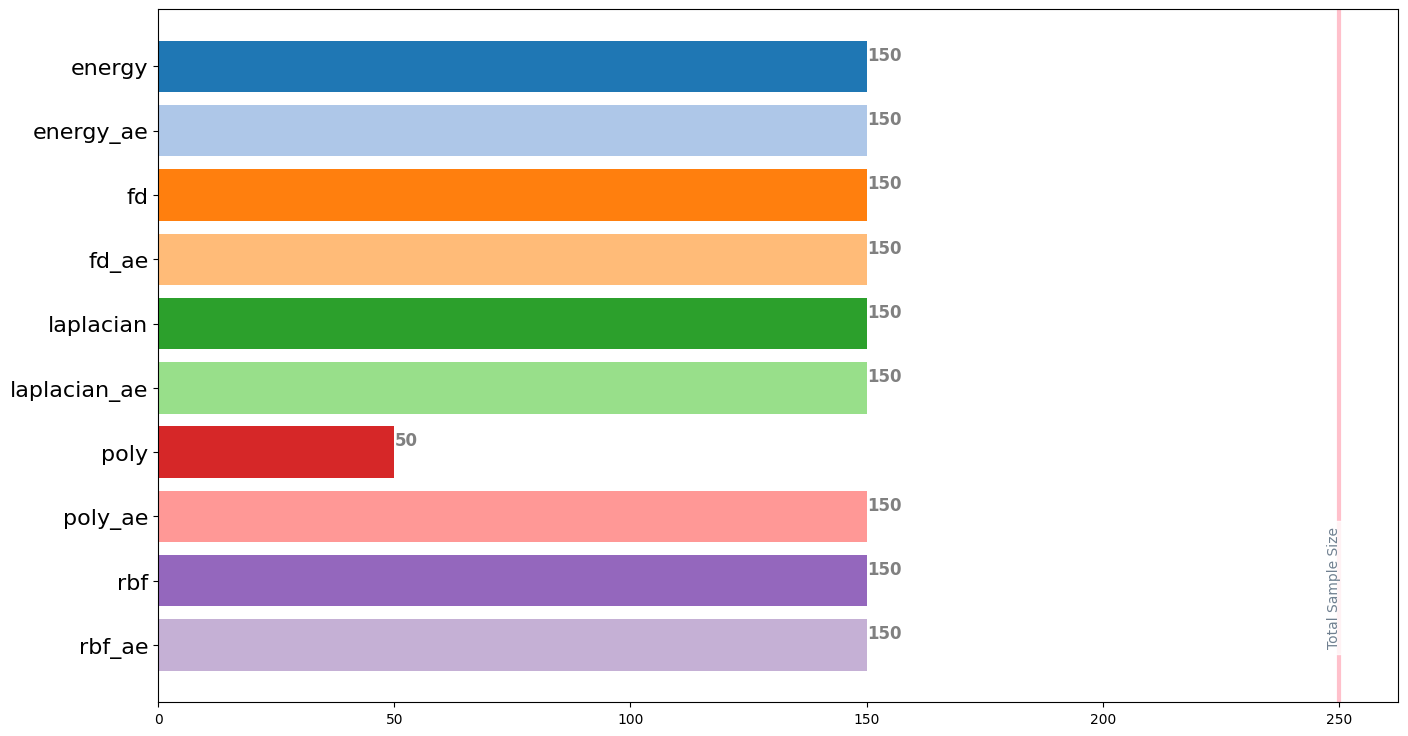} }}%
    \subfloat[\videomaeft]{{\includegraphics[width=0.18\linewidth]{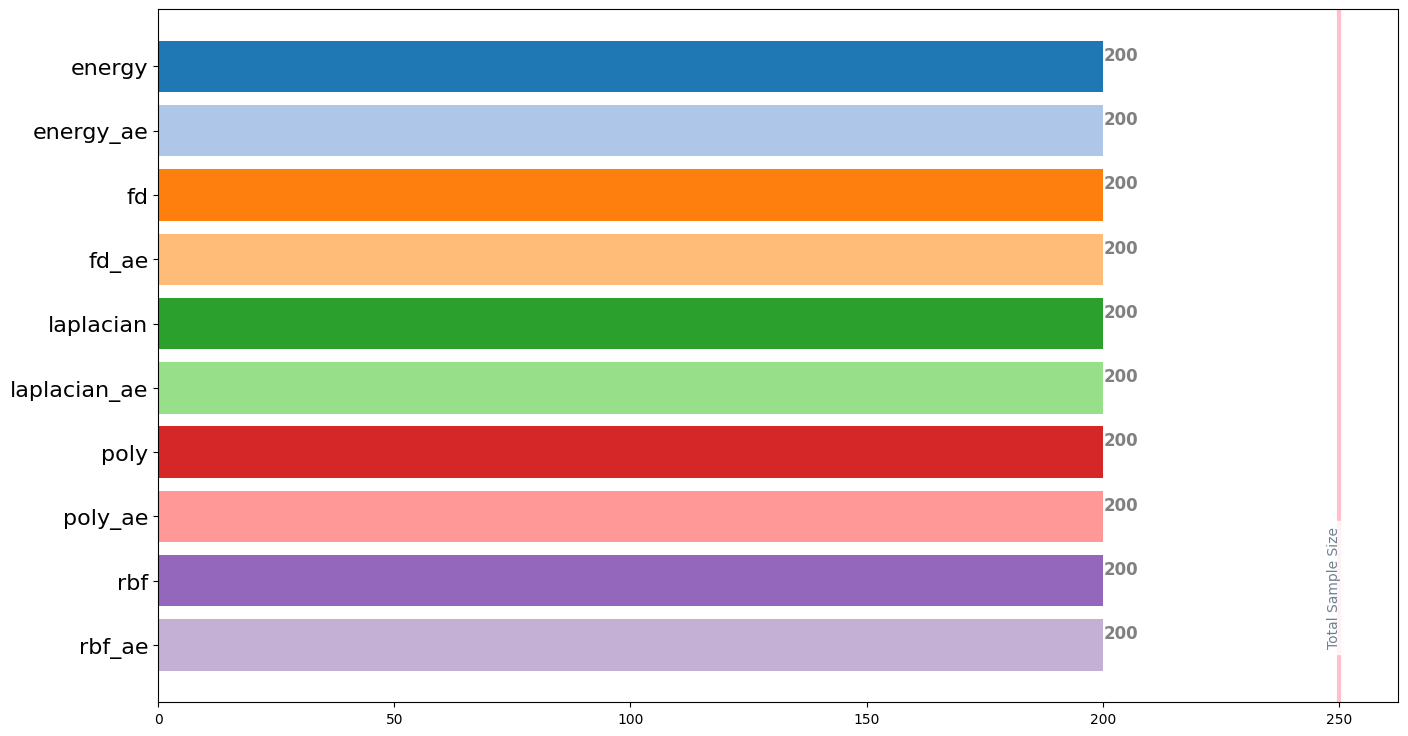} }}%
    \subfloat[\vjepapt]{{\includegraphics[width=0.18\linewidth]{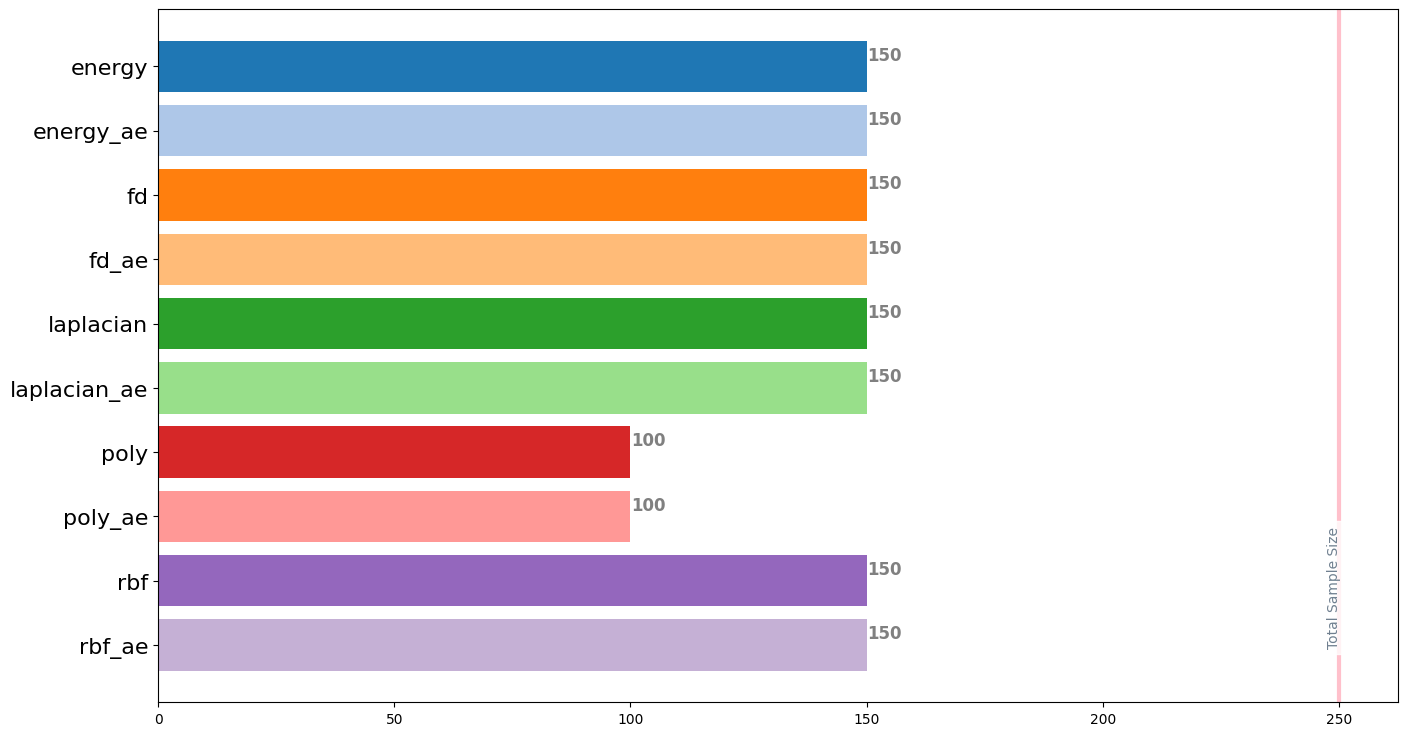} }}%
    \subfloat[\vjepaft]{{\includegraphics[width=0.18\linewidth]{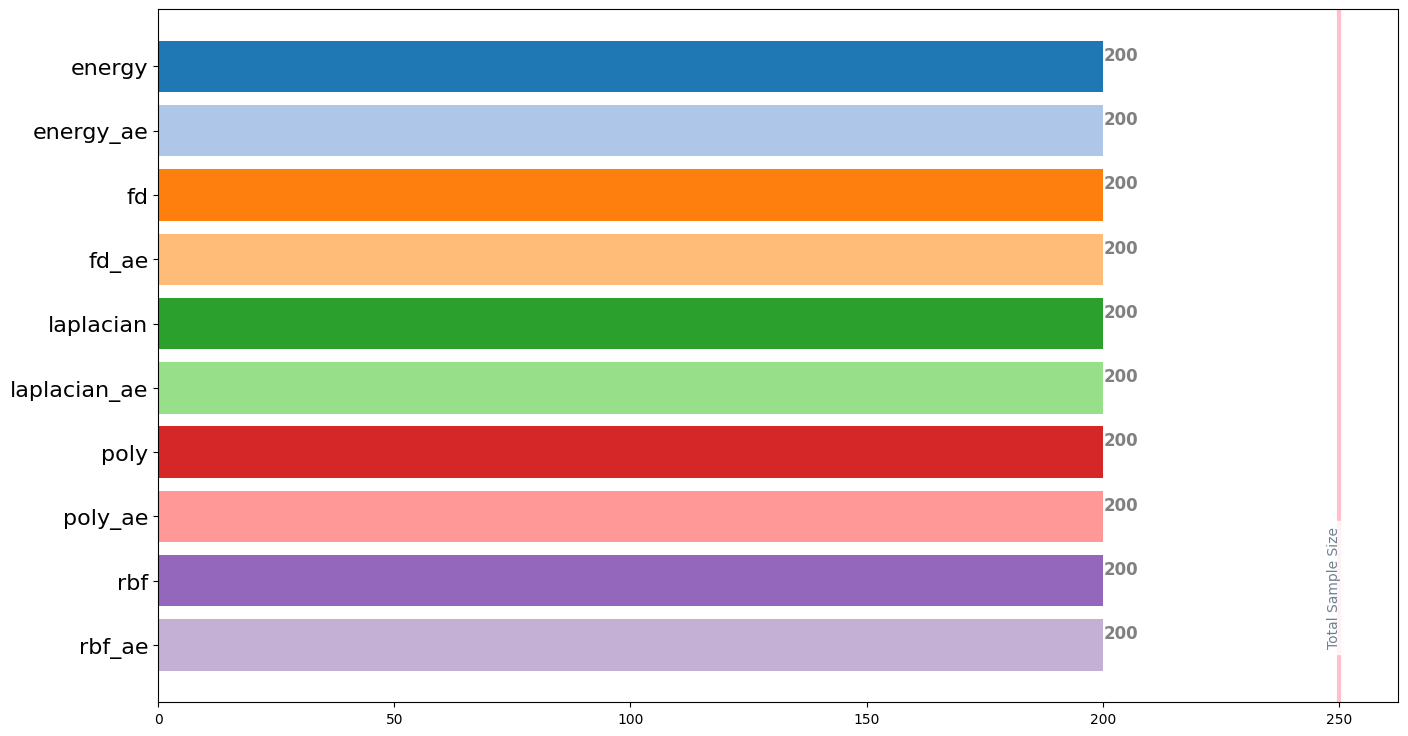} }}%
    \caption{The figures display convergence rates of distributional metrics on BAIR, comparing training and testing sets in \videomaept~and \vjepaft~feature spaces.}%
\label{fig:bair_issue}
\end{figure}

\begin{figure}[h!]%
    \centering
    \setlength\tabcolsep{3pt} % default: 6pt
    \resizebox{\textwidth}{!}{\begin{tabular}{c M{0.3\linewidth} M{0.3\linewidth} M{0.3\linewidth}}
     & \textbf{I3D}  & \textbf{\vjepapt} & \textbf{\vjepaft}\\
    Anime & \includegraphics[width=\hsize]{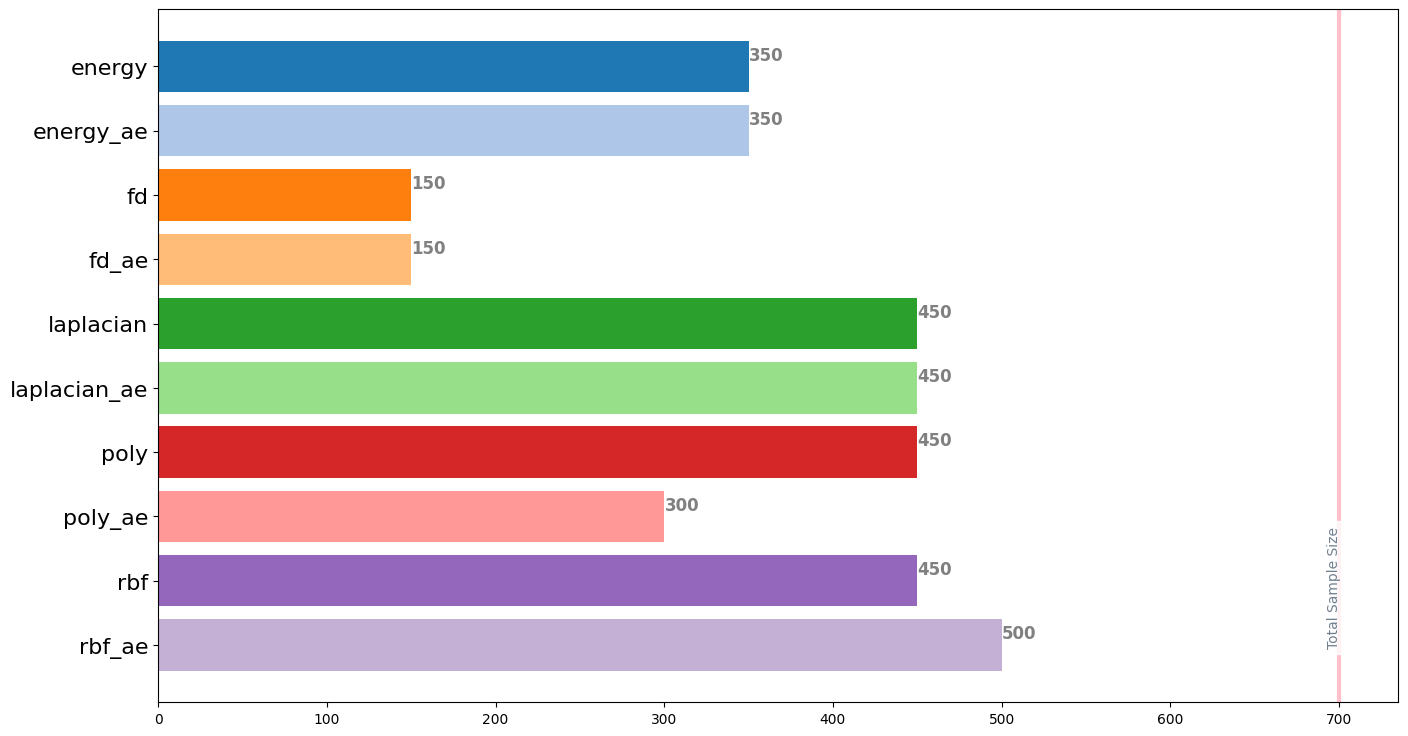}  
      & \includegraphics[width=\hsize]{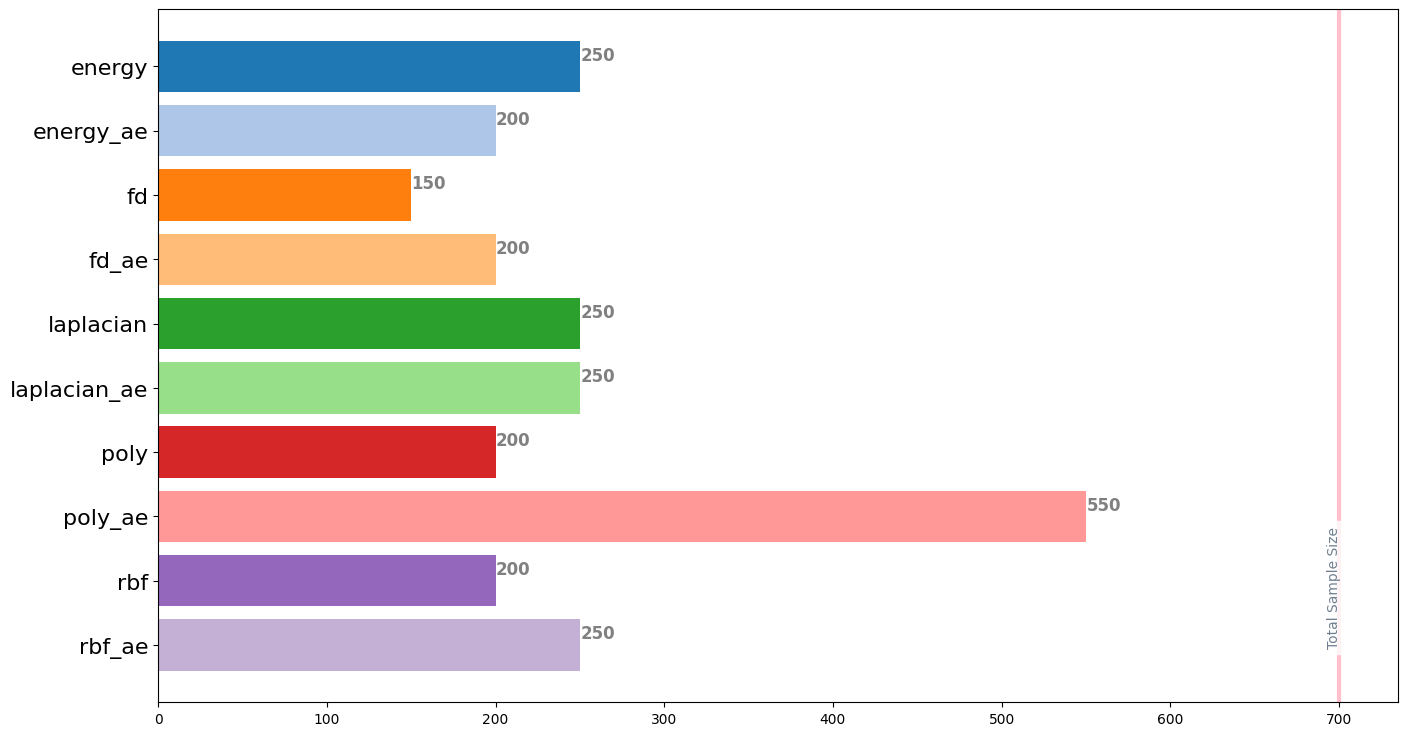}
      & \includegraphics[width=\hsize]{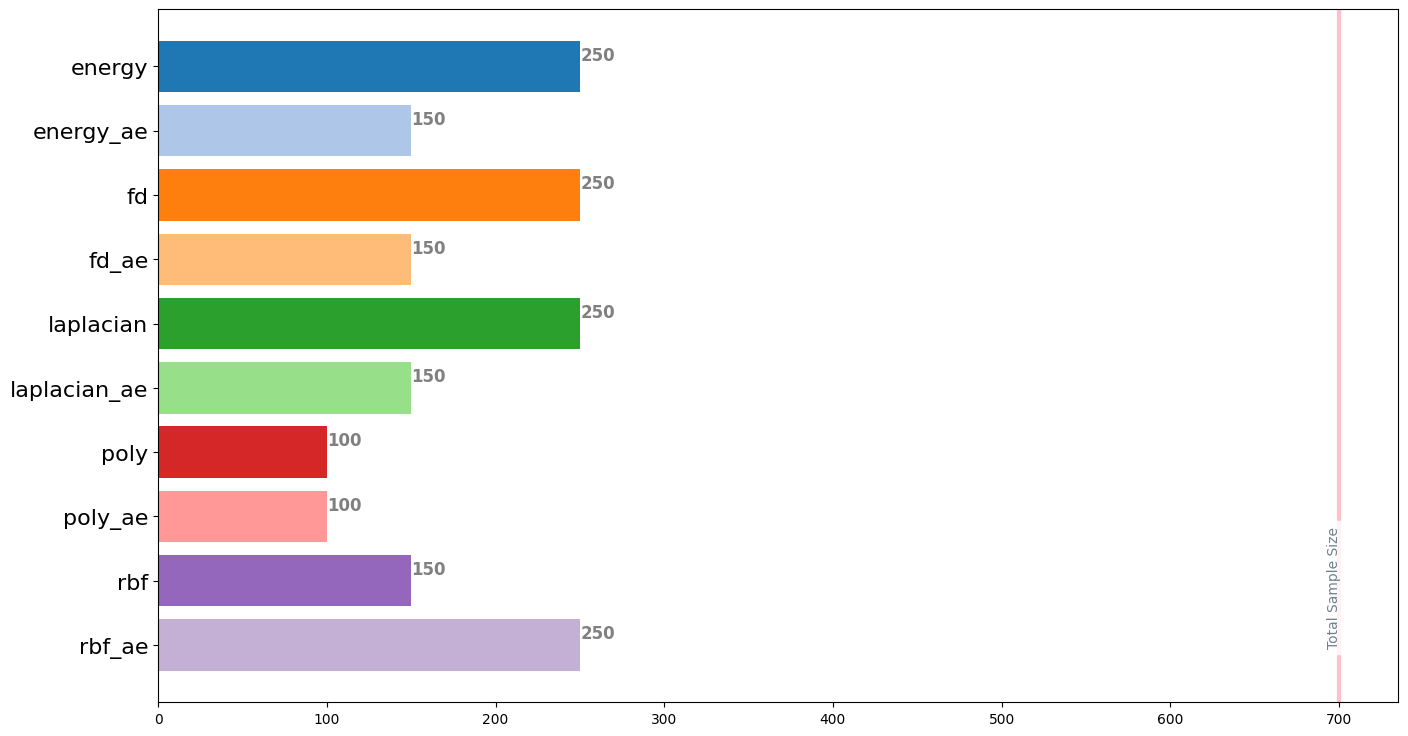}
      \\
      BDD & \includegraphics[width=\hsize]{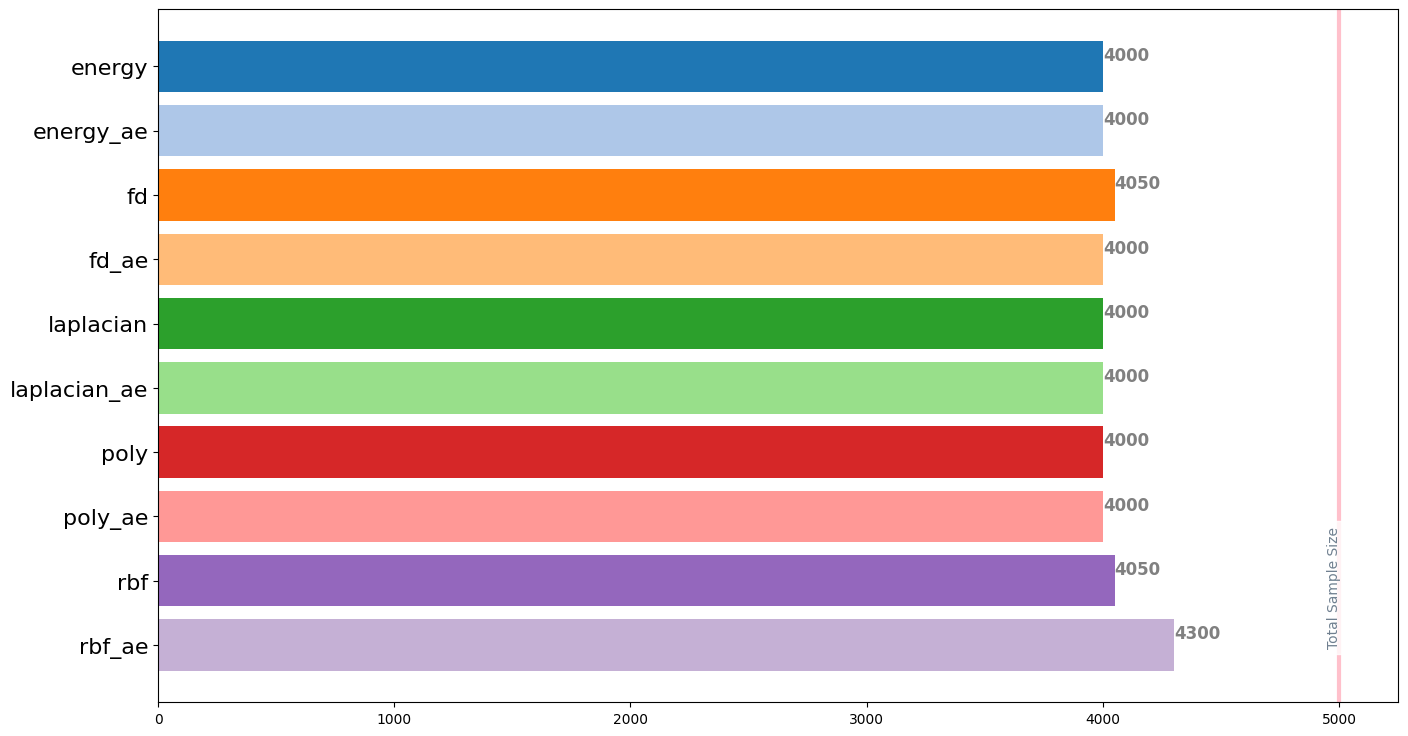}  
      & \includegraphics[width=\hsize]{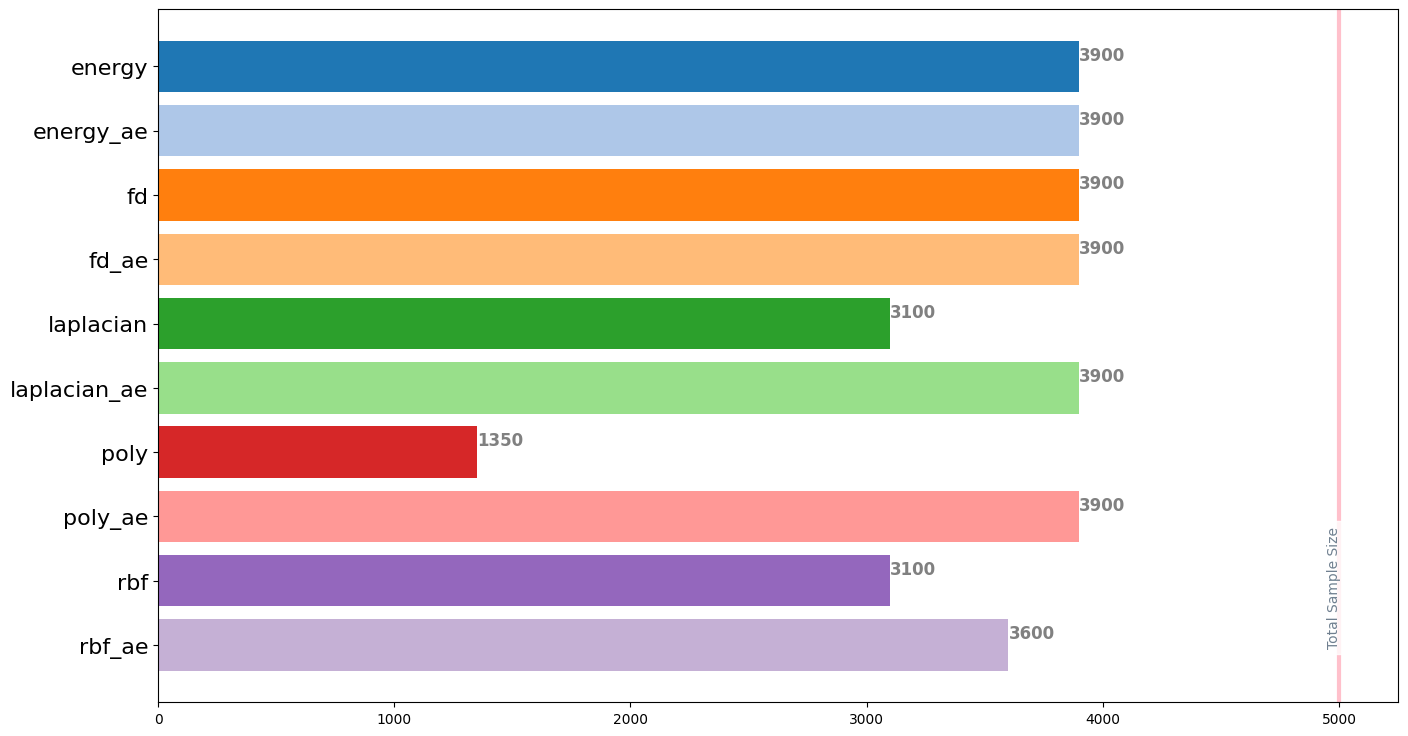}
      & \includegraphics[width=\hsize]{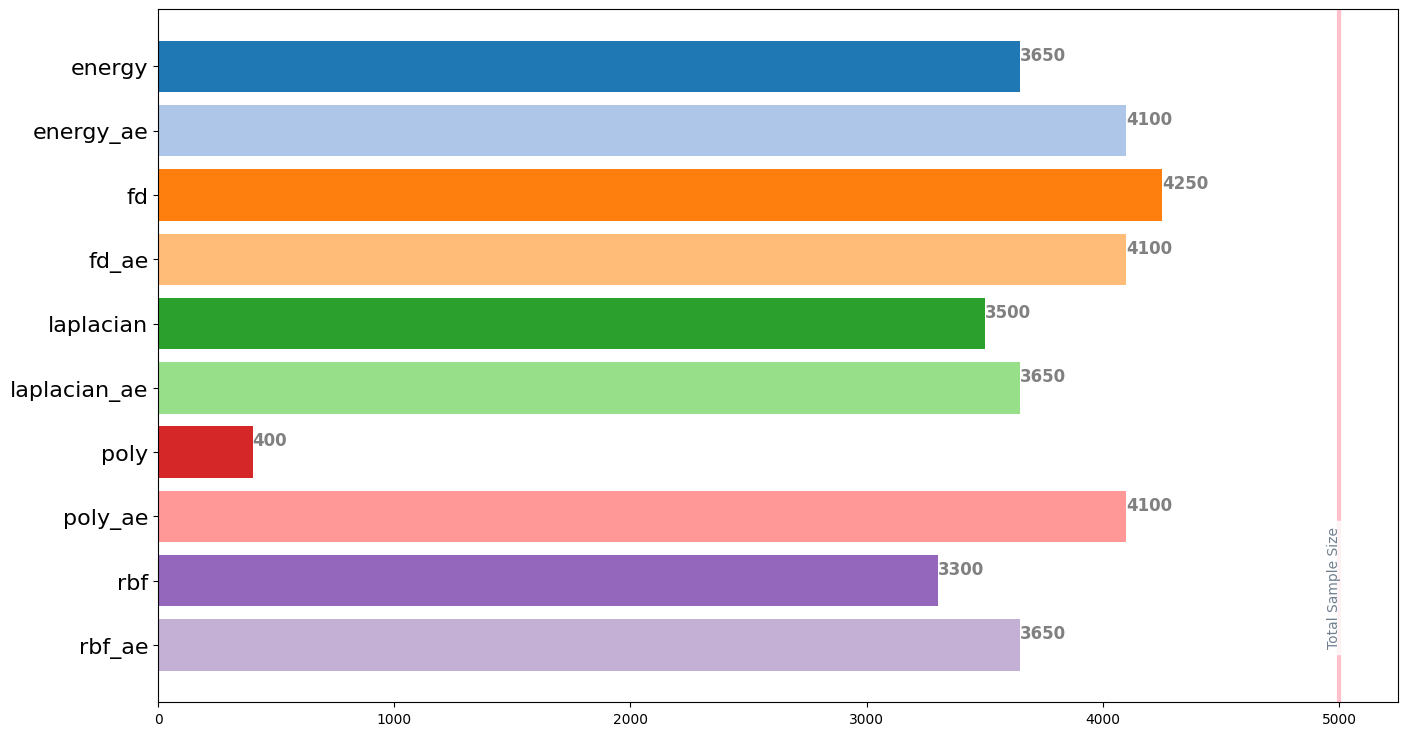}
      \\
      Davis & \includegraphics[width=\hsize]{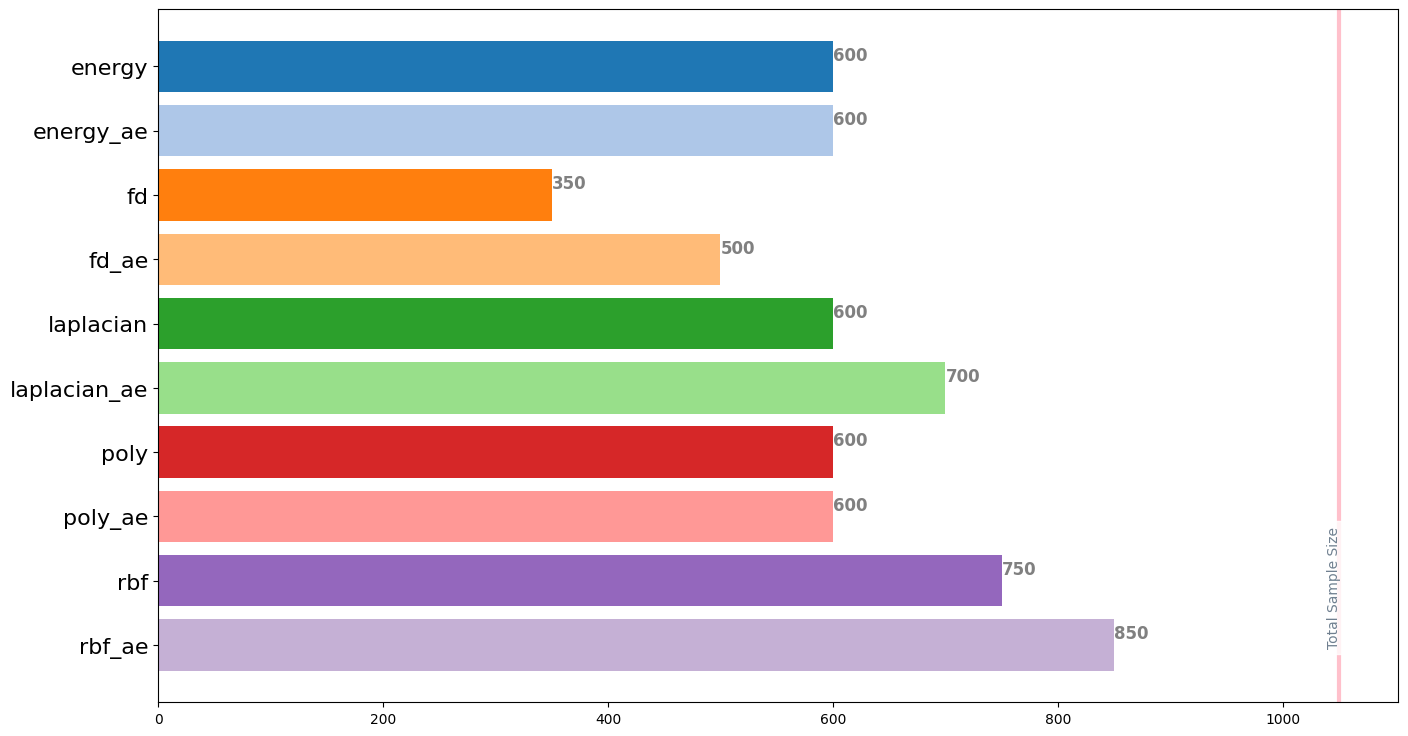}  
      & \includegraphics[width=\hsize]{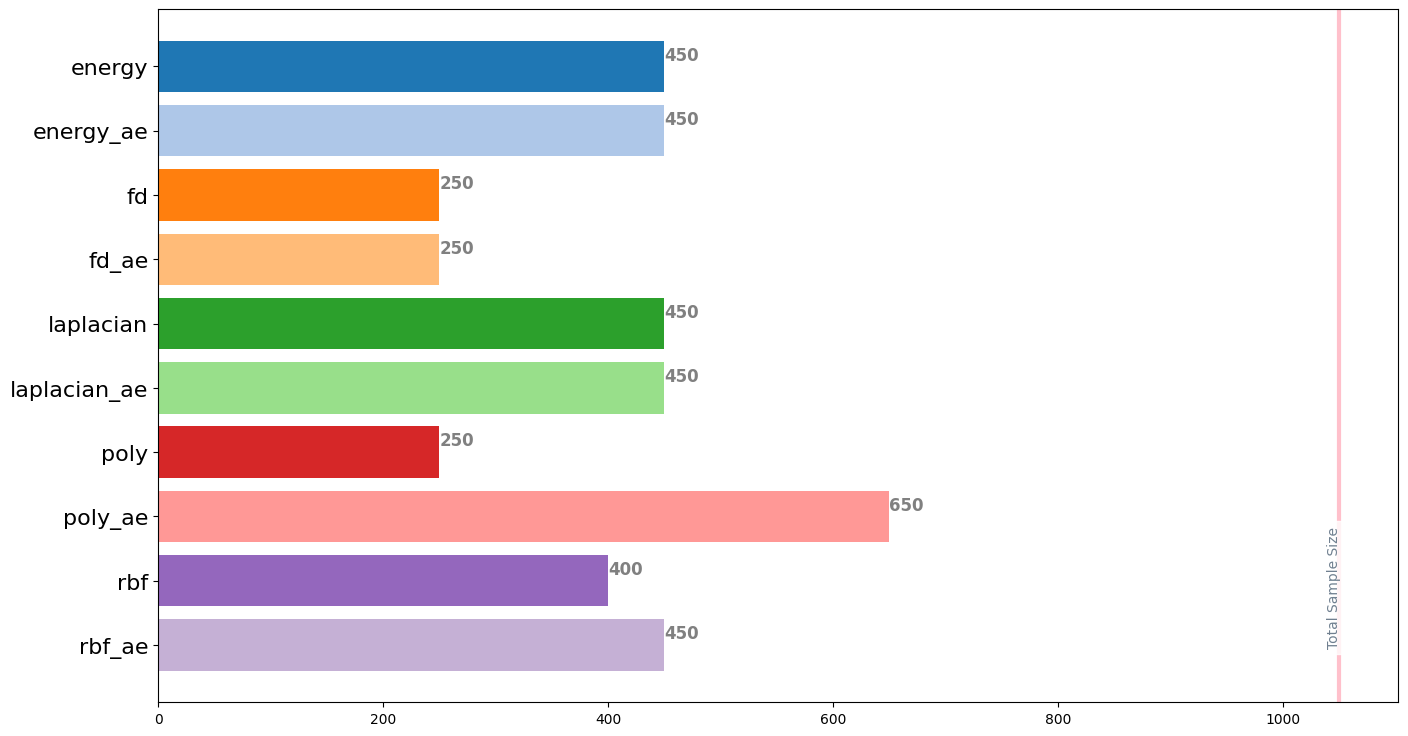}
      & \includegraphics[width=\hsize]{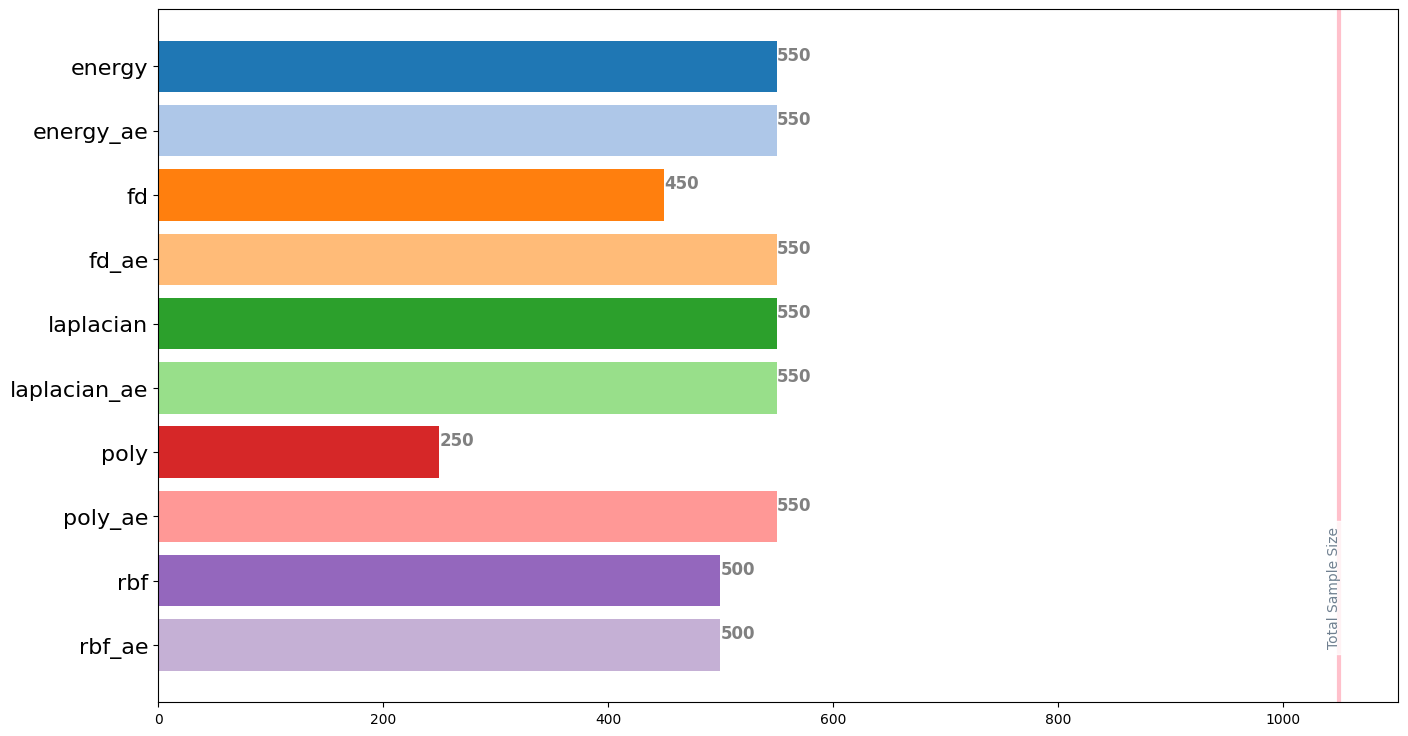}
      \\
      Fashion & \includegraphics[width=\hsize]{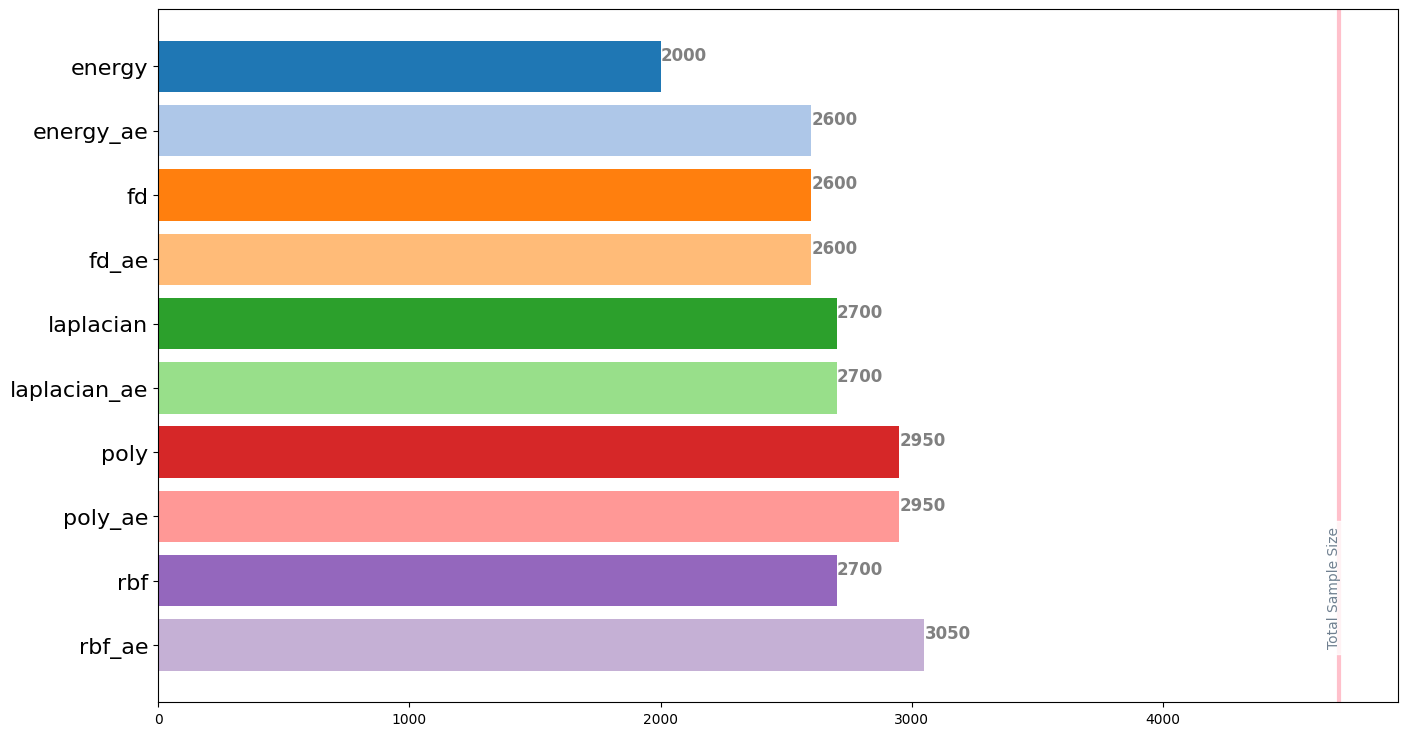}  
      & \includegraphics[width=\hsize]{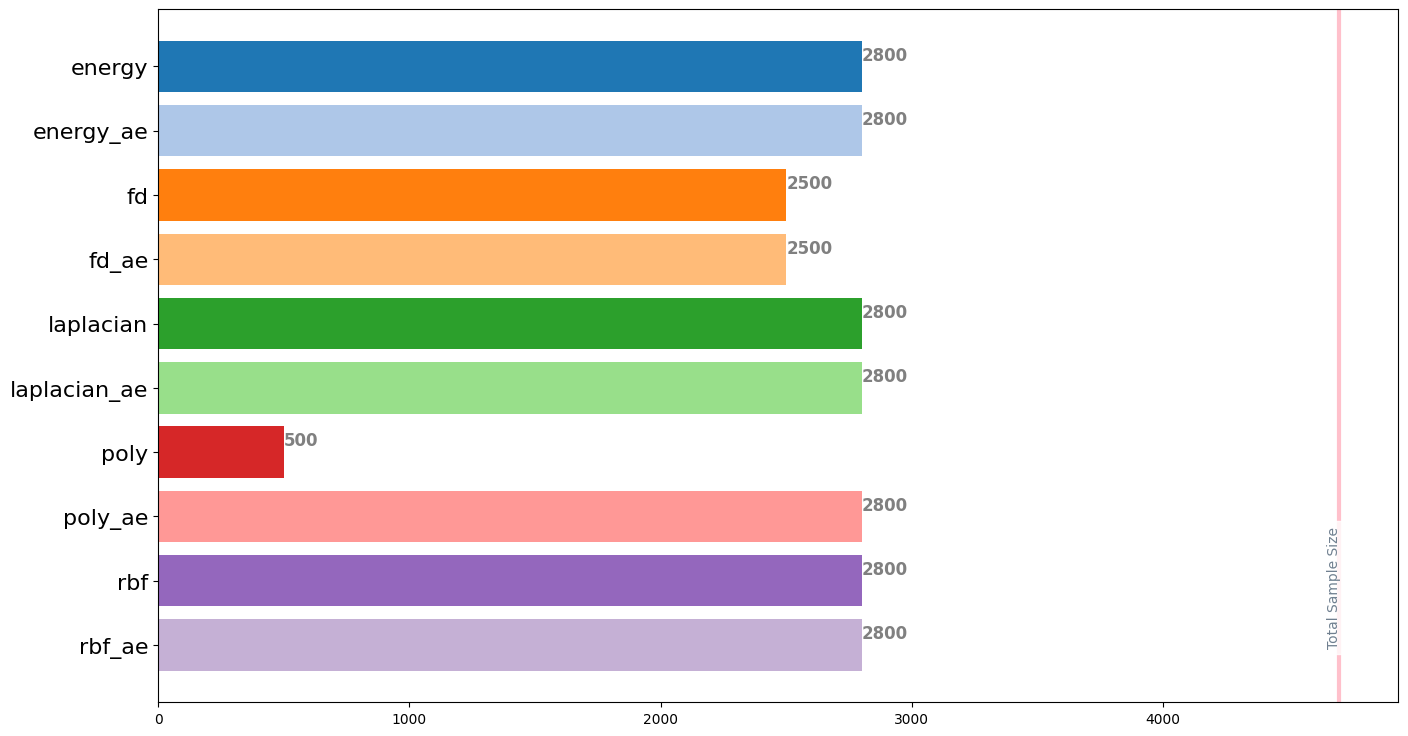}
      & 
      \includegraphics[width=\hsize]{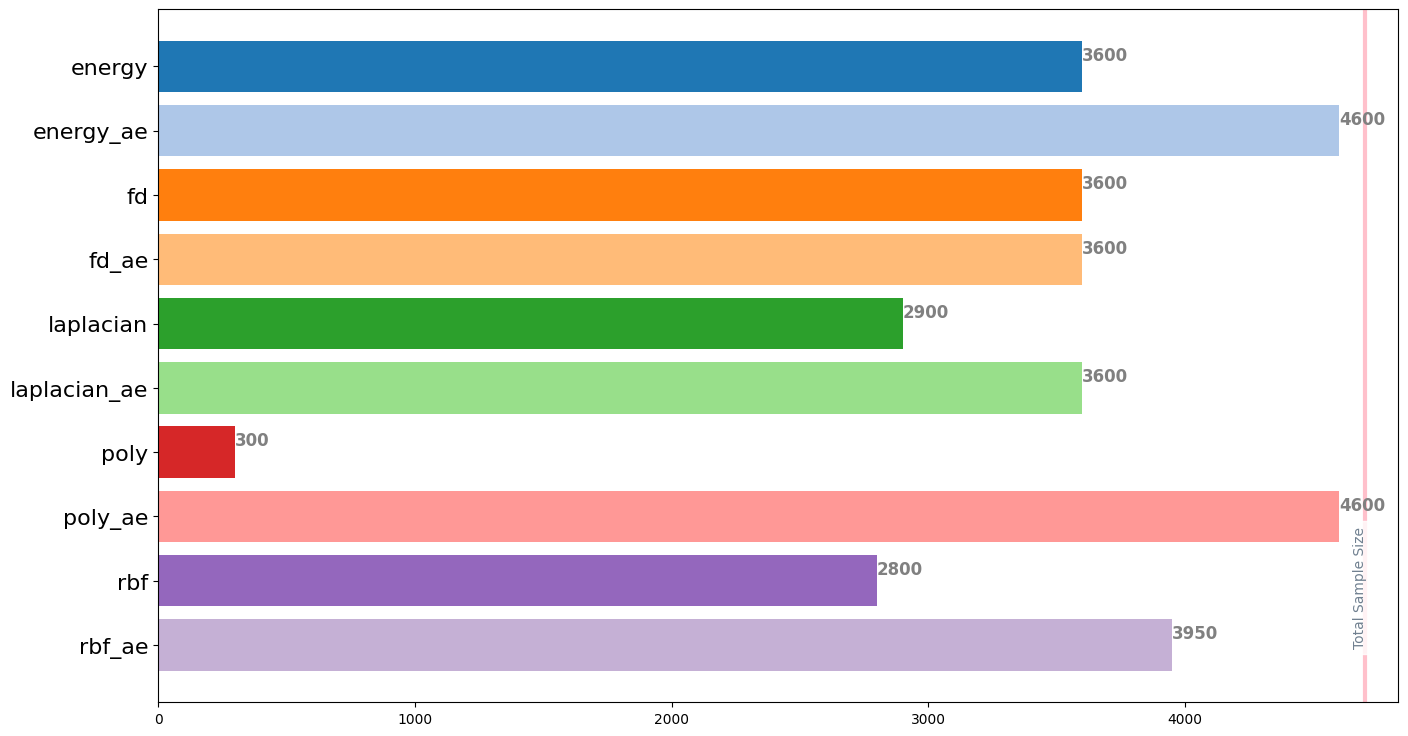}
      \\
      HMDB & \includegraphics[width=\hsize]{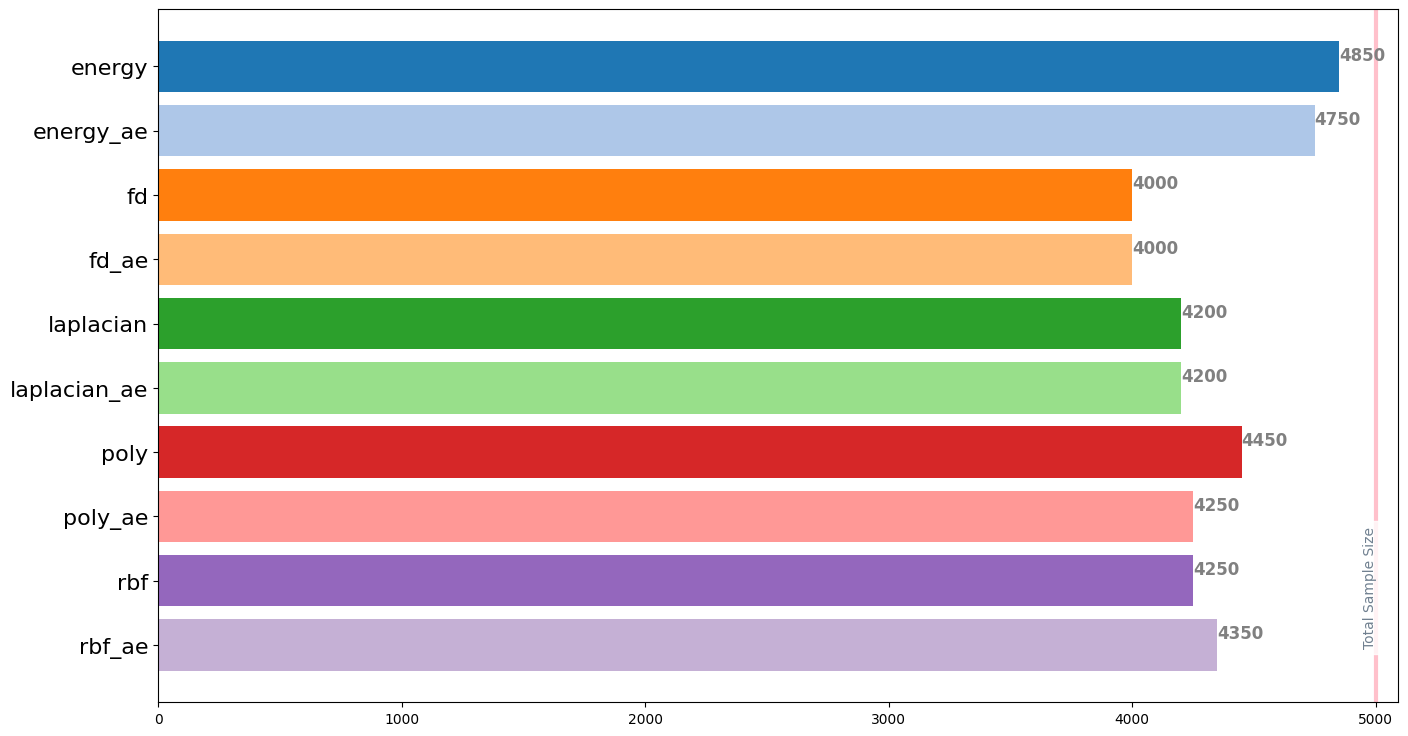}  
      & \includegraphics[width=\hsize]{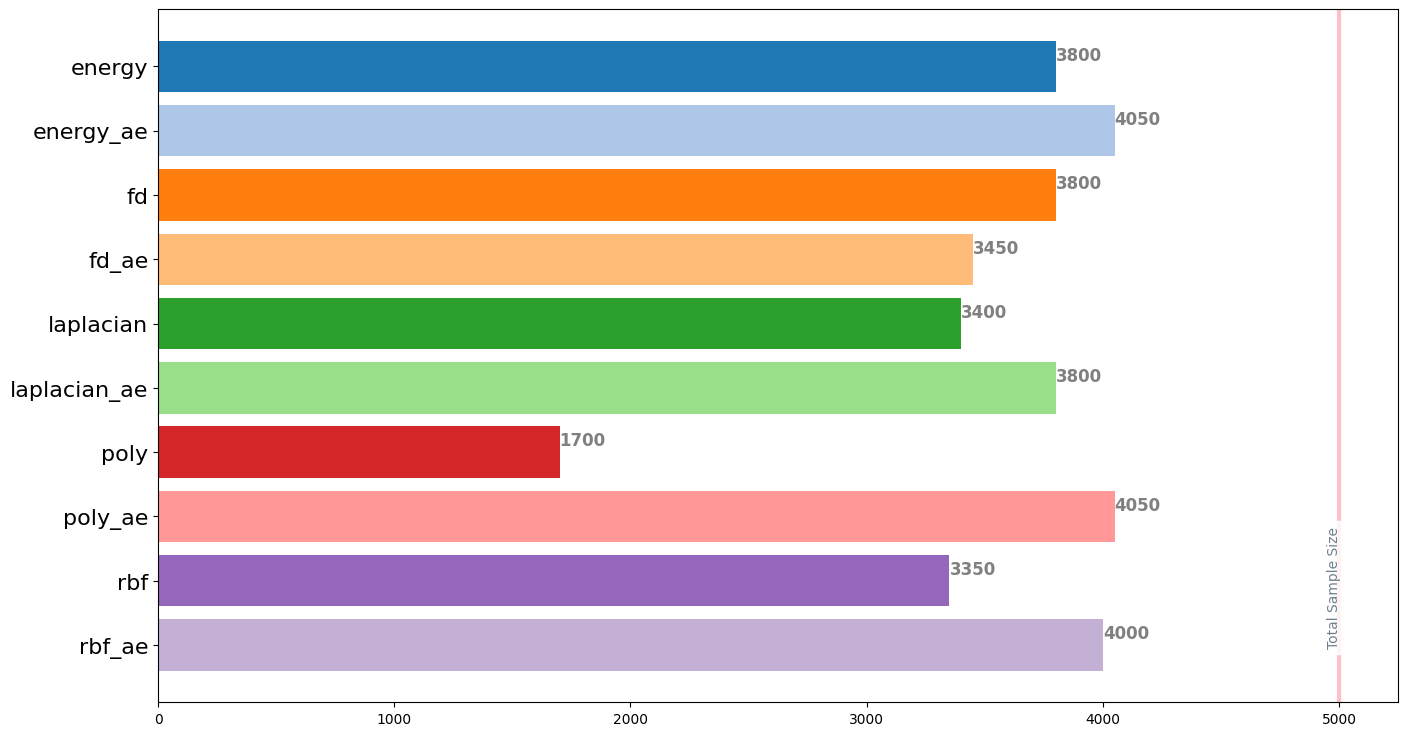}
      & 
      \includegraphics[width=\hsize]{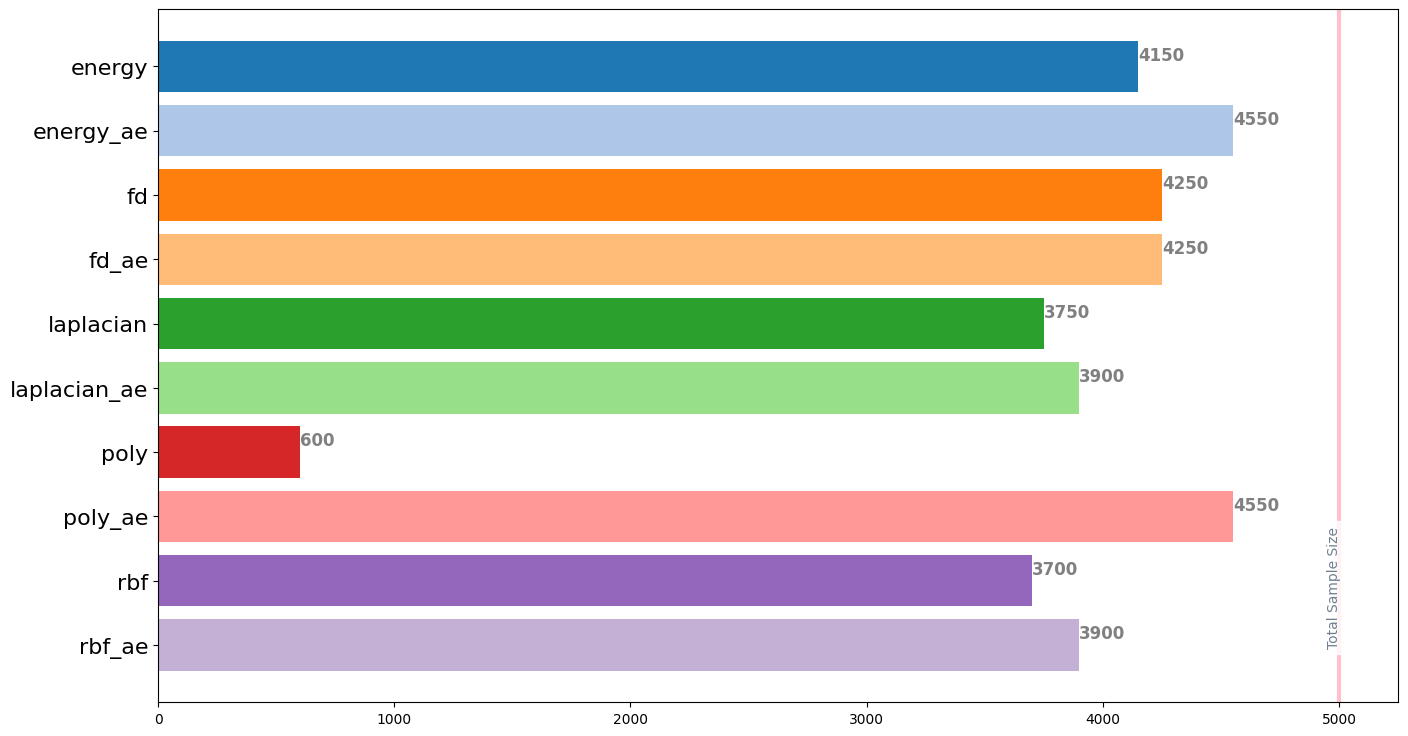}
      \\
      KITTI & \includegraphics[width=\hsize]{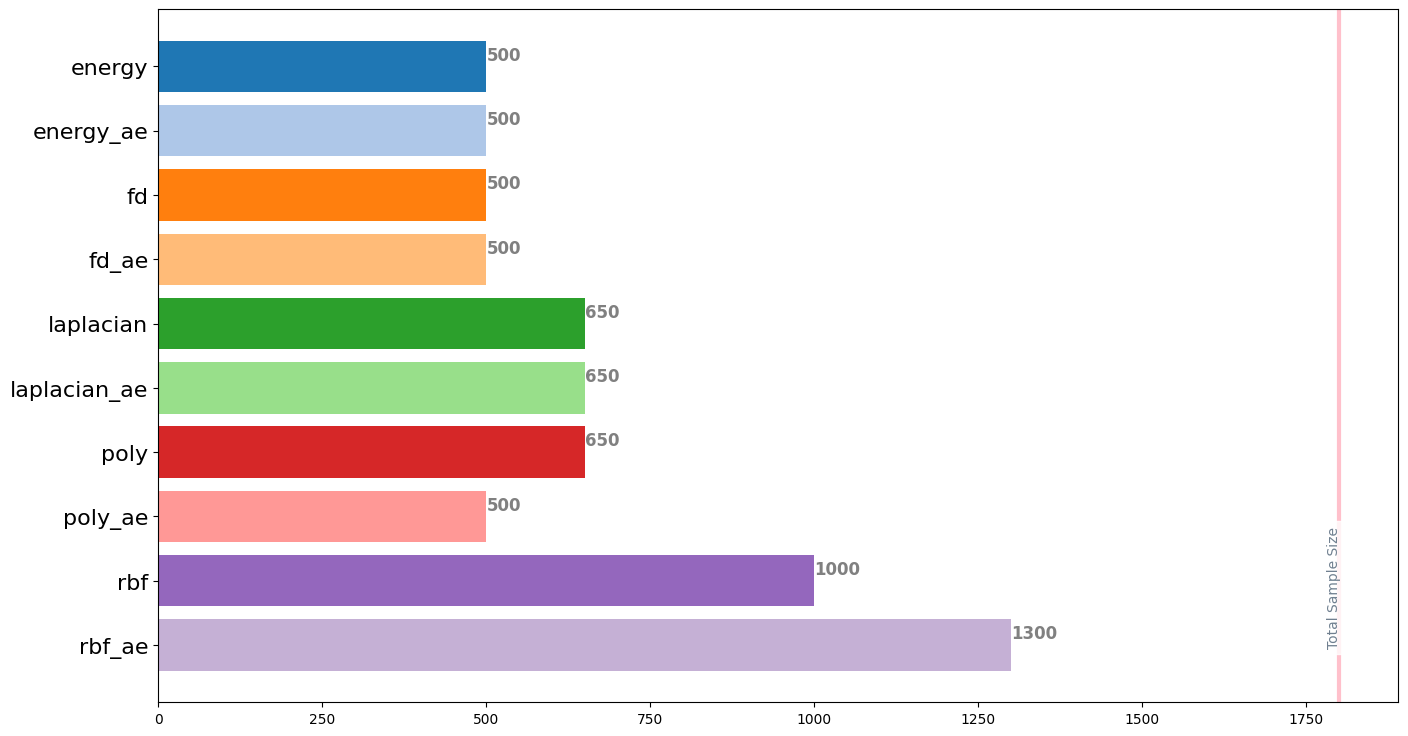}  
      & \includegraphics[width=\hsize]{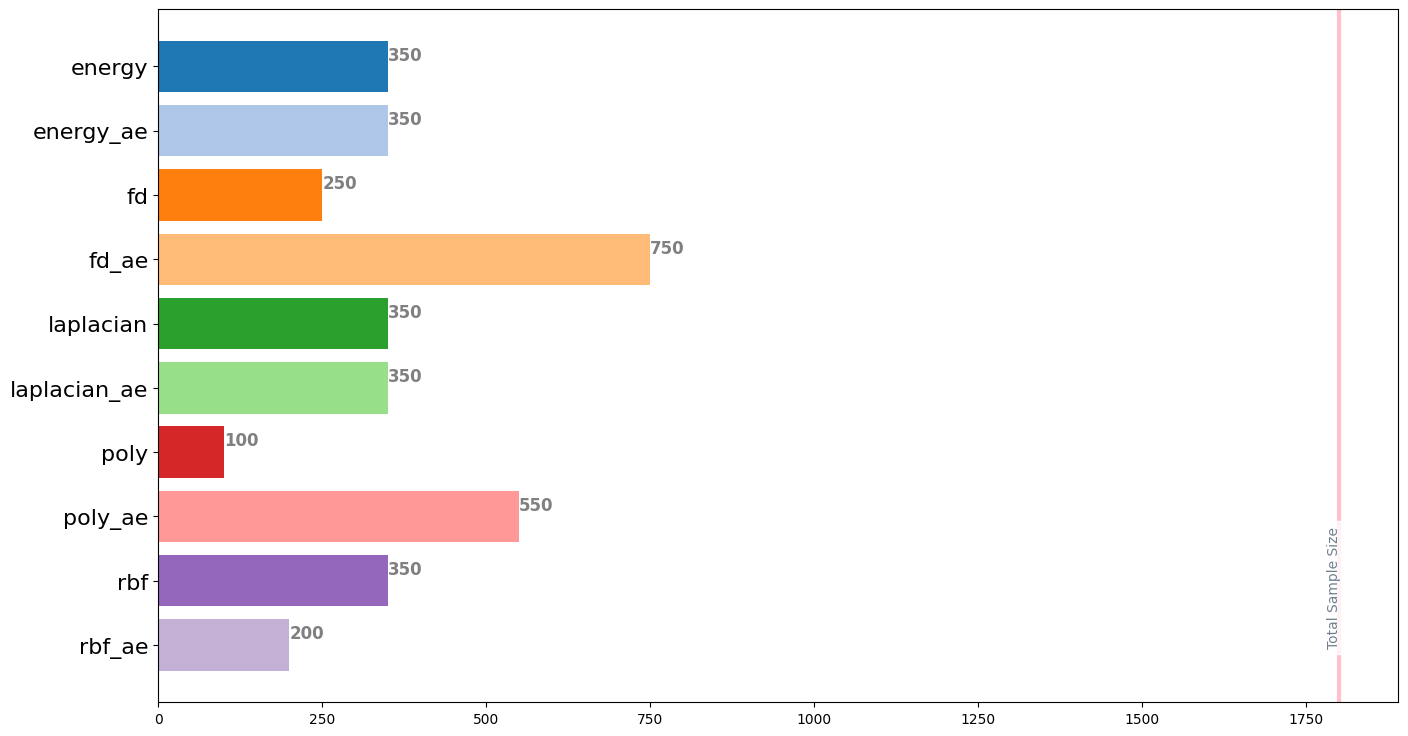}
      & \includegraphics[width=\hsize]{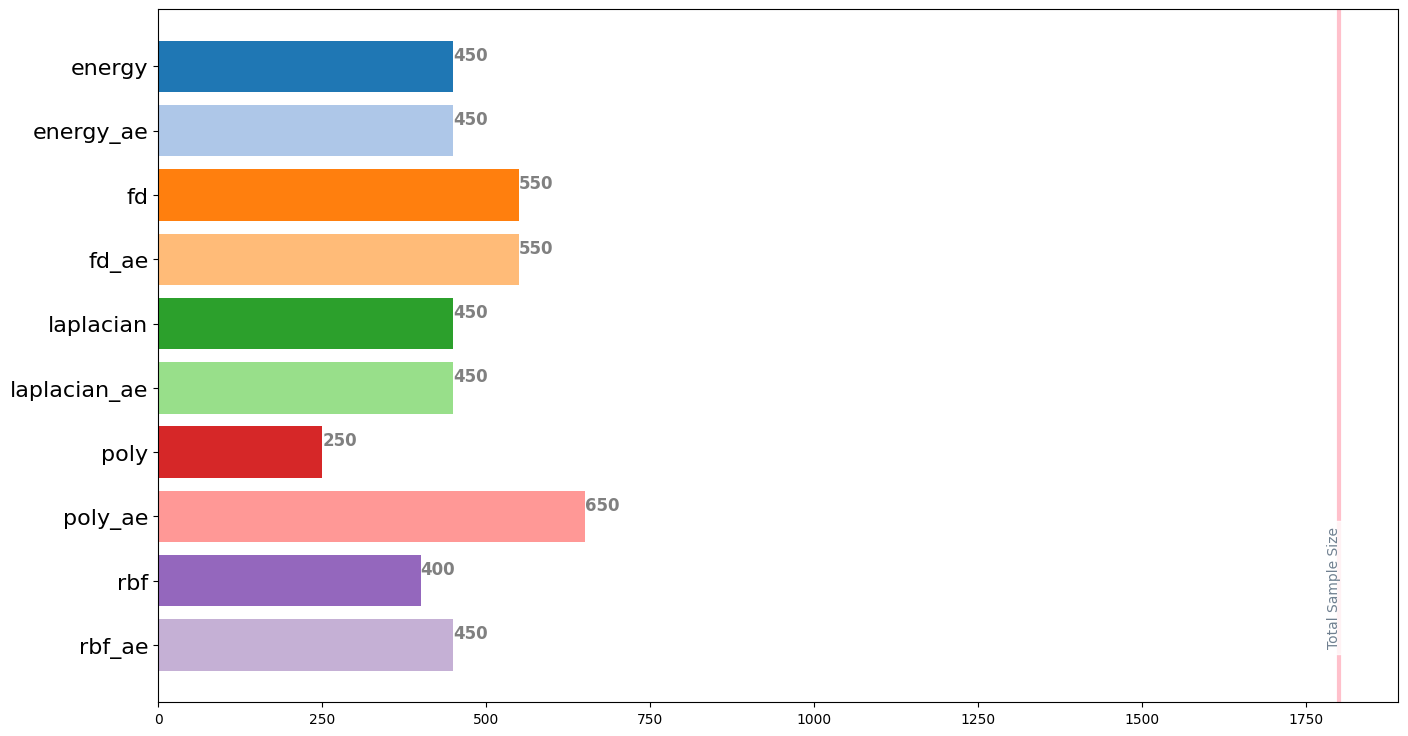}
      \\
      Sky & \includegraphics[width=\hsize]{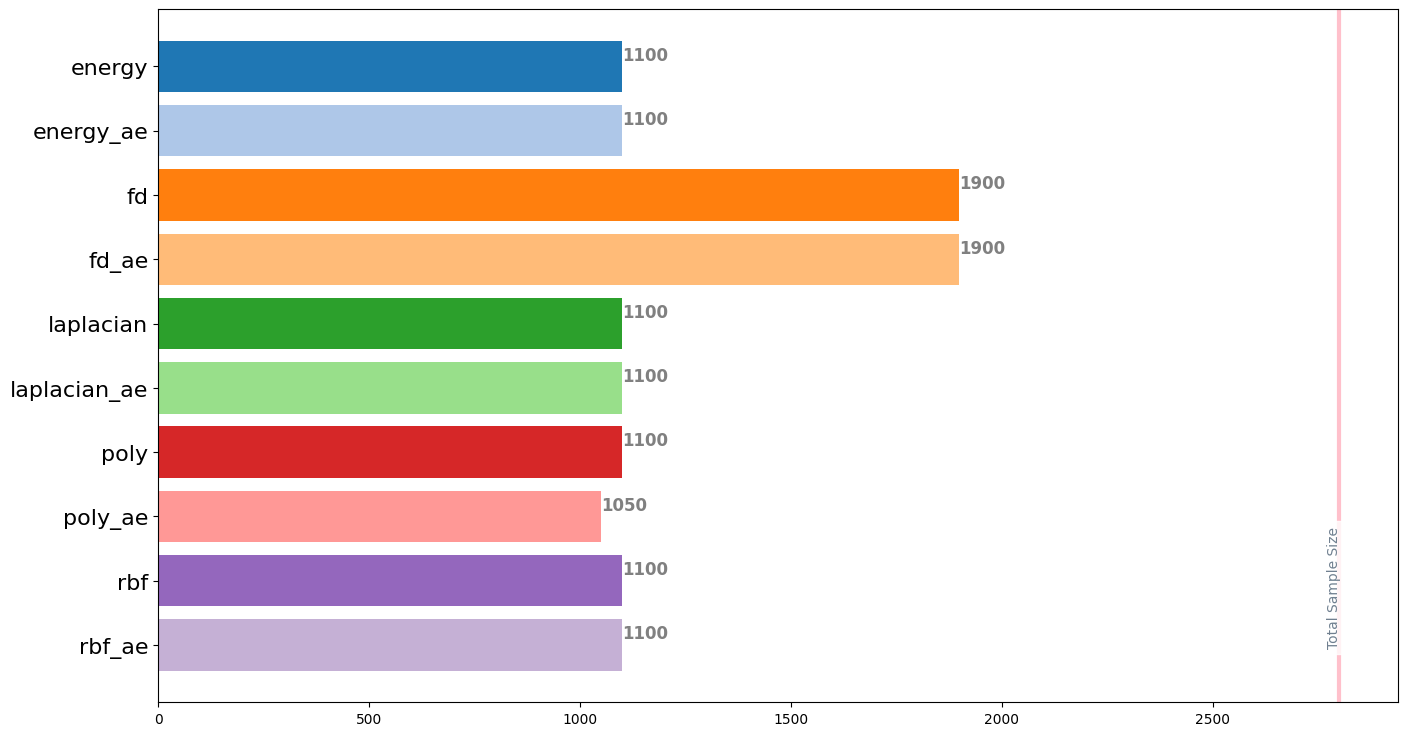}  
      & \includegraphics[width=\hsize]{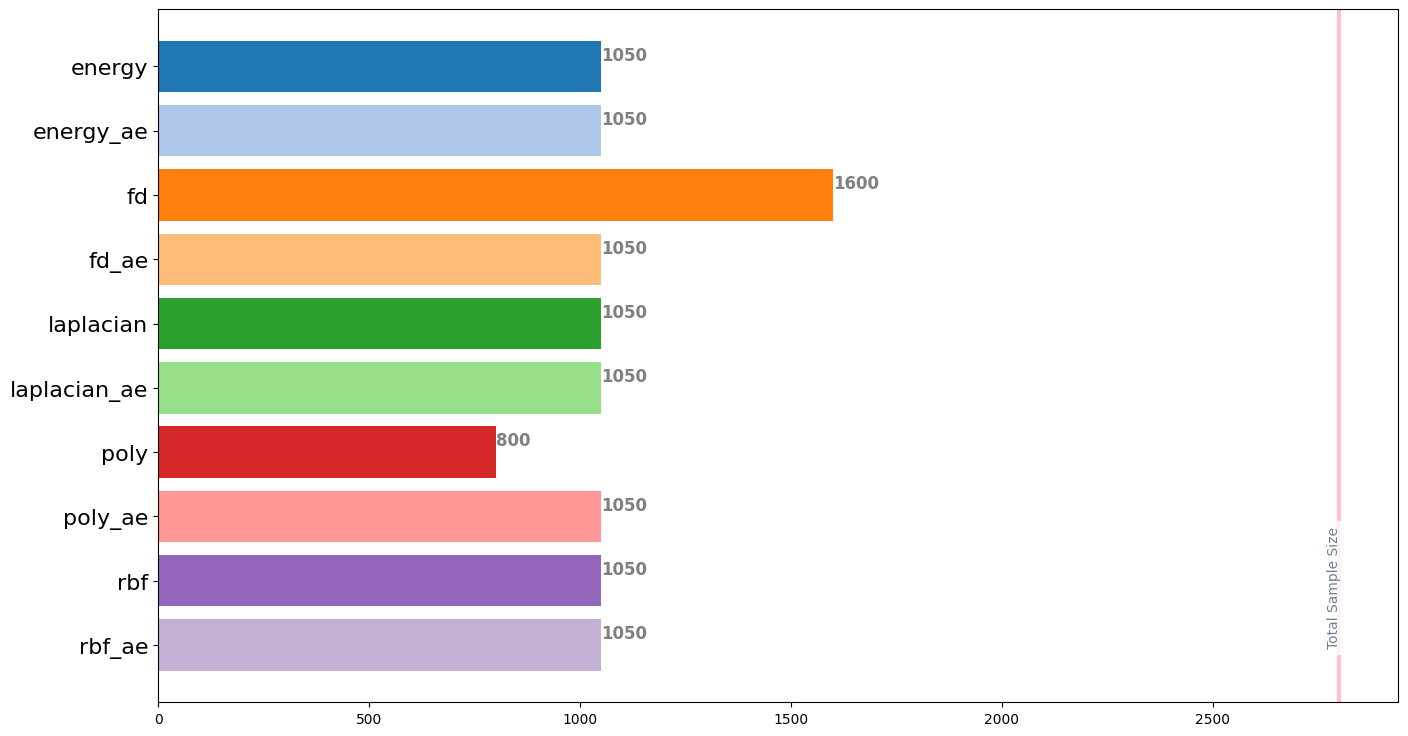}
      & \includegraphics[width=\hsize]{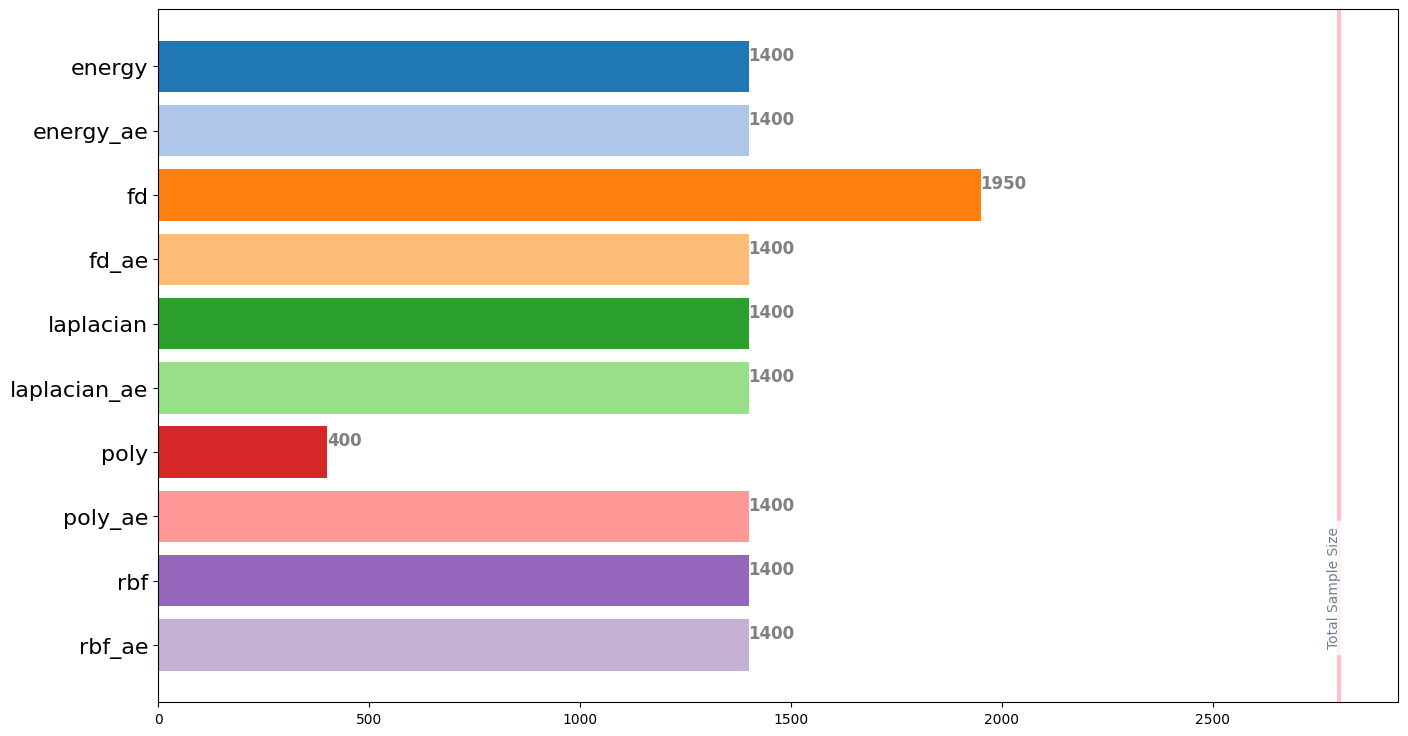}
      \\
      \end{tabular}}
    \caption{This figure shows the number of samples needed for \vjepapt~and \vjepaft~to achieve a 5\% error margin of the distance measured from 5,000 samples using the training and testing sets on most of the datasets presented in this study. The convergence requirement is stated in Figure~\ref{fig:number_sample_convergence}.}%
\label{fig:number_sample_convergence_all}
\end{figure}

Figures included in this section are:
\begin{description}
    \item 
    [\textbf{Figure ~\ref{fig:train_test_distances}}] The evolution of UCF-101 train-test distances for all metrics in all feature spaces.
    \item 
    [\textbf{Figure ~\ref{fig:train_test_convergence_rate}}] A comparison of convergence rates of FVD and \ourmetric, comparing training and testing sets on UCF101 and Something-Somethingv2.
    \item
    [\textbf{Figure ~\ref{fig:train_test_convergence_rate_other}}] Convergence rates UCF-101 train-test distances in \videomaept~and \vjepapt~feature spaces.
    \item
    [\textbf{Figure ~\ref{fig:number_sample_convergence_others}}] Visualization of the sample size required for \videomaept~and \vjepapt~features to converge to a 5\% error margin compared to the distance measured from 5,000 samples using the train and testing sets of UCF-101 and SSv2.
    \item[\textbf{Figure ~\ref{fig:number_sample_convergence_all}}]Visualization of the sample size required for \vjepapt~and \vjepapt~features to converge to a 5\% error margin compared to the distance measured using the train and testing sets of 9 other datasets. The pink vertical lines in the plots denote the sample size used to compute the target metric distance (ideally 5,000 samples). Note in Figures~\ref{fig:number_sample_convergence} and ~\ref{fig:number_sample_convergence_others} that UCF-101 and SSv2 have sufficient samples ($>5,000$) in their training and testing sets. However, many datasets in this graph have fewer samples. Importantly, convergence estimates become less reliable as bars approach the pink line, due to fewer iterations meeting the second convergence criterion (referring to Figure \ref{fig:number_sample_convergence}'s caption).
    \item[\textbf{Figure ~\ref{fig:bair_issue}}] The BAIR dataset~\citep{ebert2017bair} demonstrates a perfect example of the convergence issue due to insufficient samples. With 250 training videos, the estimated sample size for convergence (bars) nearly coincides with the target metric computation sample size (pink line). As convergence estimates degrade near this threshold due to insufficient iterations meeting the second criterion, it is difficult to confirm whether metrics truly converge at the displayed sample sizes, especially for Fr\'echet Distance-based metrics. 
\end{description}

% \begin{figure}[h]%
%     \centering
%     \setlength\tabcolsep{3pt} % default: 6pt
%     \centering
%     \begin{tabular}{c M{0.3\linewidth} M{0.3\linewidth} M{0.3\linewidth}}
%      & UCF with blur (low) &  UCF with blur (medium) & UCF with blur (high) \\
%     I3D & \includegraphics[width=\hsize]{figs/convergence_sample/i3d_ucf_32_test_blurlow_noise_num_sample.png}  
%       & \includegraphics[width=\hsize]{figs/convergence_sample/i3d_ucf_32_test_blur_noise_num_sample.png}
%       & \includegraphics[width=\hsize]{figs/convergence_sample/i3d_ucf_32_test_blurhigh_noise_num_sample.png}\\
%       \end{tabular}
%     \caption{This figure shows the number of samples needed to achieve a 10\% error margin of the distance measured from 5000 samples using the training and testing sets. }%
% \label{fig:number_sample_convergence_with_noise}
% \end{figure}

%% file: sections/appendix/noise_human_study.tex
\clearpage
\section{Noise Distortion Studies\label{sec:noise_distortion_stydies}}
\subsection{Complimentary Material for the Noise and Generative Model Study\label{sec:understanding_noise}}

\begin{figure}[h!]%
    \centering
    \setlength\tabcolsep{3pt} % default: 6pt
    \begin{tabular}{M{0.48\linewidth} M{0.48\linewidth}}
    \textbf{UCF-101} & \textbf{Sky Scene}  \\
    \includegraphics[width=\hsize]{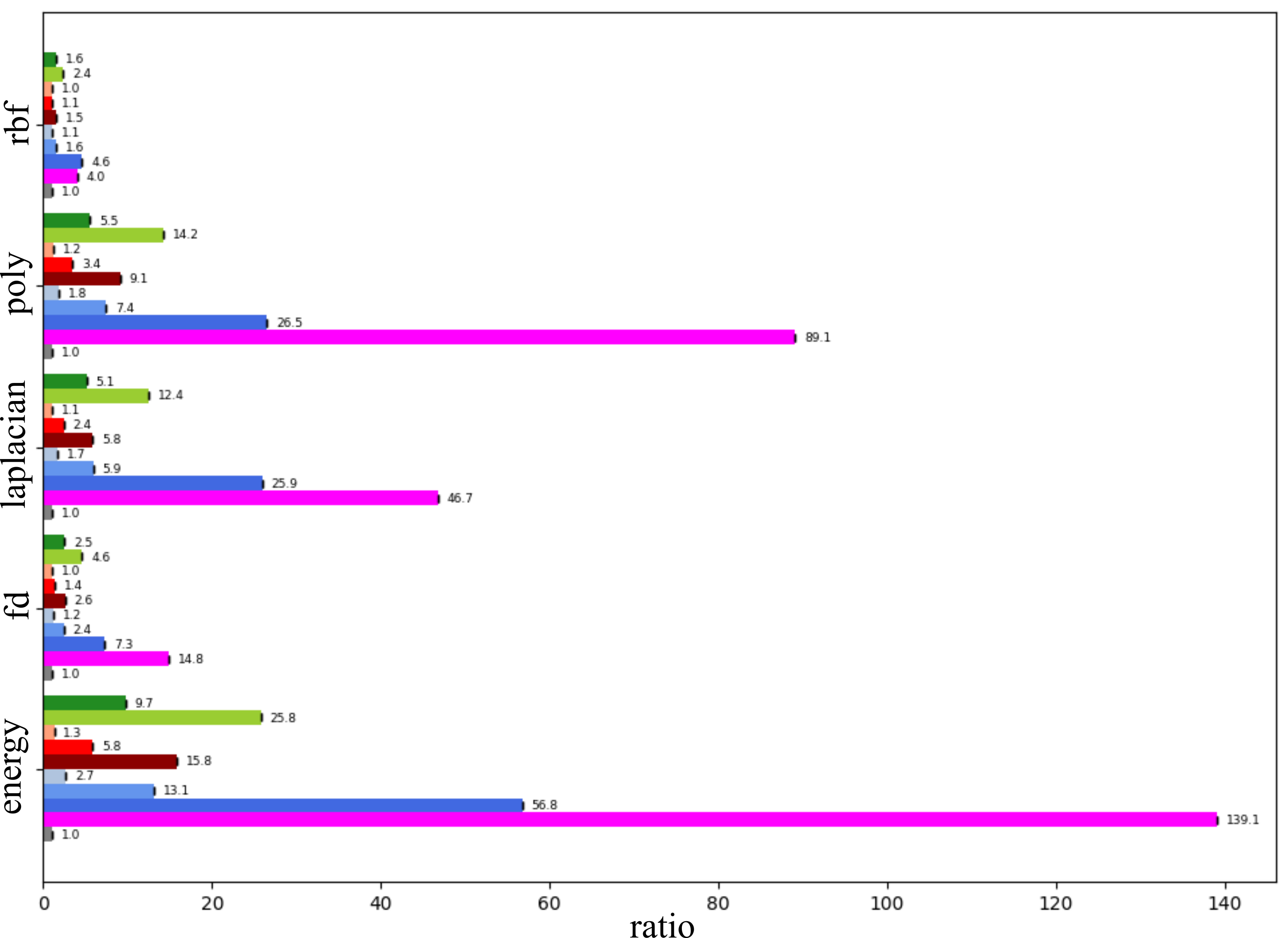}  
     & \includegraphics[width=\hsize]{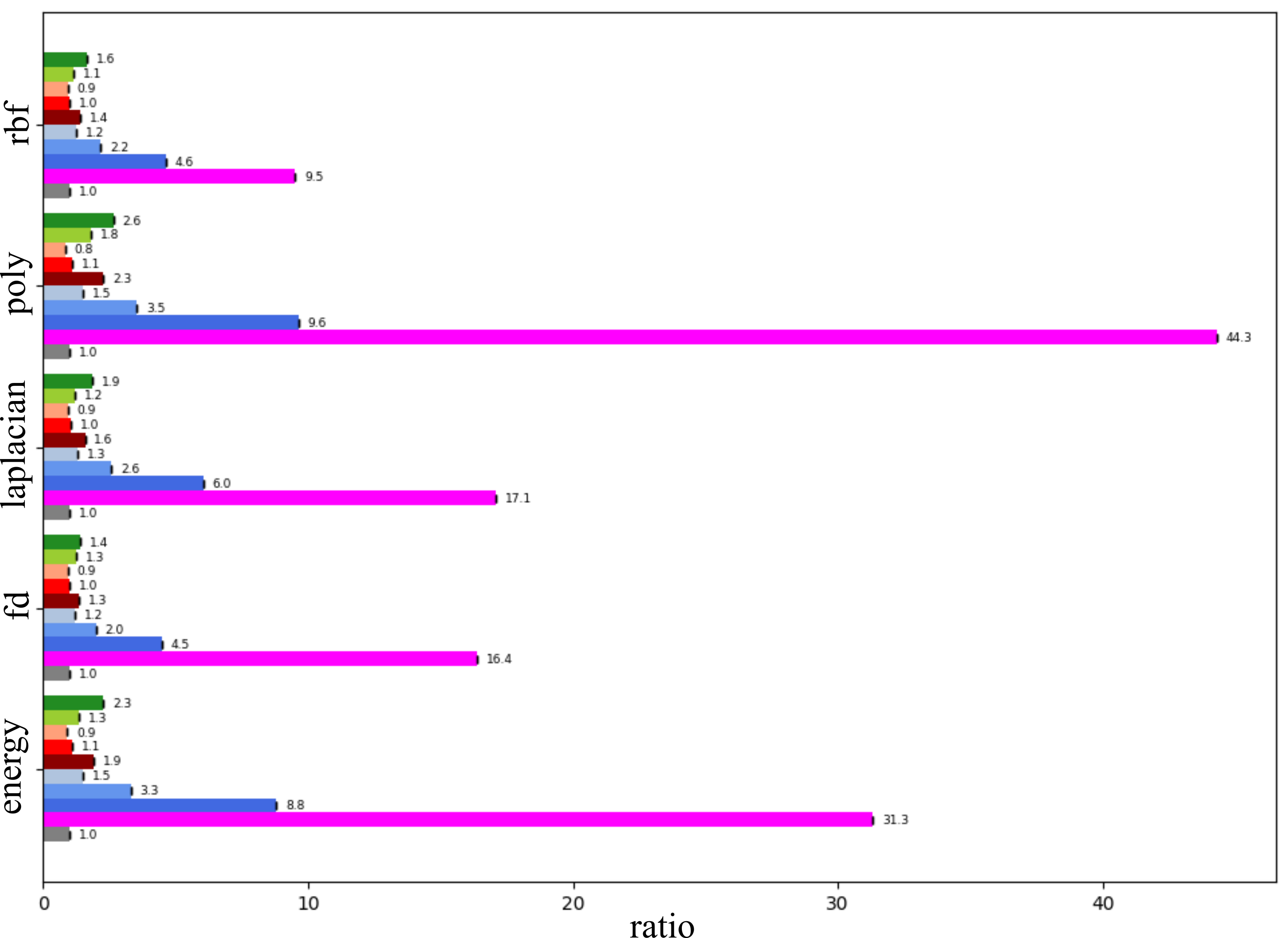}  \\
     \multicolumn{2}{c}{\includegraphics[width=0.95\hsize]{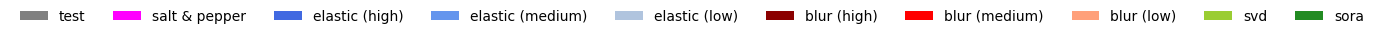}}\\
      \end{tabular}
    \caption{Comparing the FVD (I3D+FD) between training and test sets, using 5,000 samples from each, under various conditions, including noise and conditional generation. The train-test distance (gray) serves as a baseline for evaluating the reliability of the metric. For clarity, we normalize metric values using the train-test distance. The displayed bar values represent these scaled distances. \emph{Notably, the Sky Scene~\citep{xiong2018skyscene} experiment shows that low blur distortion brings the test distribution closer to the training distribution, highlighting a flaw in the FVD metric.}}%
    \label{fig:fvd_noise_impact}
    \raggedright \hyperlink{generative-specs}{\house} Back to paper
\end{figure}

\begin{figure}[h!]%
    \centering
    \setlength\tabcolsep{3pt} % default
    \begin{tabular}{c M{0.42\linewidth} M{0.42\linewidth}}
     & \textbf{UCF-101} & \textbf{Sky}  \\
    $\text{VMAE}_{\text{PT}}$& 
    \includegraphics[width=\hsize]{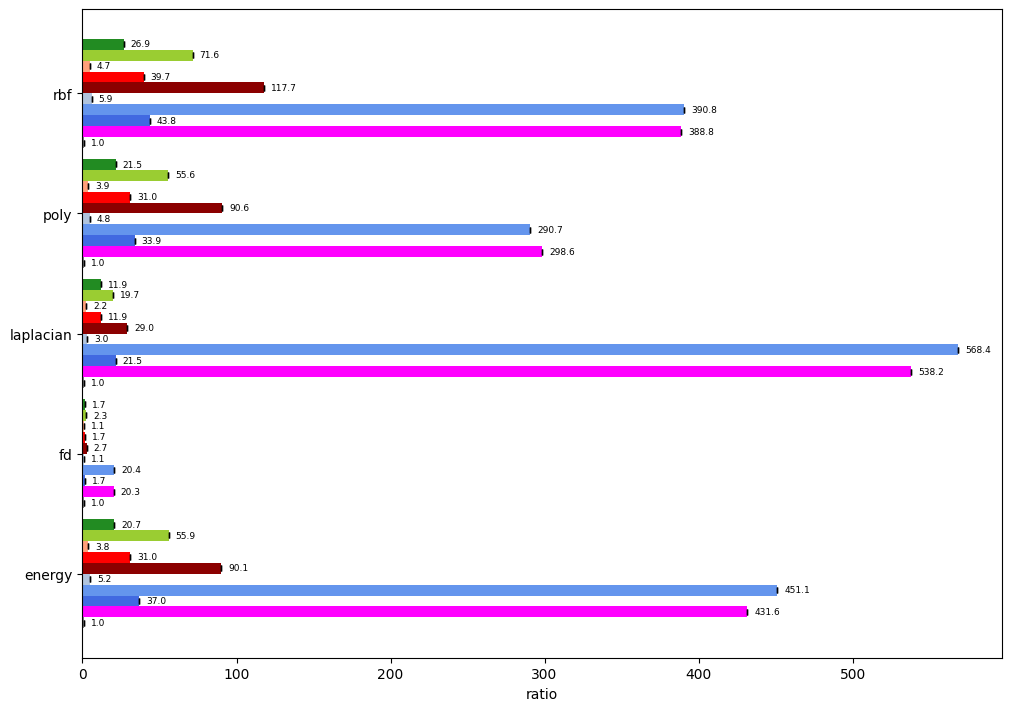}
     & \includegraphics[width=\hsize]{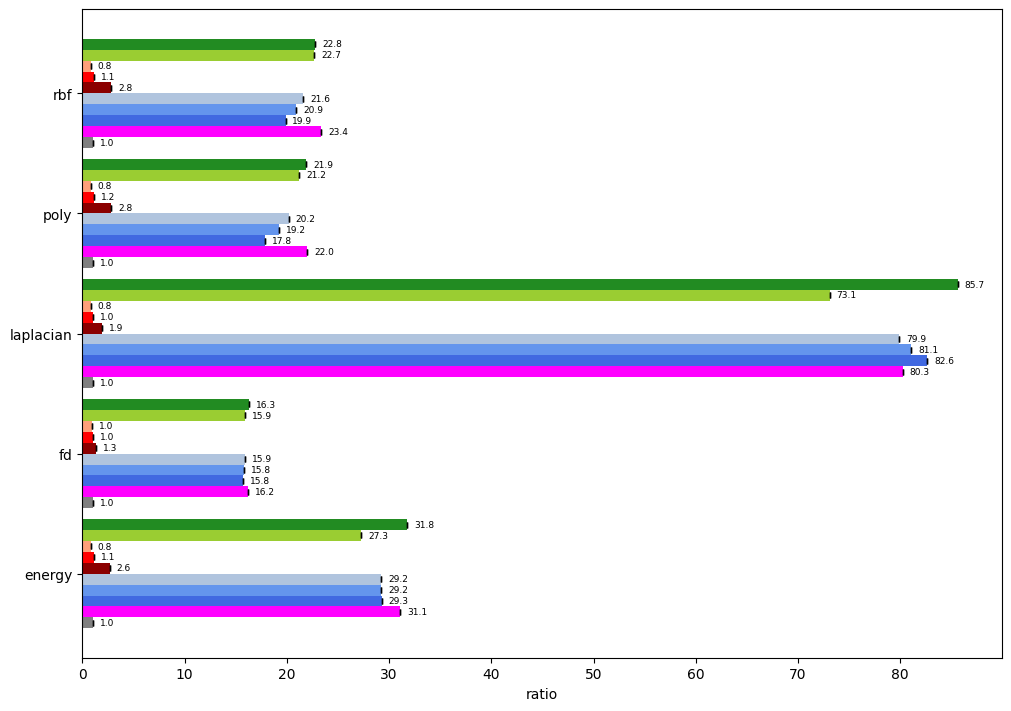}  \\
    $\text{VMAE}_{\text{SSv2}}$& 
     \includegraphics[width=\hsize]{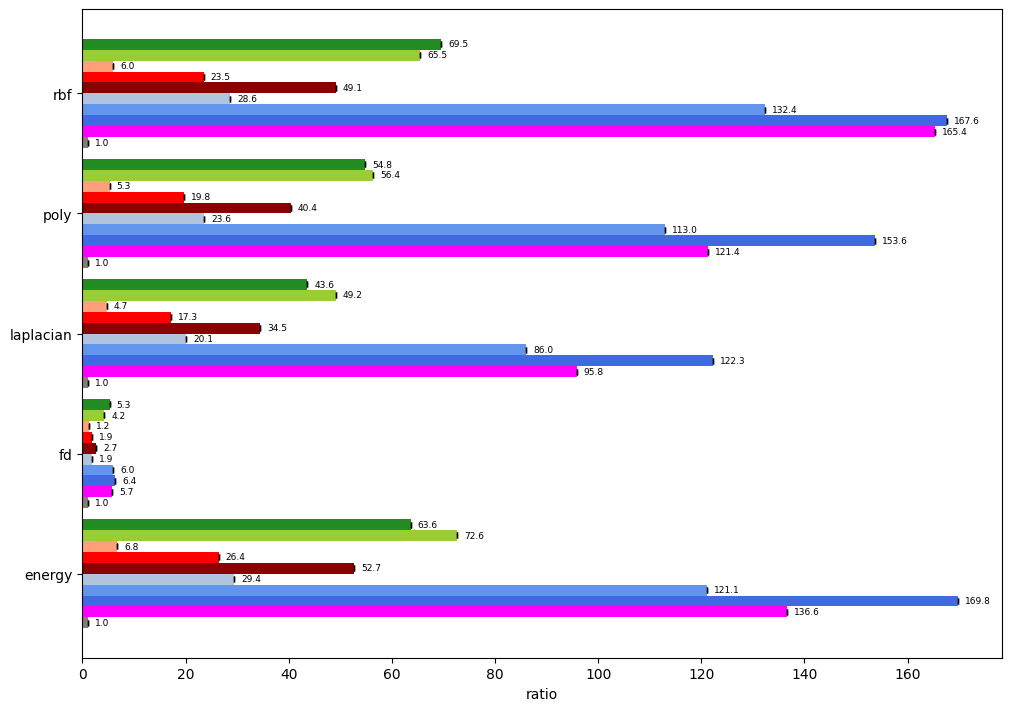}
     & \includegraphics[width=\hsize]{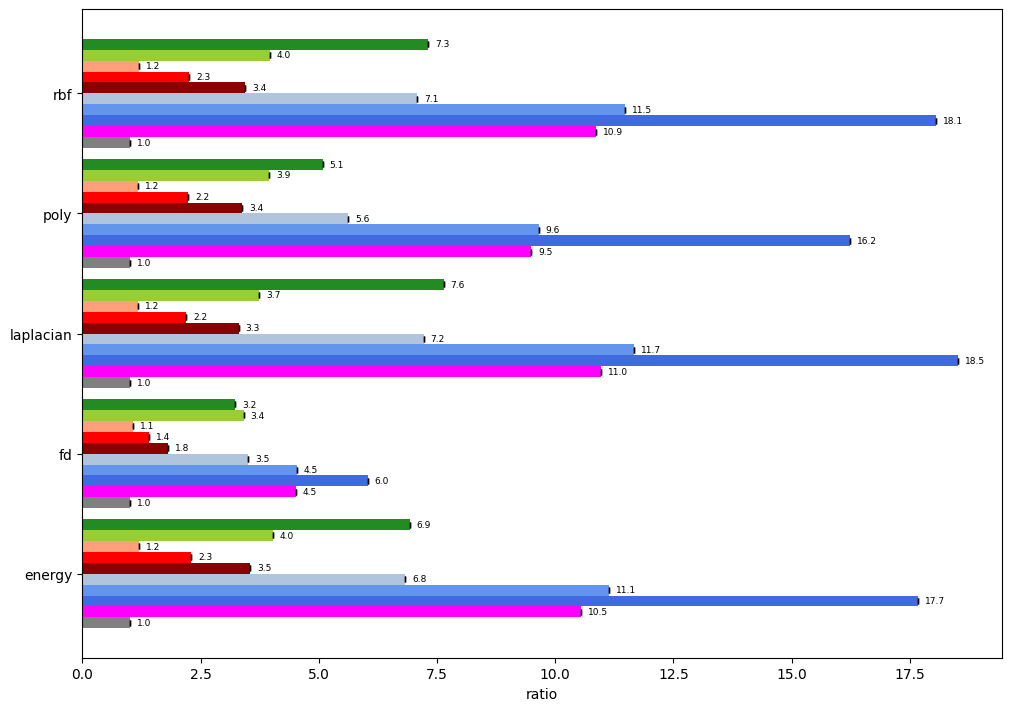}  \\
    \vjepapt &
    \includegraphics[width=\hsize]{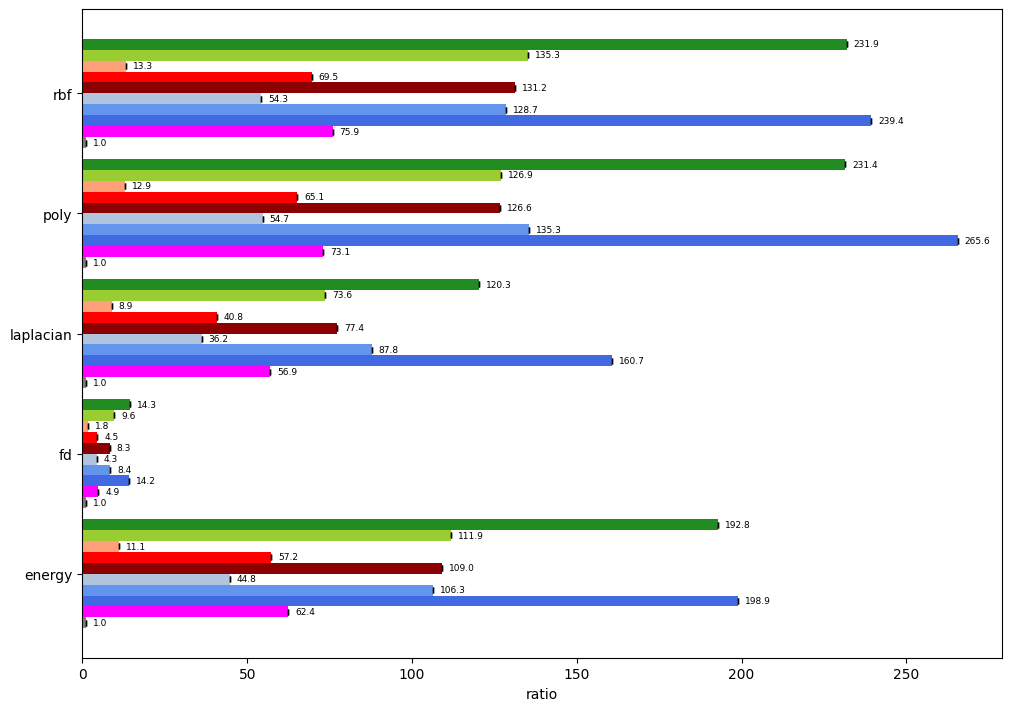}
    & \includegraphics[width=\hsize]{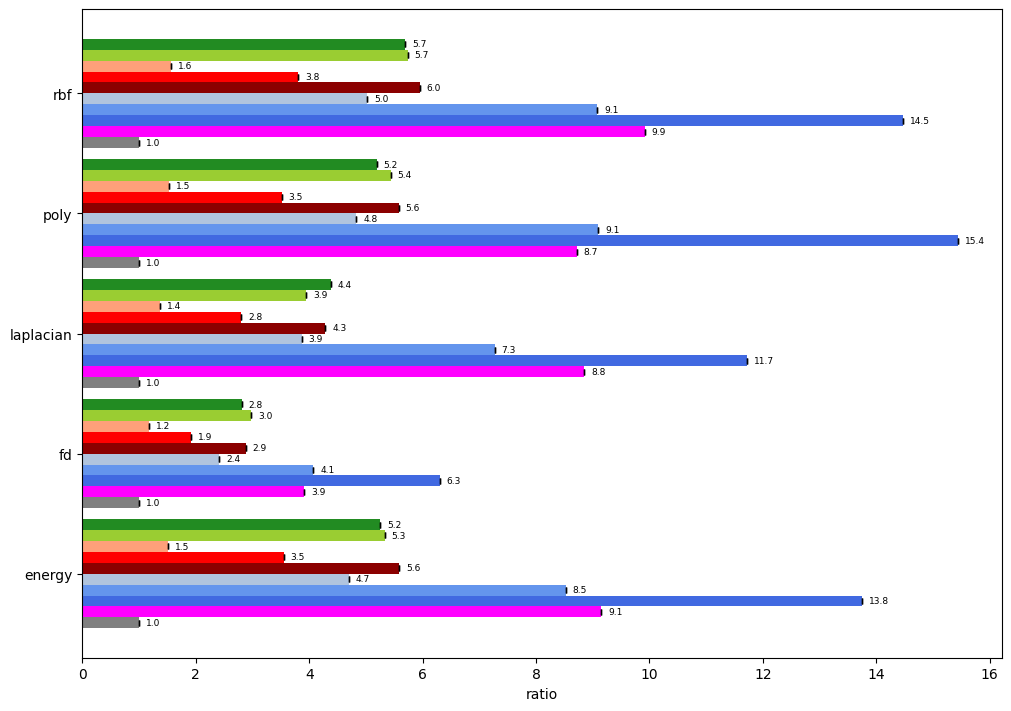}  \\
    \vjepaft &
    \includegraphics[width=\hsize]{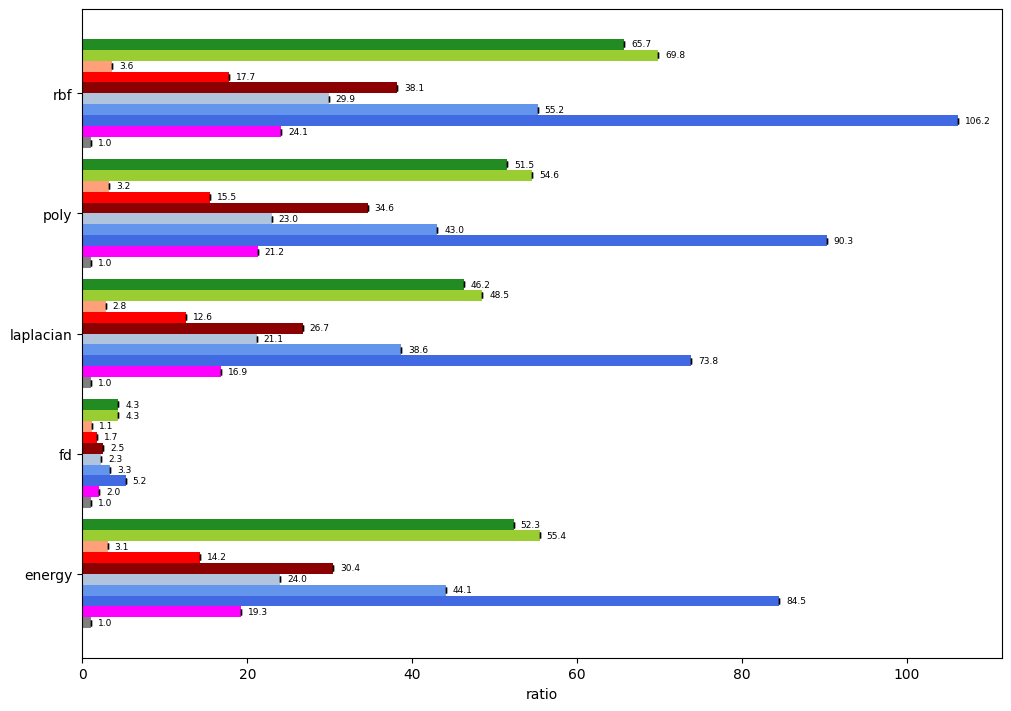}
    & \includegraphics[width=\hsize]{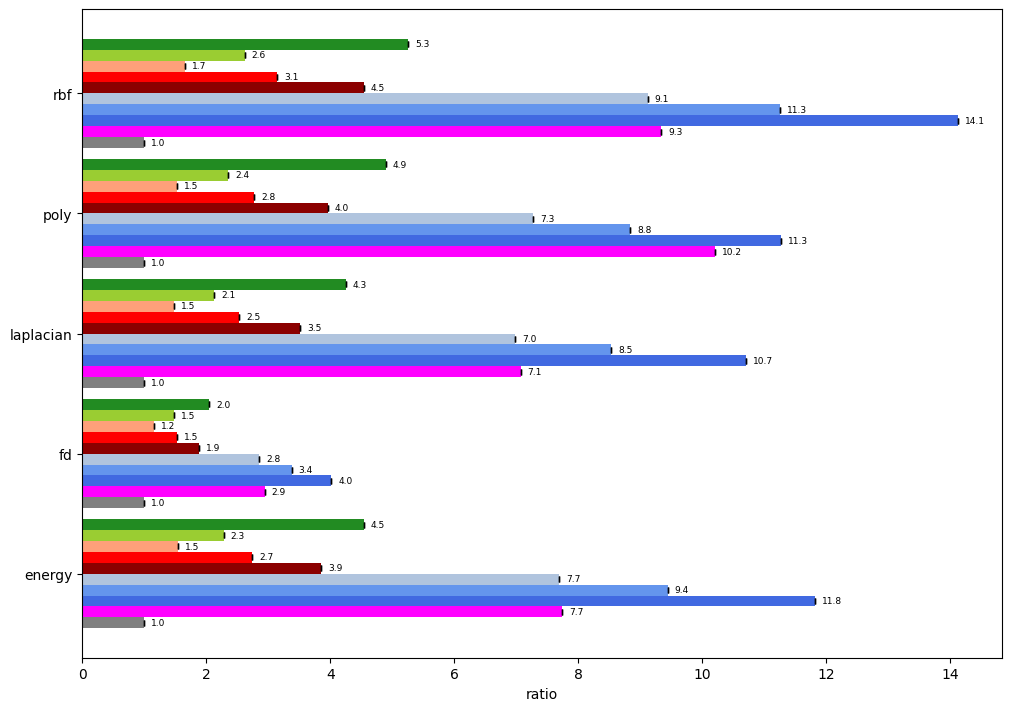}  \\
    \multicolumn{3}{c}{\includegraphics[width=0.95\hsize]{figs/noise_experiments/noise_legend.png}}\\
      \end{tabular}
    \caption{Experiments were conducted on the feature spaces of VideoMAE and V-JEPA using configurations analogous to those in Figure~\ref{fig:fvd_noise_impact}. Both the \vjepapt~and \videomaept~implementations require averaging across patch embeddings. Intuitively, \vjepapt~metrics are less affected by pixel-level salt and pepper noise as  its training is done in an abstract representation space, in direct contrast to that of a VideoMAE.}%
    \label{fig:FD_noise_impact_others}
    \raggedright \hyperlink{generative-specs}{\house} Back to paper
\end{figure}

Figure~\ref{fig:FD_noise_impact_others} illustrates the impact of noise and generative models on the metrics in VideoMAEs and V-JEPAs spaces. The study demonstrates a distinction in how different feature spaces rank these noise types. In Section~\ref{sec:human_study}, the findings from a human survey to determine which model has the closest ranking with human perception are reported.

\clearpage

\subsection{Noise and Convergence Rate\label{sec:noise_and_convergence}}
\hyperlink{noise-distortion}{\house} Back to paper

\begin{figure}[h!]%
    \setlength\tabcolsep{3pt} % default: 6pt
    \centering
    \begin{tabular}{c M{0.21\linewidth} M{0.21\linewidth} M{0.21\linewidth}M{0.21\linewidth} }
     & \textbf{No distortion} & \textbf{Blur (low)} & \textbf{Blur (med)} & \textbf{Blur (high)}\\
    UCF-101
    &\includegraphics[width=\hsize]{figs/fvd_ae_experiments/ucf_32_test_fvd_vs_final.png}   
    &\includegraphics[width=\hsize]{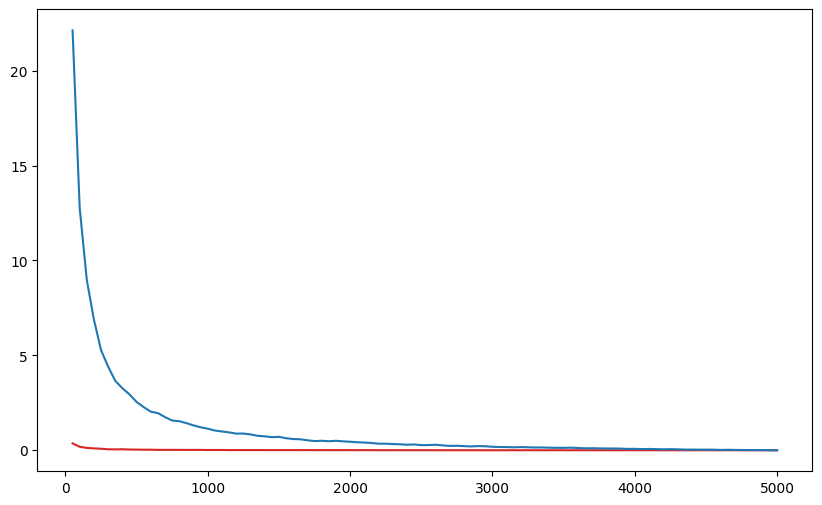}   
      & \includegraphics[width=\hsize]{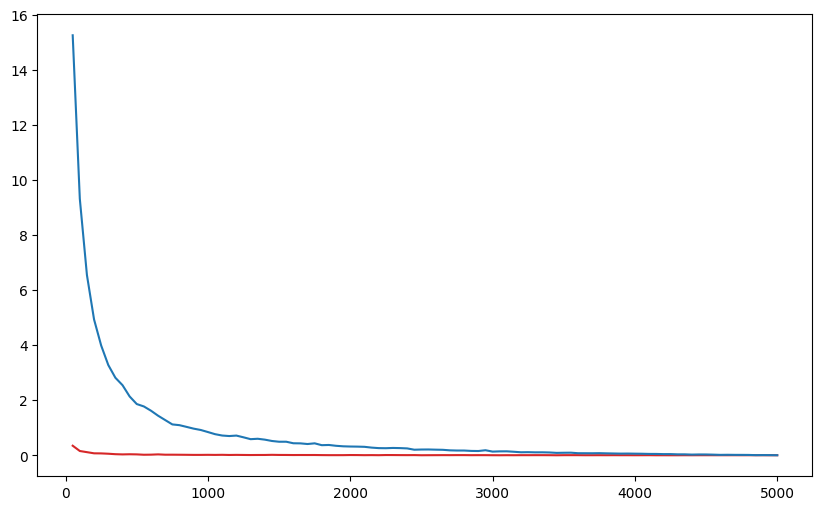}
      & \includegraphics[width=\hsize]{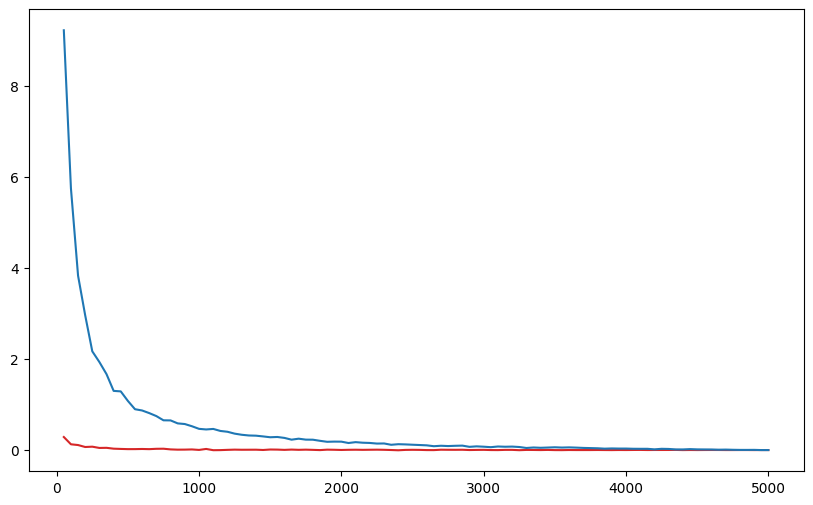}\\
    SSv2
    &\includegraphics[width=\hsize]{figs/fvd_ae_experiments/sthv2_32_test_fvd_vs_final.png}   
    &\includegraphics[width=\hsize]{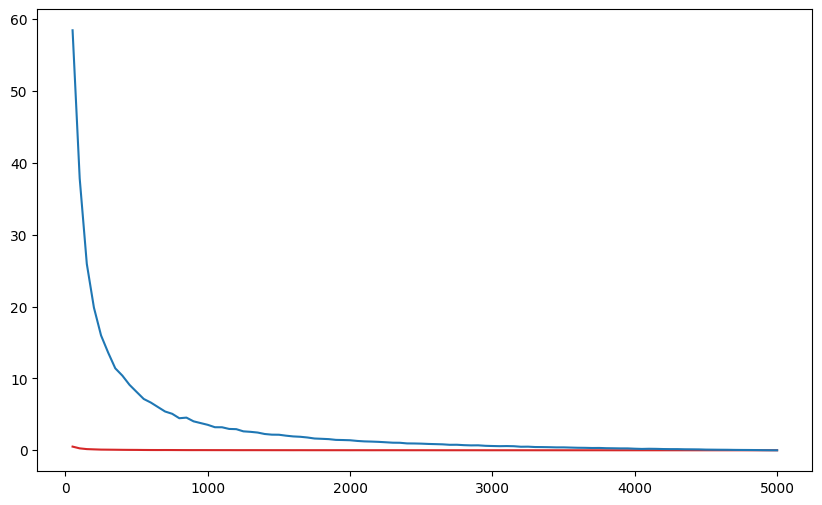}   
      & \includegraphics[width=\hsize]{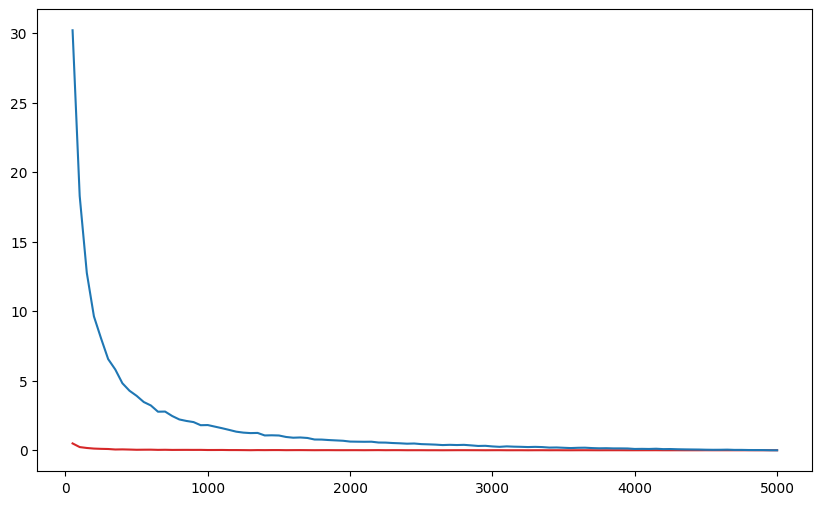}
      & \includegraphics[width=\hsize]{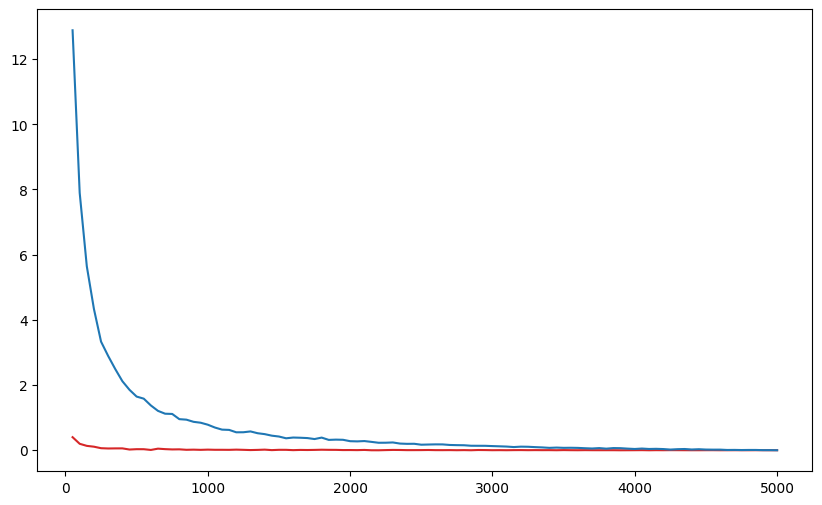}\\      \end{tabular}
      \caption{The convergence speed of FVD and the proposed metric is evaluated with blur distortion noises on the UCF-101 and Something-Something-v2 datasets. The convergence rate computation follows the same configuration as Figure \ref{fig:train_test_convergence_rate}. The blue line in the figure represents the convergence rate of FVD, while the red line represents the convergence rate of the proposed metric.\label{fig:noise_vs_metric_convergence}}
\end{figure}

Our study investigates the impact of noise and generative model outputs on the convergence speed of FVD and our proposed metric. Figure~\ref{fig:noise_vs_metric_convergence} reveals two key findings:
\begin{enumerate}[leftmargin=*,noitemsep,nolistsep]
\setlength{\itemsep}{1pt}
    \item 
    Our method exhibits significantly faster convergence than FVD even with added noise.
    \item 
    The presence of noise surprisingly accelerates convergence for both metrics.
\end{enumerate}

\subsection{Human Evaluation}\label{sec:human_eval}
\hyperlink{human-evaluation}{\house} Back to paper

To investigate human alignment on the perception of video quality degradation under various noise distortions, we conduct a small scale survey. We randomly select 24 videos from each of the UCF-101 and Sky Scene test sets, originally captured at 30 frames per second (fps), and subsample them to 25 frames at 7 fps to generate clips three seconds in length. Four types of noise distortions are systematically applied: high blur with a 7x7 kernel and a Gaussian standard deviation ranging from 0.01 to 3, medium blur with the same 7x7 kernel and a standard deviation ranging from 0.1 to 1.5, elastic distortion with a deformation strength of 30, and salt-and-pepper noise applied at a rate of 1\%.

\begin{figure}[h!]%
    \centering
    \includegraphics[width=0.7\hsize]{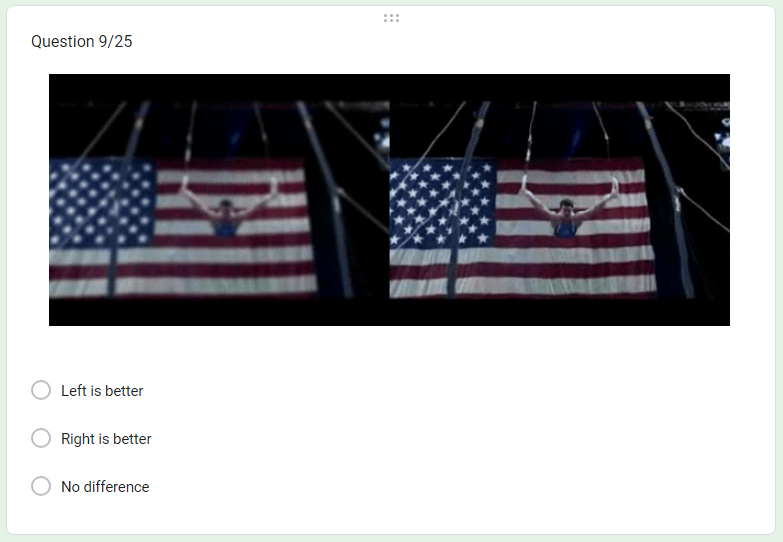}
    \caption{An example question in the human evaluation, noise distortion survey on the UCF-101 dataset.}%
\label{fig:noise_survey_sample1}
\end{figure}
\begin{figure}[h!]%
    \centering
    \includegraphics[width=0.7\hsize]{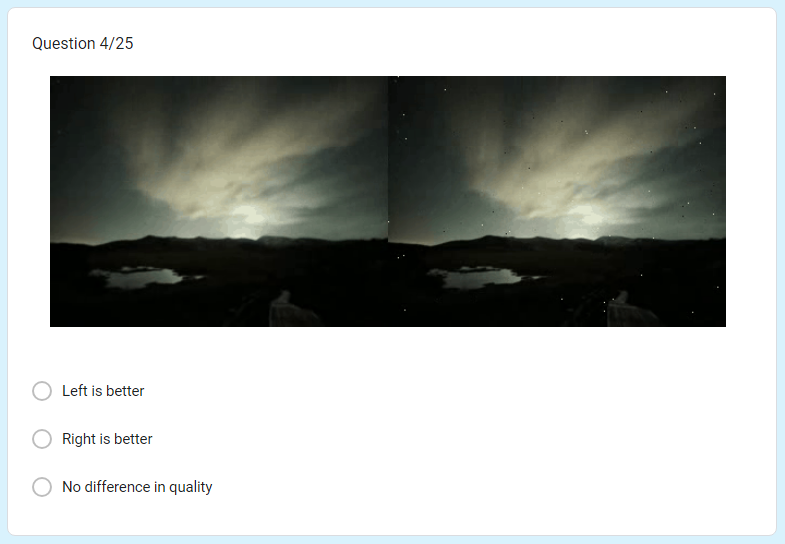}
    \caption{An example question in the human evaluation, noise distortion survey on the Sky Scene dataset.}%
\label{fig:noise_survey_sample2}
\end{figure}

Following the Analytic Hierarchy Process (AHP)~\citep{Saaty1987TheAH}, a pairwise comparison matrix is used to aggregate the responses. It is important to note that the sums of corresponding comparisons may not total 100\% due to the inclusion of an option indicating no discernible difference in quality between two videos (i.e., all self-to-self comparisons yield a value of 0\%). To account for this, the columns are normalized before computing the priority vector, which captures human preference over the different types of noise included in our study. Raw metric values can be found in Appendix~\ref{sec:understanding_noise}.

\begin{table}[h!]
\centering

\label{tab:pairwise_comparison}
\begin{tabular}{ccccc}
\toprule
\textbf{Noise Type} & \textbf{Blur (high)} & \textbf{Blur (medium)} & \textbf{Elastic (medium)} &\textbf{Salt and pepper} \\
\midrule
Blur (high) & 0.00\% & 0.00\% & 12.50\% & 2.50\% \\
Blur (med) & 93.75\% & 0.00\% & 68.75\% & 16.25\% \\
Elastic (med) & 81.25\% & 16.25\% & 0.00\% & 3.75\% \\
Salt and pepper & 95.00\% & 69.68\% & 93.75\% & 0.00\% \\
\bottomrule
\end{tabular}
\caption{Pairwise comparison matrix, noise distortion (UCF-101). Results are aggregated from 20 participants.}
\end{table}

\begin{table}[h!]
\centering

\label{tab:pairwise_comparison2}
\begin{tabular}{ccccc}
\toprule
\textbf{Noise Type} & \textbf{Blur (high)} & \textbf{Blur (medium)} & \textbf{Elastic (medium)} &\textbf{Salt and pepper} \\
\midrule
Blur (high) & 0.00\% & 7.63\% & 33.23\% & 37.50\% \\
Blur (med) & 68.23\% & 0.00\% & 41.25\% & 48.75\% \\
Elastic (med) & 56.78\% & 42.50\% & 0.00\% & 41.25\% \\
Salt and pepper & 53.75\% & 46.25\% & 51.25\% & 0.00\% \\
\bottomrule
\end{tabular}
\caption{Pairwise comparison matrix, noise distortion (Sky Scene). Results are aggregated from 20 participants.}
\end{table}

\begin{table}[h!]

\label{tab:human_evaluation_results}
\centering
\small
\begin{tabular}{ccccc}
\toprule
\textbf{Rank} & \textbf{Method} & \textbf{Feature Space} & \textbf{UCF-101 (\%)} & \textbf{Sky Scene (\%)} \\
\midrule
1 & Energy & \vjepapt & 83.26 & 84.06\\
2 & $\text{MMD}_{\text{RBF}}$ & \vjepapt & 83.02 & 84.07\\
3 & $\text{MMD}_{\text{LAP}}$ & \vjepapt+AE & 83.02 & 85.60\\
4 & FD & \vjepapt & 82.31 & 86.82\\
5 & $\text{MMD}_{\text{POLY}}$ & \vjepapt & 81.82 & 81.07\\
6 & Energy & \vjepapt+AE & 80.88 & 83.17\\
7 & FD & \vjepapt+AE & 79.33 & 84.13\\
\rowcolor{lightgray}\textbf{8} & \textbf{$\text{MMD}_{\text{POLY}}$} & \textbf{\vjepaft} & \textbf{78.96} & \textbf{78.39}\\
9 & $\text{MMD}_{\text{LAP}}$ & \vjepapt & 78.85 & 81.34\\
10 & $\text{MMD}_{\text{POLY}}$ & \vjepapt+AE & 78.78 & 82.33\\
11 & $\text{MMD}_{\text{RBF}}$ & \vjepaft & 75.51 & 79.10\\
12 & $\text{MMD}_{\text{LAP}}$ & \vjepaft & 75.49 & 79.98\\
13 & Energy & \vjepaft & 75.42 & 79.85\\
14 & $\text{MMD}_{\text{RBF}}$ & \vjepapt+AE & 75.32 & 94.67\\
15 & FD & \vjepaft & 74.63 & 86.32\\
16 & $\text{MMD}_{\text{RBF}}$ & I3D+AE & 70.32 & 75.43\\
17 & $\text{MMD}_{\text{RBF}}$ & \vjepaft+AE & 63.78 & 84.03\\
18 & FD & \vjepaft+AE & 61.15 & 84.12\\
19 & Energy & \vjepaft+AE & 59.12 & 79.09\\
20 & $\text{MMD}_{\text{POLY}}$ & \vjepaft+AE & 58.47 & 78.60\\
21 & $\text{MMD}_{\text{LAP}}$ & \vjepaft+AE & 57.93 & 80.34\\
22 & FD & \videomaeft & 55.72 & 79.08\\
23 & Energy & \videomaeft & 51.76 & 74.15\\
24 & $\text{MMD}_{\text{RBF}}$ & I3D & 50.93 & 68.07\\
25 & $\text{MMD}_{\text{LAP}}$ & \videomaeft & 50.16 & 72.76\\
26 & $\text{MMD}_{\text{POLY}}$ & \videomaeft+AE & 49.76 & 75.54\\
27 & $\text{MMD}_{\text{POLY}}$ & \videomaeft & 48.34 & 75.35\\
28 & Energy & \videomaeft+AE & 46.56 & 72.09\\
29 & $\text{MMD}_{\text{RBF}}$ & \videomaeft & 46.36 & 73.47\\
30 & $\text{MMD}_{\text{POLY}}$ & \videomaept & 46.14 & 62.87 \\
31 & $\text{MMD}_{\text{RBF}}$ & \videomaept & 45.98 & 62.29\\
32 & $\text{MMD}_{\text{LAP}}$ & \videomaeft+AE & 42.85 & 69.20 \\
33 & FD & \videomaeft+AE & 42.38 & 72.42 \\
34 & Energy & \videomaept & 40.03 & 60.73\\
35 & $\text{MMD}_{\text{RBF}}$ & \videomaeft+AE & 37.90 & 73.47\\
36 & Energy  & \videomaept+AE & 37.64 & 58.12\\
37 & $\text{MMD}_{\text{LAP}}$ & \videomaept+AE & 37.47 & 57.99\\
38 & $\text{MMD}_{\text{POLY}}$ & \videomaept+AE & 36.88 & 58.06\\
39 & FD & \videomaept & 35.36 & 61.94\\
40 & $\text{MMD}_{\text{RBF}}$ & \videomaept+AE & 32.44 & 57.04\\
41 & FD & \videomaept+AE & 32.01 & 57.42\\
42 & FD & I3D+AE & 31.66 & 57.54\\
\rowcolor{lightgray} \textbf{43} & \textbf{FD} & \textbf{I3D} & \textbf{31.32} & \textbf{58.07}\\
44 & $\text{MMD}_{\text{LAP}}$ & \videomaept & 29.90 & 57.93\\
45 & $\text{MMD}_{\text{POLY}}$ & I3D+AE & 25.97 & 67.25\\
46 & $\text{MMD}_{\text{LAP}}$ & I3D & 25.55 & 54.57\\
47 & Energy & I3D+AE & 25.52 & 48.07\\
48 & $\text{MMD}_{\text{LAP}}$ & I3D+AE & 25.17 & 57.30\\
49 & Energy & I3D & 23.74 & 47.25\\
50 & $\text{MMD}_{\text{POLY}}$ & I3D & 22.38 & 67.17\\
\bottomrule
\end{tabular}
\caption{Human Evaluation: Cosine Similarity Results (Ranked on UCF-101). $\text{VJEPA}_{\text{SSv2}}$+$\text{MMD}_\text{POLY}$ and FVD (I3D+FD) results are highlighted.}
\end{table}

%% file: sections/appendix/generative_models.tex
\section{Generative Model Specifications~\label{sec:generative_model}}
\hyperlink{generative-specs}{\house} Back to paper

\subsection{Open-Sora}

For videos generated from the UCF-101 dataset, a single frame was provided as the image prompt with its corresponding video class label given as the text prompt. For videos generated from the Sky Scene dataset, a single frame was provided as the image prompt with ``A sky timelapse'' provided as the text prompt.

Videos were generated at an output resolution of 240p with an aspect ratio of 3:4. The '4s' preset for the number of frames was used which translates to 102 frames at 24 fps. These videos were converted to 16 frames at 7 fps for both the metric calculations. We used the open-source implementation loaded with the hpcai-tech/OpenSora-VAE-v1.2 checkpoint for inference.

\subsection{Ctrl-V}

\begin{figure}[ht!]
    \setlength\tabcolsep{3pt} % default: 6pt
    \centering
    \begin{tabular}{@{} r M{0.18\linewidth} M{0.18\linewidth} M{0.18\linewidth} M{0.18\linewidth} M{0.18\linewidth} @{}}
    & Frame 1 & Frame 5 & Frame 9 & Frame 13  & Frame 16\\
      Iter 0 & \includegraphics[width=\hsize]{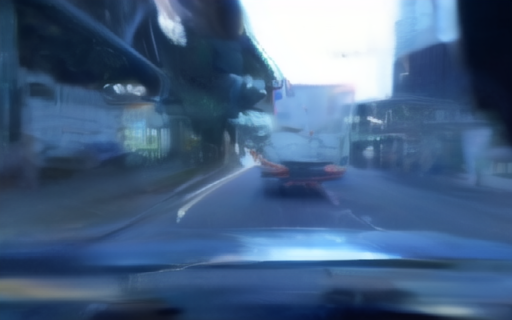}
      & \includegraphics[width=\hsize]{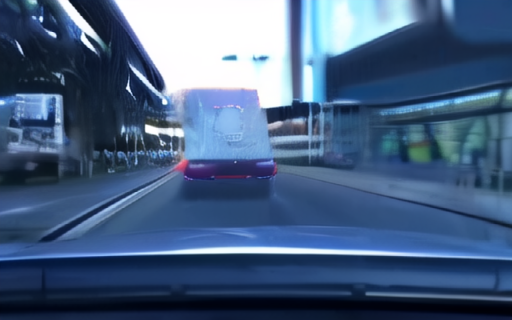}  
      & \includegraphics[width=\hsize]{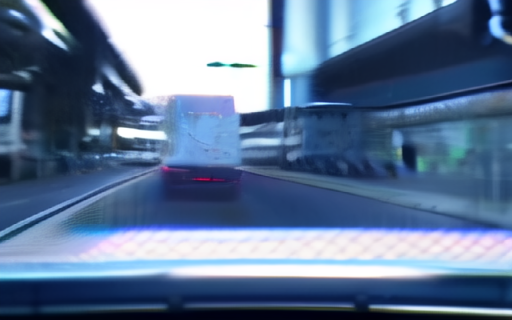}
      & \includegraphics[width=\hsize]{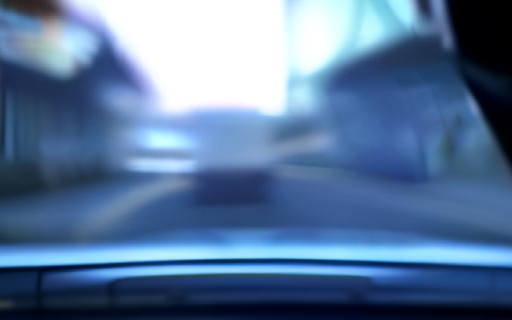} 
      & \includegraphics[width=\hsize]{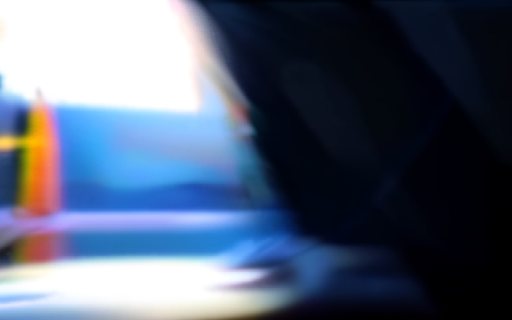}\\
      Iter 1 & \includegraphics[width=\hsize]{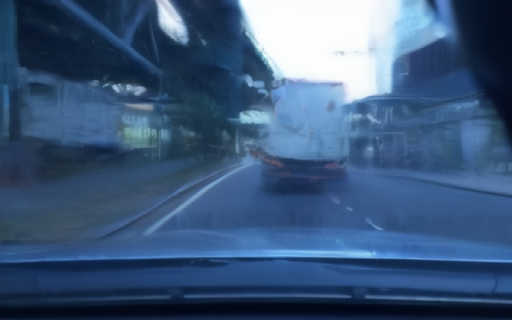}
      & \includegraphics[width=\hsize]{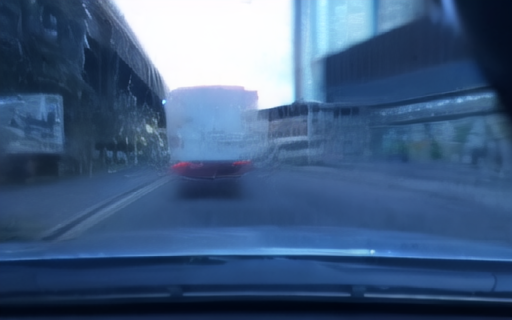}  
      & \includegraphics[width=\hsize]{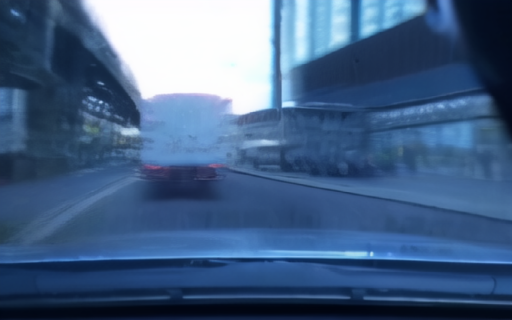}
      & \includegraphics[width=\hsize]{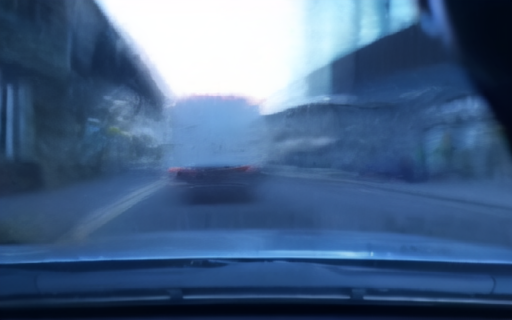} 
      & \includegraphics[width=\hsize]{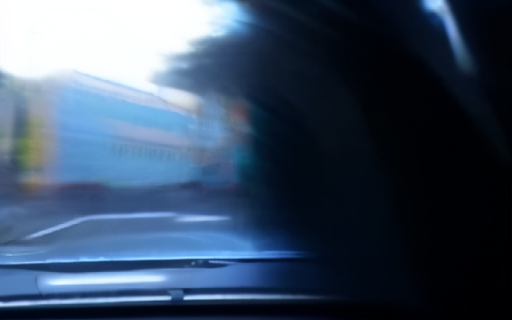}\\
      Iter 1200 & \includegraphics[width=\hsize]{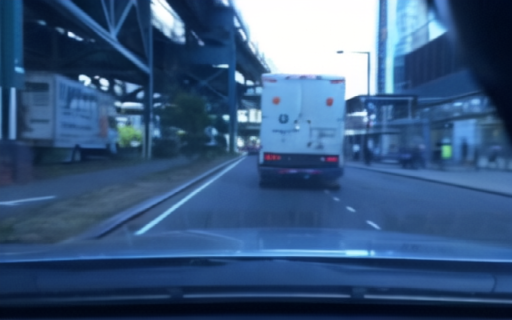}
      & \includegraphics[width=\hsize]{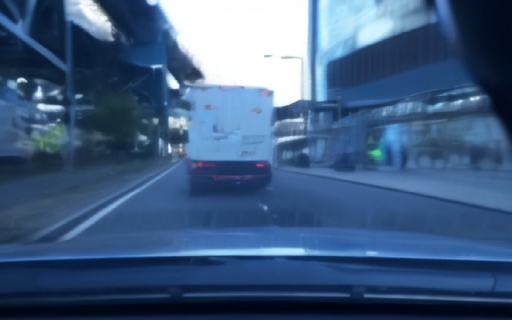}  
      & \includegraphics[width=\hsize]{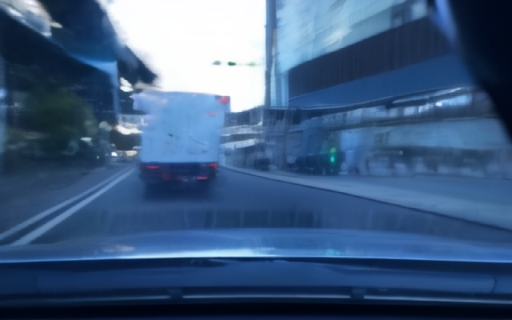}
      & \includegraphics[width=\hsize]{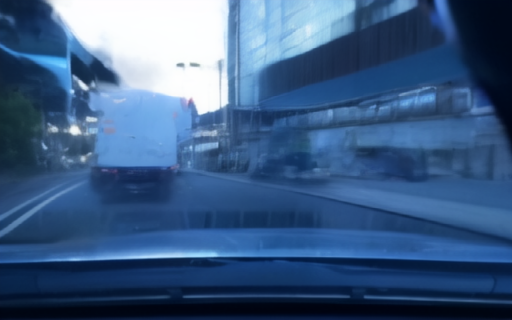} 
      & \includegraphics[width=\hsize]{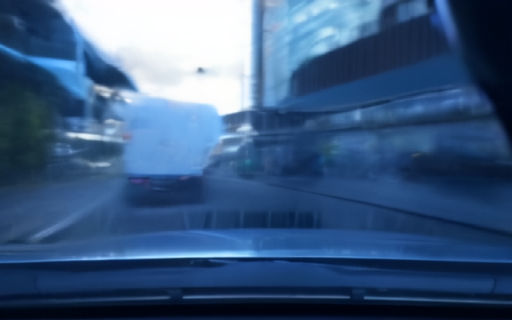}\\
\end{tabular}
    \caption{Training progression of Ctrl-v SVD generation at iteration 0, 1 and 1200.}
    \label{fig:ctrlv_training}
\end{figure}
\hyperlink{ctrlv-training}{\house} Back to paper 
 
We use the original code and model configuration from Ctrl-v~\citep{luo2024ctrlv} in this study, and their Stable Video Diffusion's model backbone is HuggingFace's \texttt{stabilityai/stable-video-diffusion-img2vid-xt} model. Figure~\ref{fig:ctrlv_training} contains a demo of samples generated from the Ctrl-v's SVD checkpoints.